\documentclass{egpubl_arxiv}
\usepackage{egsr2023}
 
 \JournalSubmission    
\usepackage[T1]{fontenc}
\usepackage{dfadobe}
\usepackage{bm}
\usepackage{float}
\usepackage{cite}  

\BibtexOrBiblatex
\electronicVersion
\PrintedOrElectronic
\ifpdf \usepackage[pdftex]{graphicx} \pdfcompresslevel=9
\else \usepackage[dvips]{graphicx} \fi

\usepackage{egweblnk} 
\usepackage{svg}

\usepackage[english]{babel}

\usepackage[letterpaper,top=2cm,bottom=2cm,left=3cm,right=3cm,marginparwidth=1.75cm]{geometry}

\usepackage[export]{adjustbox}
\usepackage{amssymb}
\usepackage{amsmath}
\usepackage{graphicx}
\usepackage{caption}
\usepackage{subcaption}
\usepackage{xcolor}

\usepackage{mathtools,stmaryrd,nicefrac,bbm}
\usepackage[many]{tcolorbox}
\usepackage{ifthen,comment,xspace}
\usepackage{overpic}
\usepackage[outline]{contour}
\contourlength{0.075em}
\usepackage{tikz}

\usepackage{listings}
\definecolor{codegreen}{rgb}{0,0.6,0}
\definecolor{codegray}{rgb}{0.5,0.5,0.5}
\definecolor{codepurple}{rgb}{0.58,0,0.82}
\definecolor{backcolour}{rgb}{0.95,0.95,0.92}

\lstdefinestyle{mystyle}{
    backgroundcolor=\color{white},   
    commentstyle=\color{codegreen},
    keywordstyle=\color{magenta},
    numberstyle=\tiny\color{codegray},
    stringstyle=\color{codepurple},
    basicstyle=\ttfamily\footnotesize,
    breakatwhitespace=false,         
    breaklines=true,                 
    captionpos=b,                    
    keepspaces=true,                 
    numbers=left,                    
    numbersep=5pt,                  
    showspaces=false,                
    showstringspaces=false,
    showtabs=false,                  
    tabsize=2
}

\newcommand{\final}[1]{#1}

\definecolor{orange_cubic}{rgb}{.9765, .5887, .3569}
\definecolor{purple_cubic}{rgb}{.4706, 0, .5216}
\definecolor{green_cubic}{rgb}{.28603, .81178, .5008}

\definecolor{grayLL}{rgb}{.98, .98, .98}
\definecolor{grayL}{rgb}{.9, .9, .9}
\definecolor{purpleL}{rgb}{.9735, .95, .9761}
\definecolor{purpleD}{rgb}{.8941, .8, .9043}
\definecolor{greenL}{rgb}{.9643, .9906, .9750}
\definecolor{greenD}{rgb}{.7145, .9249, .7999}
\definecolor{greenDD}{rgb}{.3145, .6249, .3999}
\definecolor{orangeLL}{rgb}{0.9991, 0.9846, 0.9759}
\definecolor{orangeL}{rgb}{.9982, .9692, .9518}
\definecolor{orangeD}{rgb}{.9929, .8766, .8071}

\definecolor{redL}{rgb}{1.0, 0.95, 0.95}
\definecolor{redD}{rgb}{1.0, 0.8, 0.8}
\definecolor{redDD}{rgb}{1.0, 0.4, 0.4}
\definecolor{yellowL}{rgb}{1.0, 1.0, 0.95}
\definecolor{yellowD}{rgb}{0.95, 0.95, 0.6}
\definecolor{yellowDD}{rgb}{0.8, 0.8, 0.2}
\definecolor{blueLL}{rgb}{0.98, 0.98, 1.0}
\definecolor{blueL}{rgb}{0.95, 0.95, 1.0}
\definecolor{blueD}{rgb}{0.8, 0.8, 1.0}
\definecolor{blueDD}{rgb}{0.6, 0.6, 1.0}

\definecolor{uciBlue}{RGB}{0,100,164}
\definecolor{uciBlueL}{RGB}{127.5, 177.5, 209.5}
\definecolor{uciOrange}{RGB}{247,141,45}
\definecolor{uciOrangeL}{RGB}{251,198,150}

\newtcolorbox{myBox}{%
	colback=grayLL,colframe=grayL,top=1mm,bottom=1mm,left=1mm,right=1mm%
}
\newtcolorbox{myTitledBox}[2][]{%
	colback=grayLL,colframe=grayL,top=1mm,bottom=1mm,left=1mm,right=1mm,enlarge top by=0.5em,title={#2},fonttitle=\bfseries\small\color{gray},#1%
}

\newcommand{\mymathbox}[3]{%
    \tcboxmath[top=0mm,bottom=0mm,left=0mm,right=0mm,fonttitle=\bfseries\scriptsize\color{gray},colbacktitle=white,enhanced,attach boxed title to top center={yshift=-1mm},boxed title style={top=0mm,bottom=0mm,left=0mm,right=0mm},colframe=#1,colback=white,title=#2]{#3}
}

\newcommand{\samplingdenseqn}[3]{%
    \mymathbox{uciBlue}{Visibility}{#1} \times \mymathbox{uciOrange}{FOV Sample Variation}{#2} \times \mymathbox{greenDD}{Distance Decay}{#3}%
}
\newcommand{\samplingdenseqnsimp}[2]{%
    \mymathbox{uciBlue}{Visibility}{#1} \times \mymathbox{greenDD}{Distance Decay}{#2}%
}
\newcommand{\imagewithsquare}[3]{%
        \begin{tikzpicture}
        \node[anchor=south west,inner sep=0] at (0,0) {\includegraphics[width=\textwidth]{#1}};
        \draw[white,very thick,rounded corners] #2 rectangle #3;
        \end{tikzpicture}}

\newcommand{\imagewithtwosquare}[5]{%
        \begin{tikzpicture}
        \node[anchor=south west,inner sep=0] at (0,0) {\includegraphics[width=\textwidth]{#1}};
        \draw[white,very thick,rounded corners] #2 rectangle #3;
        \draw[white,very thick,rounded corners] #4 rectangle #5;
        \end{tikzpicture}}

\title[\final{Floaters No More: Radiance Field Gradient Scaling for Improved Near-Camera Training}]{\final{Floaters No More: Radiance Field Gradient Scaling for Improved Near-Camera Training}}

\author[J. Philip \& V. Deschaintre]
{\parbox{\textwidth}{\centering Julien Philip\orcid{0000-0003-3125-1614} and Valentin Deschaintre\orcid{0000-0002-6219-3747}
}
        \\
{\parbox{\textwidth}{\centering Adobe Research
       }
}
}

\begin{document}

\teaser{
 \includegraphics[width=\linewidth]{Figures/EG_teaser.ai}
 \centering
  \caption{Rendering of different NeRF implementation depths results, without and with our proposed gradient scaling. Most NeRF representations suffer from background collapse, building density close to the cameras. We propose a simple gradient scaling, which can be easily implemented with any volumetric representation, preventing this build-up. We show examples of DVGOv2~\cite{SunSC22,SunSC22_2}, TensoRF~\cite{Chen2022ECCV}, Instant NGP~\cite{mueller2022instant} and MipNeRF 360~\cite{Barron2022MipNeRF3U} (without $\mathcal{L}_{dist}$) without (Original) and with our gradient scaling.}
\label{fig:teaser}
}

\maketitle
\begin{abstract}
NeRF acquisition typically requires careful choice of near planes for the different cameras or suffers from background collapse, creating floating artifacts on the edges of the captured scene. The key insight of this work is that background collapse is caused by a higher density of samples in regions near cameras. As a result of this sampling \final{imbalance}, near-camera volumes receive significantly more gradients, leading to incorrect density buildup. We propose a gradient scaling approach to counter-balance \final{this sampling imbalance}, removing the need for near planes, while preventing background collapse. Our method can be implemented in a few lines, does not induce any significant overhead, and is compatible with most NeRF implementations. 


\end{abstract}  

\section{Introduction}

Neural Radiance Fields (NeRF) \cite{mildenhall2020nerf} introduced a new way to perform 3D reconstruction and rendering of real captured objects and scenes given a set of multi-view images. The NeRF method is based on differentiable volumetric rendering, guiding the presence or absence of density and radiance at every point of the volume.
This flexible representation and the impressive reconstruction quality inspired a new line of research, improving the original formulation with respect to a large variety of aspects including reconstruction quality, reconstruction speed, rendering speed, model compression or required memory.
Nonetheless, some issues, specifically background collapse and floaters remain present in the reconstruction across most methods, especially for real scenes, pointing towards a more fundamental problem.
In this work, we study and propose both an \final{hypothesis} and a simple solution to the problem of background collapse and near-camera floaters.

Background collapse symptoms are very visible floating artifacts appearing close to the training cameras, mistakenly baking some of the background as foreground density. This problem has been identified by previous work \cite{Barron2022MipNeRF3U} which tackles this issue with an additional term in the loss to force densities to concentrate and be close to a dirac~\cite{Barron2022MipNeRF3U}.
While this term indeed results in reduced background collapse it does not explain it and bakes some priors in the optimization, which might not be suitable for all scenes.
Another --often glossed over-- detail that plays a role in reducing background collapse is the use of a near plane during ray-marching.
It completely prevents gradients from backpropagating towards close camera regions as they are not sampled, but it is scene dependent, has to be manually tuned, and is physically inaccurate. Moreover, for complex capture with varying distances to the subjects, no good scene-wide near plane might exist.
On the contrary, we argue that any NeRF method should be able to trace rays directly from the camera without resulting in background collapse or floaters and that \final{one possible cause for such artifacts is a disproportionate amount of gradient received by near-camera volume elements}.
We thus propose to scale the gradient during backpropagation to compensate for this \final{imbalance} using a very simple approximation.
This scaling allows us to completely remove the need for a near plane while preventing background collapse. 

NeRF-based methods mainly differ in their underlying volumetric data structure, which can be as simple as a Multi-Layer Perceptron \cite{barron2021mipnerf,Barron2022MipNeRF3U,mildenhall2020nerf}, a voxel hash-grid \cite{mueller2022instant}, a tensor decomposition \cite{Chen2022ECCV} or even plain voxels \cite{SunSC22}. We show that regardless of the chosen data structure, the reconstructions benefit from our gradient scaling approach.


To summarize our contributions in this work are three-fold:
\begin{itemize}
    \item Identify \final{a possible} root cause of background collapse as an \final{imbalance} in the gradient received by near-camera regions.
    \item Propose a lightweight gradient scaling solution.
    \item Demonstrate its effectiveness for several methods with widely varying data structures.
\end{itemize}

\section{Related work}
We show that our approach is beneficial to many existing NeRF~\cite{mildenhall2020nerf} related methods. We first cover radiance field-based view-synthesis methods before further discussing the problem of background collapse and near-camera \final{sampling} in the radiance field literature and how it has been addressed so far. For more exhaustive coverage of the fast-paced recent literature, we recommend the recent survey on neural rendering by Tewari et al~\shortcite{Tewari22}.

\paragraph*{NeRF representations}
NeRF \cite{mildenhall2020nerf} introduced a different representation for novel view synthesis. Contrary to mesh-based methods that rely on pre-computed geometry and reprojection \cite{ODD15,InsideOut2016,DeepBlending2018, Riegler2020FVS,Riegler2021SVS,PMGD21} or point-cloud-based methods~\cite{Aliev2020,kopanas2021point,Rakhimov_2022_CVPR,kopanas2022neural}, radiance field jointly optimize a three-dimensional volumetric density field and a six-dimensional radiance field through differentiable ray-marching and volumetric rendering. This approach has been adapted to use a wide variety of data structures to store the underlying fields providing varying quality, compactness, optimization speed, and rendering speed trade-offs.
The original line of work \cite{mildenhall2020nerf,Zhang20arxiv_nerf++} used MLPs to represent the scene and was later extended to prevent aliasing \cite{barron2021mipnerf}, to handle unbounded scenes \cite{Barron2022MipNeRF3U} and to better represent reflections \cite{verbin2022refnerf}.
While these methods provide some of the highest-quality results, they are relatively slow to optimize and very slow to render, often requiring several seconds per frame.
To act upon these limitations, several works reintroduced some locality in the representation to avoid evaluating a big MLP for each point of space.
KiloNeRF \cite{reiser2021kilonerf} proposed to first train a NeRF before reproducing the optimized field with tiny local MLPs, enabling faster rendering.
In a similar spirit, PlenOctrees \cite{yu2021plenoctrees} bake an octree after training the original NeRF. Hedman et al propose to bake~\cite{hedman2021snerg} a pre-trained NeRF to improve rendering speed, however, it does not improve optimization time. Speeding up optimization has been shown to be possible by directly optimizing a grid or voxel-based data structure \cite{yu_and_fridovichkeil2021plenoxels,SunSC22,SunSC22_2}. In this context of fast optimization, compactness has also been improved using tensor factorization \cite{Chen2022ECCV} or hashed-voxels \cite{mueller2022instant} and custom CUDA kernels to enable extremely fast optimization and rendering.

We show that this choice of representation is orthogonal to our contribution and that our gradient scaling can be easily integrated to resolve background collapse by evaluating it on prominent methods with diverse representations.

\paragraph*{Floaters, Background collapse and \final{Sampling}}

Several works observed and proposed solutions to the problem of floaters and background collapse in radiance field-based methods.
Roessle et al. \cite{roessle2022depthpriorsnerf} propose to use depth priors to solve this issue in the context of sparse capture,
while in NeRFShop, Jambon et al. \cite{NeRFshop23} acknowledge the problem of floaters and propose an editing method to remove them.
MipNeRF360 \cite{Barron2022MipNeRF3U} proposes a loss that encourages density to concentrate around a single point along the ray, effectively reducing near-camera, semi-transparent radiance.
This relatively heavy loss was further made more efficient \cite{SunSC22_2}.
NeRF in the Dark \cite{mildenhall2022rawnerf} also proposes a loss on the weight variance to reduce floaters.
FreeNeRF \cite{Yang2023FreeNeRF} discusses this issue in detail, referring to it as "walls" and "floaters", noting that they are present near the camera and thus propose to penalize density near the camera with an occlusion loss.
Using these losses however imposes a prior on the scenes density distribution, which might not be suitable for all content. Further, these methods do not explain the root cause of the phenomenon and why the density builds up close to the camera and not somewhere else.
In this work, we provide a \final{possible} explanation and solution to the problem of background collapse and floaters caused by it, by noting that near-camera regions are over-sampled and thus receive more intense and more frequent gradients.
A similar sampling problem was identified by Nimier-David et al.~\shortcite{nimierdavid2022unbiased} in the context of volumetric effect optimization (e.g. smoke) where sampling proportional to the product of transmittance and density leads to patches of density buildup, while only sampling proportional to transmittance significantly reduces artifacts.
In our work, we do not modify the sampling but account for its \final{imbalance} in the backpropagation step.

\section{\final{Analysis}}
\final{In the following section we briefly review radiance fields optimization (Section~\ref{sec:nerfoptim}), we define the problem (Section~\ref{sec:problematic}) and present our hypothesis for the cause of this issue (Section~\ref{sec:cause}).}
\subsection{Neural Radiance Fields optimization}
\label{sec:nerfoptim}
Most NeRF methods share common components and assumptions to optimize their volumetric representation. Starting from a set of images captured from calibrated cameras, the goal is to optimize an emissive volumetric density to reproduce the appearance of the input pictures when rendered with piece-wise constant volume rendering.
The general optimization framework selects pixels in the input training images, generates a ray starting at the camera and towards the chosen pixel, and performs ray-marching by sampling the data structure at discrete positions along the ray to obtain colors and density. The colors and density of these samples are  integrated to get a color for each ray cast through that pixel. The aggregation of these colors is finally compared with the original pixel value, resulting in a loss for optimization using stochastic gradient descent.

\subsection{Problem Statement}
\label{sec:problematic}
Given only a set of calibrated input images, the NeRF reconstruction problem is ill-posed and naive solutions exist. For instance, a planar surface close to each camera, reproducing their image content, could lead to reconstruction loss of 0. 
In practice, the nature of the data structures and loss regularizers partially prevent this from happening. However, some artifacts often remain: two of the most prominent are floaters and a phenomenon referred to as \emph{background collapse}, where some geometry is reconstructed near the camera.
In turn, this incorrectly reconstructed geometry is seen, from other viewpoints, as floating geometries.
\final{Note that while background collapse results in floaters, some floaters might have a different origin that our method does not address.}

\begin{figure}[t]
    \centering
    \begin{tabular}{@{}c@{\,}|@{\,}c@{}}
    \includegraphics[width=0.49\linewidth]{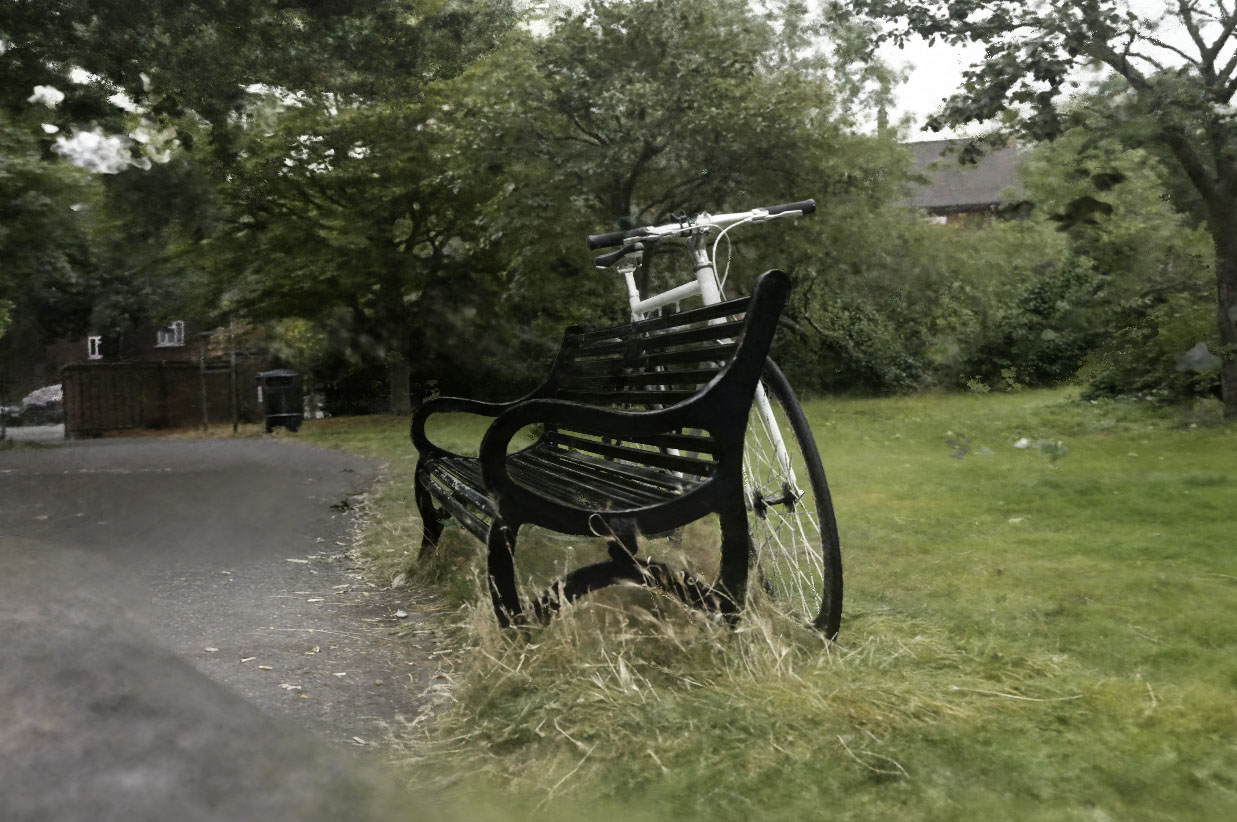} & 
    \includegraphics[width=0.49\linewidth]{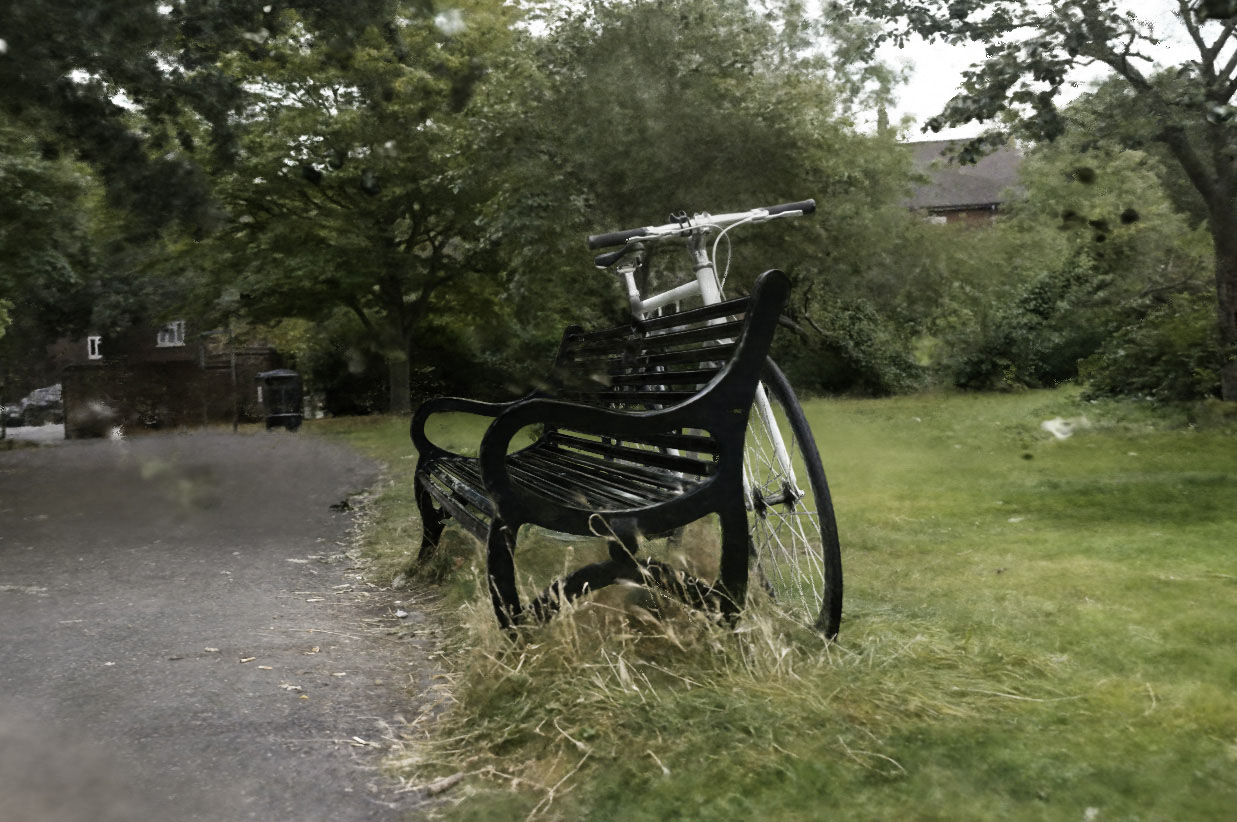} \\
    \includegraphics[width=0.49\linewidth]{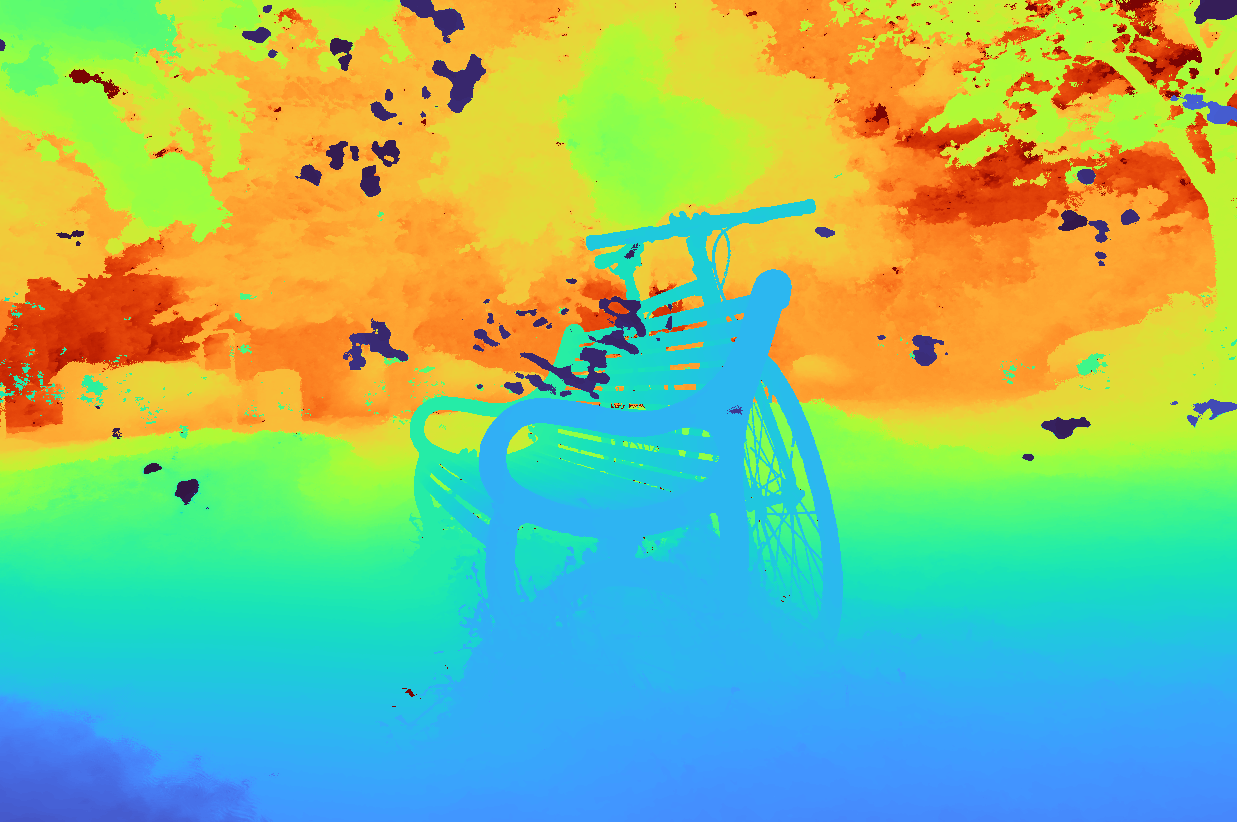} & 
    \includegraphics[width=0.49\linewidth]{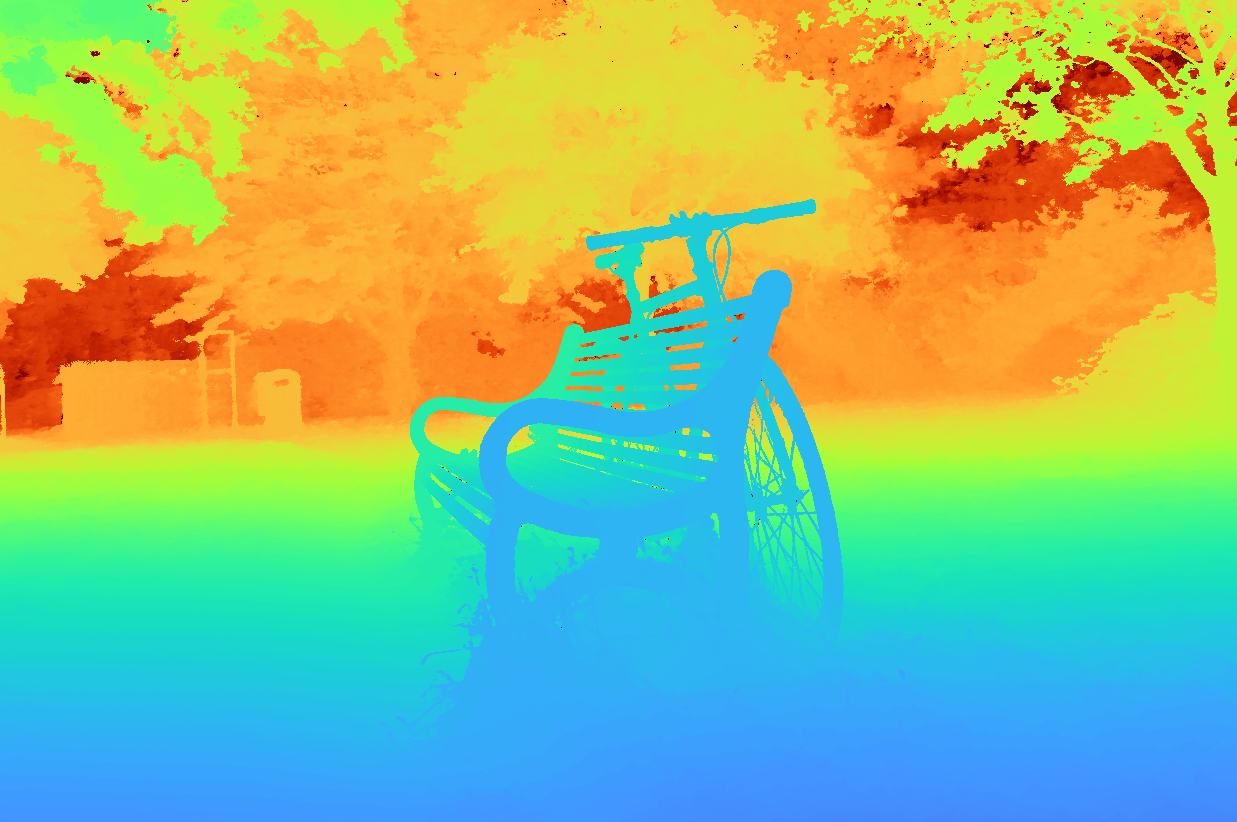} \\
    \scriptsize (a) no $\mathcal{L}_{dist}$ 
    & \scriptsize (c) with $\mathcal{L}_{dist}$
    \end{tabular}
    \vspace{-0.05in}
    \caption{Illustration of the problem of background collapse. $\mathcal{L}_{dist}$ is a loss proposed in MipNeRF360~\cite{Barron2022MipNeRF3U} to prevent it.}
        \label{fig:mip360}
\end{figure}

In Fig.~\ref{fig:mip360} we visualize depth maps in false colors, going from dark blue (close) to red (far). We can see dark blue regions in the depth maps without regularizer loss (a), indicating that some geometry is reconstructed near the camera.
In MipNeRF360 the authors propose a relatively complex loss $\mathcal{L}_{\mathrm{dist}}$ to partially solve both floaters and background collapse -- These issues are linked as the close-camera density appears as floating geometry from other viewpoints. For a given ray, the loss \final{ aims at consolidating sample weights into as small a region
as possible}

While this loss partially prevents background collapse, it does not explain it. Further, it pushes density to be concentrated, which can be a problem if semi-transparent surfaces appear in the scene as they are represented with partial density.

Another simple way to mitigate background collapse is to set a near plane for the rays superior to zero, i.e. the rays do not start from the camera center but a certain distance from it.
In practice, this trick is used in most NeRF-related work, but rarely discussed, nor its implications.
Using a non-zero near plane prevents any gradient to influence the pyramid formed by the camera center and the near plane when optimizing for a pixel.
On the other hand, it prevents reconstruction and rendering in this pyramid, meaning that one should carefully pick the near plane distance. 
If it is too close to the camera, background collapse may arise, if it is too far, some geometry would be missed during reconstruction and might lead to other artifacts. Indeed, with a near plane positioned too far, the model has to represent the color of the training pixel with density after the near plane.
While reasonable values can often be found for the near plane distance, it requires per-scene tuning in the general case. In the case of capture from varying distances from the main subject(s), this near-plane distance may need to be set for every camera independently. Further, when some content is captured very close to the camera, no good value might exist.


\begin{figure}
     \centering
         \centering
         \includegraphics[width=\columnwidth]{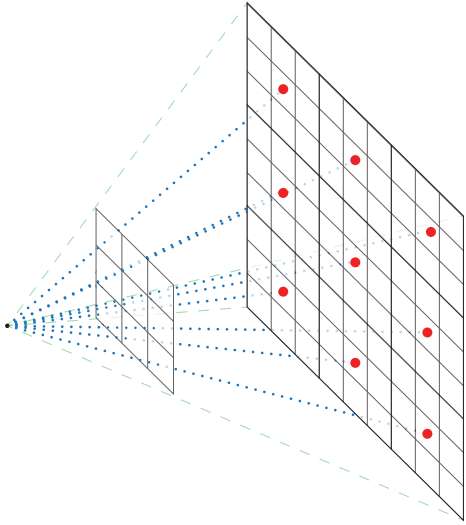}
     
        \caption{As rays spread from the camera toward the scene, the density of points sampled decreases: all areas were sampled in the first rectangle, while only $\frac{1}{9}$ were in the second (marked with red circles)}
        \label{fig:Spread}
\end{figure}

\subsection{Cause}
\label{sec:cause}
We \final{hypothesize} that background collapse is primarily caused by a \final{disproportionate} amount of gradient near-camera volumes receive.
As illustrated in Fig.~\ref{fig:Spread}, ray casting from a camera is akin to the propagation of light and suffers from a similar quadratic decay.
Given a camera and a visible volume element and assuming equally spaced samples along the ray, the density of samples falling in the volume element is proportional to the inverse of the square of the distance from that camera.
This means that the volume close to the camera is disproportionately more sampled than the rest of it and that near-camera regions receive significantly more gradients per volume element, encouraging a fast build-up of density, and creating floating artifacts.

Indeed, as data structures used to represent radiance fields are generally continuous, a higher sampling rate of volume elements directly translates to stronger and more frequent gradients for the variables used to represent the density and color of the volume. For instance, in the case of a semi-discretized representation, which uses a voxel-like structure \cite{yu_and_fridovichkeil2021plenoxels,mueller2022instant, SunSC22, SunSC22_2}, weights have a local arrangement. In Direct Voxel Grid Optimization (DVGO) \cite{SunSC22} \final{and follow-up work (DVGOv2) \cite{SunSC22_2}} only the eight weights associated with the corners of the voxel containing a sampled point are affected by the backward pass. In these cases, a higher density of sampling directly translates to gradients received more often, and therefore faster updates. This same reasoning can also be applied to the different levels of hash grids in NGP \cite{mueller2022instant}. In the case of MLP-like implicit representations~\cite{mildenhall2020nerf,barron2021mipnerf,Barron2022MipNeRF3U}, this  \final{higher sampling rate} translates to the MLP receiving a lot more signal for near-camera space than elsewhere.

We also note that this sampling \final{imbalance} has the strongest effect early in the training when the low frequencies are not fitted yet.
At this early training stage, the gradients are likely to be locally very aligned as they all push toward the same global direction.
For instance, if the colors predicted at early iterations for a small volume are varying around grey but the target is red, all points receive approximately the same gradient to change the color to be redder.
In such cases, this means that the gradient for a weight influencing a volume element scales linearly with the sampling density of this volume \final{and the weight changes faster.}


\begin{figure}
     \begin{subfigure}{\linewidth}
        \centering
        \includegraphics[width=\linewidth]{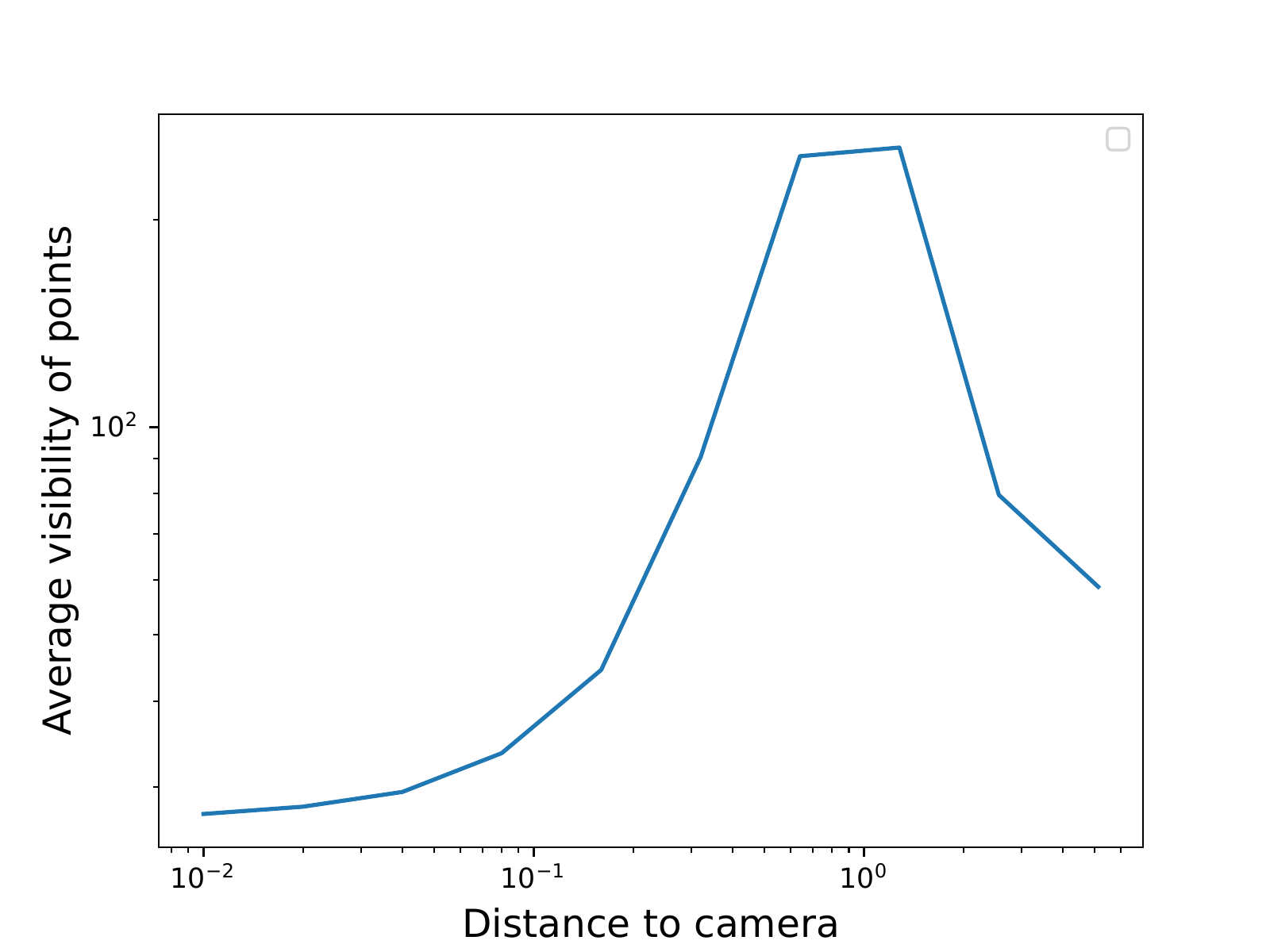}
        \caption{Visualisation of the average number of cameras that see a point on a ray as a function of distance to the ray origin. The score is averaged over all cameras in 12 scenes from various methods \cite{Barron2022MipNeRF3U,Chen2022ECCV,SunSC22,mueller2022instant}. Most points near cameras are only seen by a few cameras. Visibility increases until it reaches the average subject distance before decreasing again.}
    \label{fig:vis_cam}
    \end{subfigure}
    
     \begin{subfigure}{\linewidth}
        \centering
        \includegraphics[width=\linewidth]{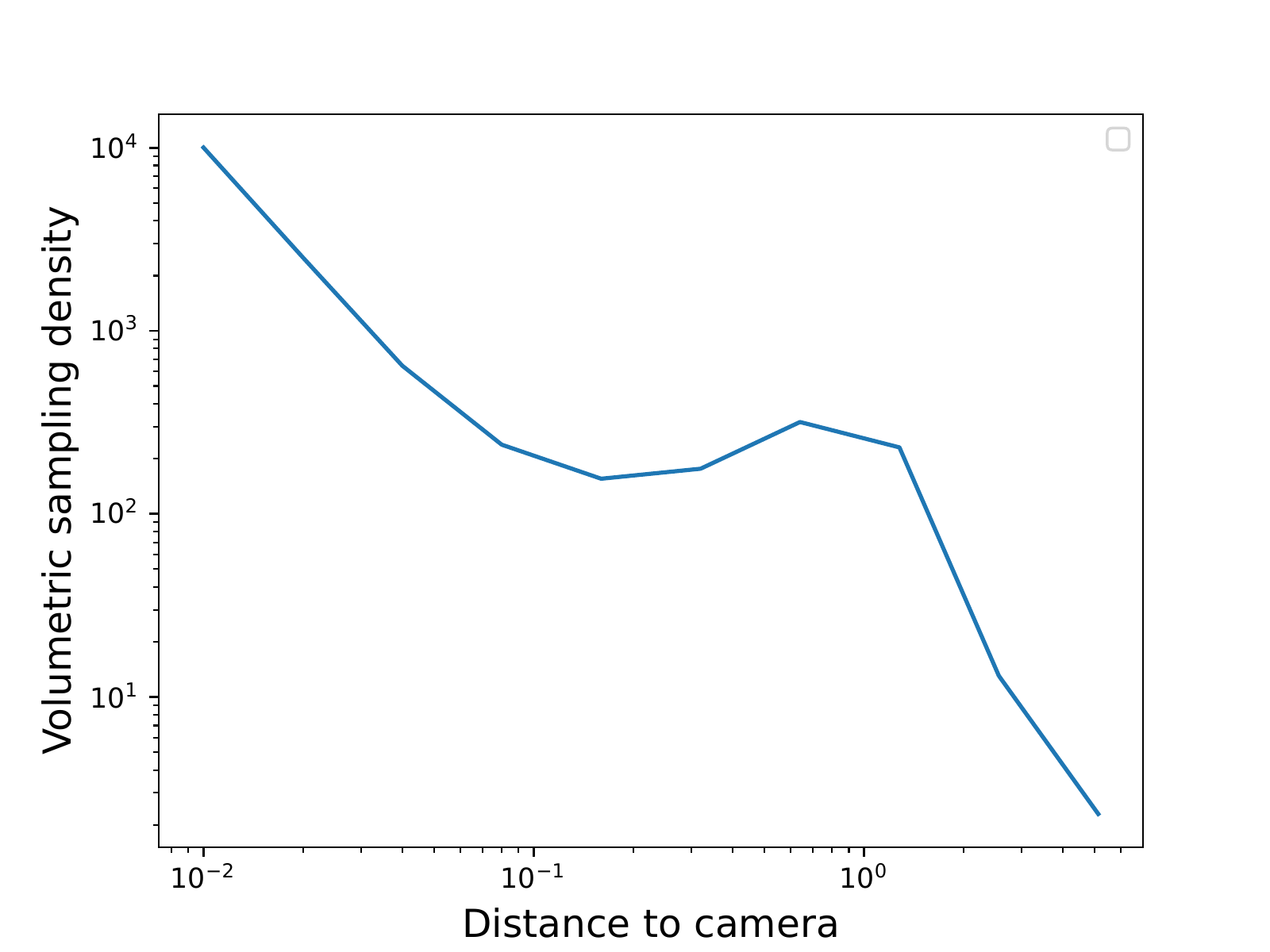}
        \caption{Visualisation of the average volumetric sampling for rays in the Bonsai scene as a function of the distance to the camera. We see that volume units close to the camera are over-sampled despite a low visibility. }
    \label{fig:vis_cam2}
    \end{subfigure}
    \caption{Visibility of points on a ray (a) and average volumetric sampling density (b), with respect to distance from a given camera. Both axes use a logarithmic scale.}
\end{figure}

\begin{figure*}[t!]
\begin{lstlisting}[language=Python]
    class GradientScaler(torch.autograd.Function):
    @staticmethod
    def forward(ctx, colors, sigmas, ray_dist):
        ctx.save_for_backward(ray_dist)
        return colors, sigmas, ray_dist

    @staticmethod
    def backward(ctx, grad_output_colors, grad_output_sigmas, grad_output_ray_dist):
        (ray_dist,) = ctx.saved_tensors
        scaling = torch.square(ray_dist).clamp(0, 1)
        return grad_output_colors * scaling.unsqueeze(-1), grad_output_sigmas * scaling, grad_output_ray_dist
\end{lstlisting}
\caption{PyTorch source code for our gradient scaling operation. \texttt{colors}, \texttt{sigmas}, \texttt{ray\_dist} are respectively the per-point color, density, and distance to the cameras. The forward pass is the identity and the backward scales the gradients during back-propagation based on Eq.~\ref{eq:weighting}. This module has to be called before integrating the points along the rays.}
\label{code}
\end{figure*}

\subsection{Sampling in NeRF}
Given a pinhole camera $c_i$, with a view direction $\vec{d_i}$, rays are sampled uniformly for pixels on the image plane. Along those rays, assuming points are sampled linearly, the sampling density at a given point $p$ is given by:

\begin{equation}
    \label{eqn:samplingdens}
    \rho_i(p)=
    \samplingdenseqn{%
        v_i(p)
    }{%
        \frac{|p-c_i|}{\vec{d_i}\cdot(p-c_i)}
    }{%
        \frac{1}{|p-c_i|^2}\
    }
\end{equation}

Where $v_i(p)$ is a visibility function (1 if $p$ is in the camera field of view and 0 otherwise).
The second term accounts for the lower spatial density of rays on the border while the third accounts for the ray spreading with distance. For reasonable camera FOV, the effect of the second term is negligible in comparison to the distance decay and we can approximate $\frac{|p-c_i|}{\vec{d_i}\cdot(p-c_i)} \approx 1$, giving us:

\begin{equation}
    \label{eqn:samplingdenssimplified}
    \rho_i(p)\approx
    \samplingdenseqnsimp{%
        v_i(p)
    }{%
        \frac{1}{|p-c_i|^2}\
    }
    =v_i(p)\times\frac{1}{(\delta^i_{p})^2}
\end{equation}


With $\delta^i_{p}$ the distance between $c_i$ and $p$.

For a complete scene with $n$ cameras, we can compute the sampling density at a given point $p$ as the sum of density from all cameras:
\begin{equation} \label{eq:rho}
\rho(p)\approx\sum_{i=0}^{n} v_i(p)\times\frac{1}{(\delta^i_{p})^2}
\end{equation}

The main intuition given by the sum in Eq.\ref{eq:rho} is that for a point, visible and close to a given camera, the sum is dominated by this single camera term, while for points further away and at roughly equal distance from the cameras, the visibility term is what plays a significant role.
For points near cameras, the inverse squared distance has a very significant impact, while these points tend to only be visible to a few cameras.
On the other hand, points around the main subject of the capture tend to be visible by a lot more cameras.
This camera visibility phenomenon is illustrated in Fig.~\ref{fig:vis_cam}.
In Fig.~\ref{fig:vis_cam2} we illustrate --on a log scale-- the average sampling density along camera rays. We can see that near cameras, the density is decaying quadratically and that despite lower visibility, the volumes close to the camera are disproportionately densely sampled, leading to a disproportionate amount of gradient for these regions. 

\section{Method}
\label{sec:method}
To compensate for the sampling density \final{imbalance} close to cameras, we propose to scale the gradient that per-point characteristics (such as density or color) back-propagate to the NeRF representation (MLP, voxel grid, etc...) during the backward pass. We propose to apply the following gradient scaling:
\begin{equation} \label{eq:weighting}
s_{\nabla p}=min(1,(\delta^i_p)^2)
\end{equation}
i.e we replace $\nabla p$ by $\nabla p \times s_{\nabla p}$.
Where $\delta^i_p$ is the distance between the point and the camera ray origin.

This scaling compensates for the dominating square density close to the cameras while leaving the rest of the gradients unchanged.
Note that the scaling for a given point depends on the camera from which the rays are cast.

For this proposed approach we assume that the typical distance between the camera and captured content is in the order of 1 unit of distance.
We use this assumption to derive Eq.~\ref{eq:weighting} and did not tune scene scales during our experiments as most scenes respect this assumption. In the case where the scene scale significantly differs from this assumption and the captured content is at an order of $\sigma$ units of distances, the weighting can be replaced by:
\begin{equation} \label{eq:weighting_sig}
s_{\nabla p}=min(1,\frac{(\delta^i_p)^2}{\sigma^2})
\end{equation}

\final{$\sigma$ could potentially be estimated automatically based on camera calibration.}

\final{Acting directly on the gradient is uncommon but designing a loss to achieve a similar effect would be  challenging: the loss would need access to individual density/color of samples as it is not possible to influence individual points differently based on distance after volumetric integration. Regularizing individual densities/colors would impose a prior on their values while scaling gradients reduces the speed at which they change. Further, a sample with 0 density does not contribute to the forward color but may receive a significant gradient. Adding a new loss means adding the gradient of that loss to the gradient of the other losses/regularizers (differentiation sum rule), making it hard to reproduce the effect of the scaling.
In contrast, gradient scaling is straightforward and changes the update magnitude of density/color for samples near the camera, helping to avoid local minima.}

\subsection{Non-linear space parameterization}
\label{sec:method_jacobian}
\final{
Some methods \cite{Barron2022MipNeRF3U} use a non-linear parameterization of coordinates $f(p)\in \mathbb{R}^3\rightarrow\mathbb{R}^3$ to fit an unbounded scene into bounded coordinates.
In such cases, the volume contraction of space should be taken into account to scale the gradients.
This contraction factor is the absolute value of the determinant of the Jacobian $\mathbb{J}_f$ of $f$. The scaling thus becomes:
}
\begin{equation}
\final{
s_{\nabla p}=min(1,\frac{(\delta^i_p)^2}{|\det(\mathbb{J}_f(p)|})
}
\end{equation}
\final{
Depending on the mapping $\det(\mathbb{J}_f(p))$ might be non-trivial to compute. In our experiments, we do not use it, but future work may require it. Indeed in MipNeRF360, the central space of the scene, containing most cameras is unaffected by the mapping and thus the Jacobian is the identity.
}

\subsection{Implementation and performance} \label{implem}
Implementing this operation in PyTorch is straightforward using custom \texttt{autograd.Function} as shown in the 10 lines of code in Figure~\ref{code}. We also provide a JAX implementation in supplemental materials, directly compatible with the multinerf \cite{multinerf2022} codebase. Using this operation, a single call to it can be inserted after the data structure (MLP, hash-grid, voxels, etc) has been sampled, and just before point integration along the ray. This ensures that each point gradient is scaled independently while influencing the weights that control their characteristics (density and color). The code presented in section \ref{implem}, induces an overhead of 100$\mu$s in the backward pass for 300k points, which is negligible in the context of $\sim $50ms iterations. It can readily be used in most codebases with minimal adaptations.\\


\section{Evaluations}
\final{We empirically evaluate our solution for a wide range of volumetric reconstruction methods and representations. While we do not provide a theoretical analysis of the convergence properties of the modified optimization, we find that the proposed scaling reduces floaters for all tested methods while preserving or improving quantitative measures, including optimization loss.}


\begin{figure*}[t!]
     \centering
     \rotatebox{90}{\hspace{13mm}\footnotesize No Scaling}
     \begin{subfigure}[b]{0.19\textwidth}
         \centering
         \footnotesize Depth 500 Steps
         \includegraphics[width=\textwidth]{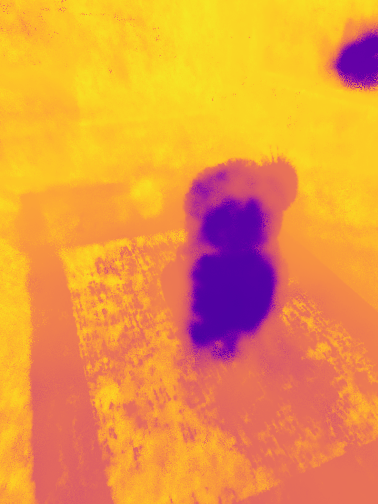}
     \end{subfigure}
     \hfill
     \begin{subfigure}[b]{0.19\textwidth}
         \centering
         \footnotesize Depth 1250 Steps
         \includegraphics[width=\textwidth]{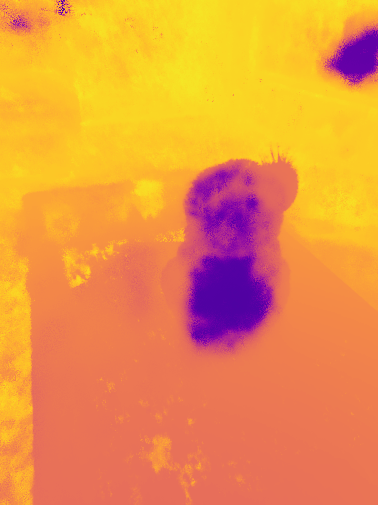}
     \end{subfigure}
     \hfill
     \begin{subfigure}[b]{0.19\textwidth}
         \centering
         \footnotesize Depth 2000 Steps
         \includegraphics[width=\textwidth]{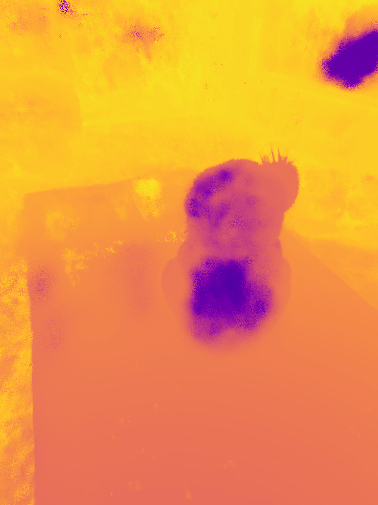}
     \end{subfigure}
     \hfill
     \begin{subfigure}[b]{0.19\textwidth}
         \centering
         \footnotesize Depth 4000 Steps
         \includegraphics[width=\textwidth]{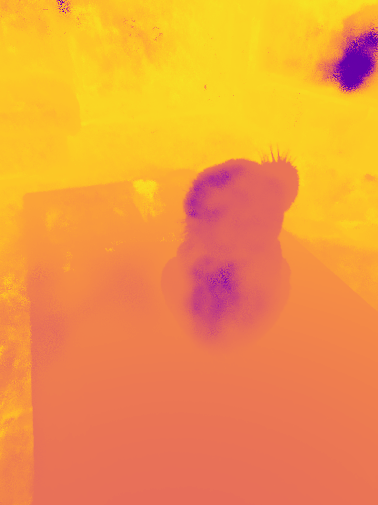}
     \end{subfigure}
     \hfill
     \begin{subfigure}[b]{0.19\textwidth}
         \centering
         \footnotesize Renderings 4000 Steps
         \includegraphics[width=\textwidth]{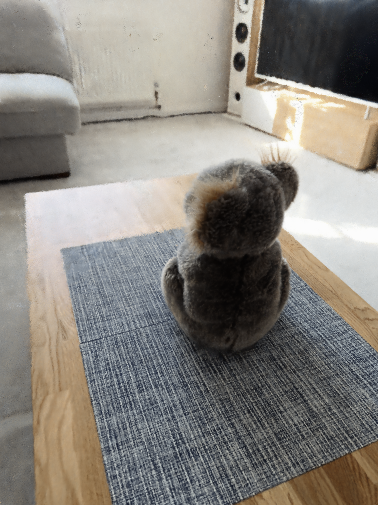}
     \end{subfigure}
     
    \rotatebox{90}{\hspace{10mm}\footnotesize Quadratic Scaling} 
    \begin{subfigure}[b]{0.19\textwidth}
         \centering
         \includegraphics[width=\textwidth]{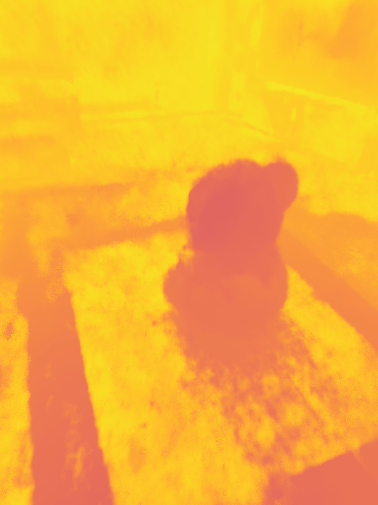}
     \end{subfigure}
     \hfill
     \begin{subfigure}[b]{0.19\textwidth}
         \centering
         \includegraphics[width=\textwidth]{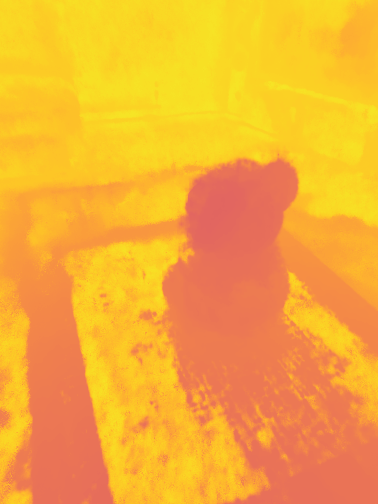}
     \end{subfigure}
     \hfill
     \begin{subfigure}[b]{0.19\textwidth}
         \centering
         \includegraphics[width=\textwidth]{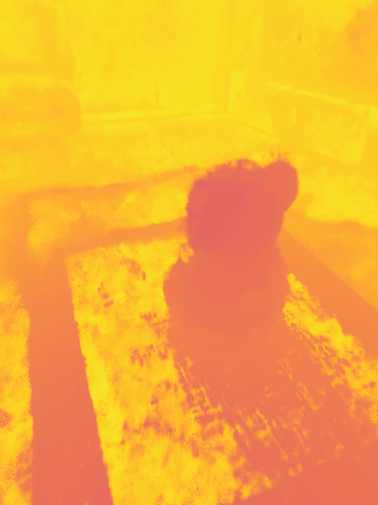}
     \end{subfigure}
     \hfill
     \begin{subfigure}[b]{0.19\textwidth}
         \centering
         \includegraphics[width=\textwidth]{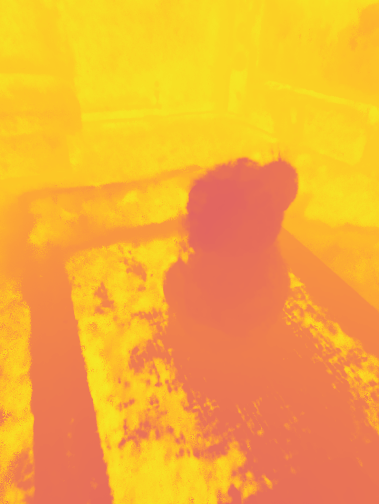}
     \end{subfigure}
     \hfill
     \begin{subfigure}[b]{0.19\textwidth}
         \centering
         \includegraphics[width=\textwidth]{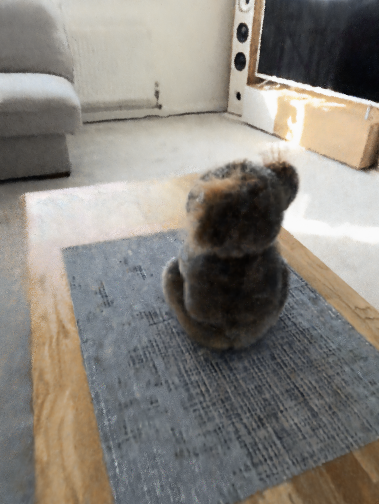}
     \end{subfigure}
     
    \rotatebox{90}{\hspace{7mm}\footnotesize Our Proposed Scaling} 
    \begin{subfigure}[b]{0.19\textwidth}
         \centering
         \includegraphics[width=\textwidth]{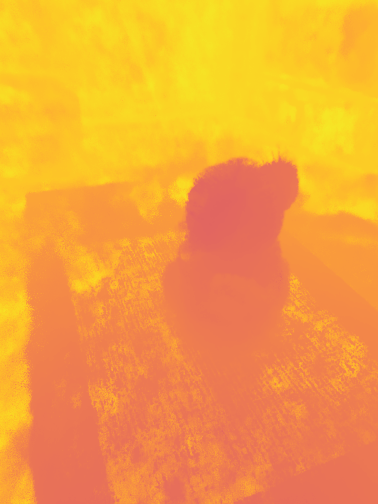}
     \end{subfigure}
     \hfill
     \begin{subfigure}[b]{0.19\textwidth}
         \centering
         \includegraphics[width=\textwidth]{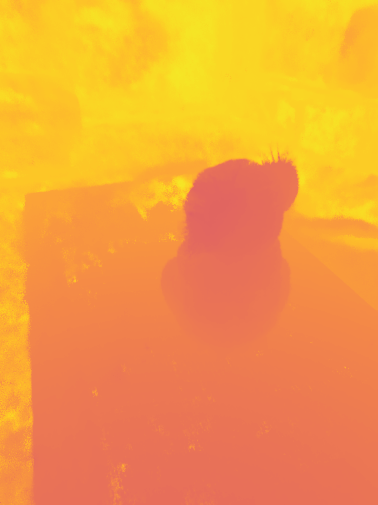}
     \end{subfigure}
     \hfill
     \begin{subfigure}[b]{0.19\textwidth}
         \centering
         \includegraphics[width=\textwidth]{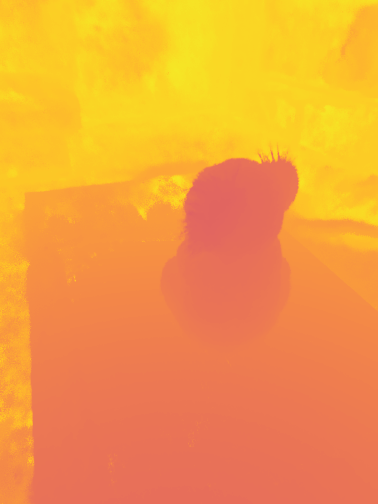}
     \end{subfigure}
     \hfill
     \begin{subfigure}[b]{0.19\textwidth}
         \centering
         \includegraphics[width=\textwidth]{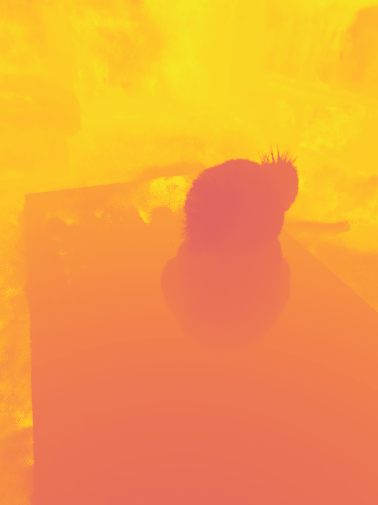}
     \end{subfigure}
     \hfill
     \begin{subfigure}[b]{0.19\textwidth}
         \centering
         \includegraphics[width=\textwidth]{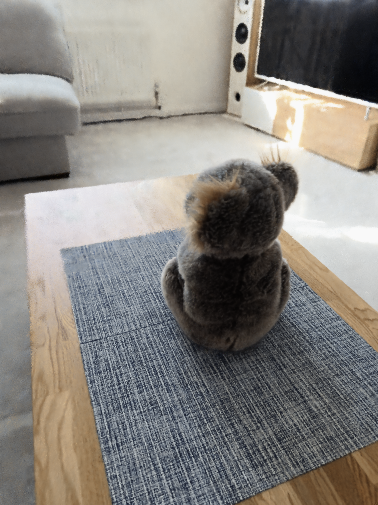}
     \end{subfigure}
     
        \caption{From left to right for each row: Expected Ray depth at 500, 1250, 2000 and 4000 iterations and rendered view at 4000 iterations. Top: No scaling, Middle: Quadratic scaling without clamping, Bottom: our scaling approach. Without Scaling (top row) near camera density arises early on and is never totally removed. With Quadratic Scaling, the table is badly reconstructed with far density and the method only partially recovers. With our Scaling the table is correctly reconstructed early on and no near camera density builds up. The depth is represented here with the plasma color palette, with purple close to the camera and yellow far.}
        \label{fig:comp}
\end{figure*}
\begin{figure*}
     \centering
          \rotatebox{90}{\hspace{1.5mm}\footnotesize No Scaling}
     \begin{subfigure}[b]{0.155\textwidth}
         \centering \footnotesize Depth
         \imagewithsquare{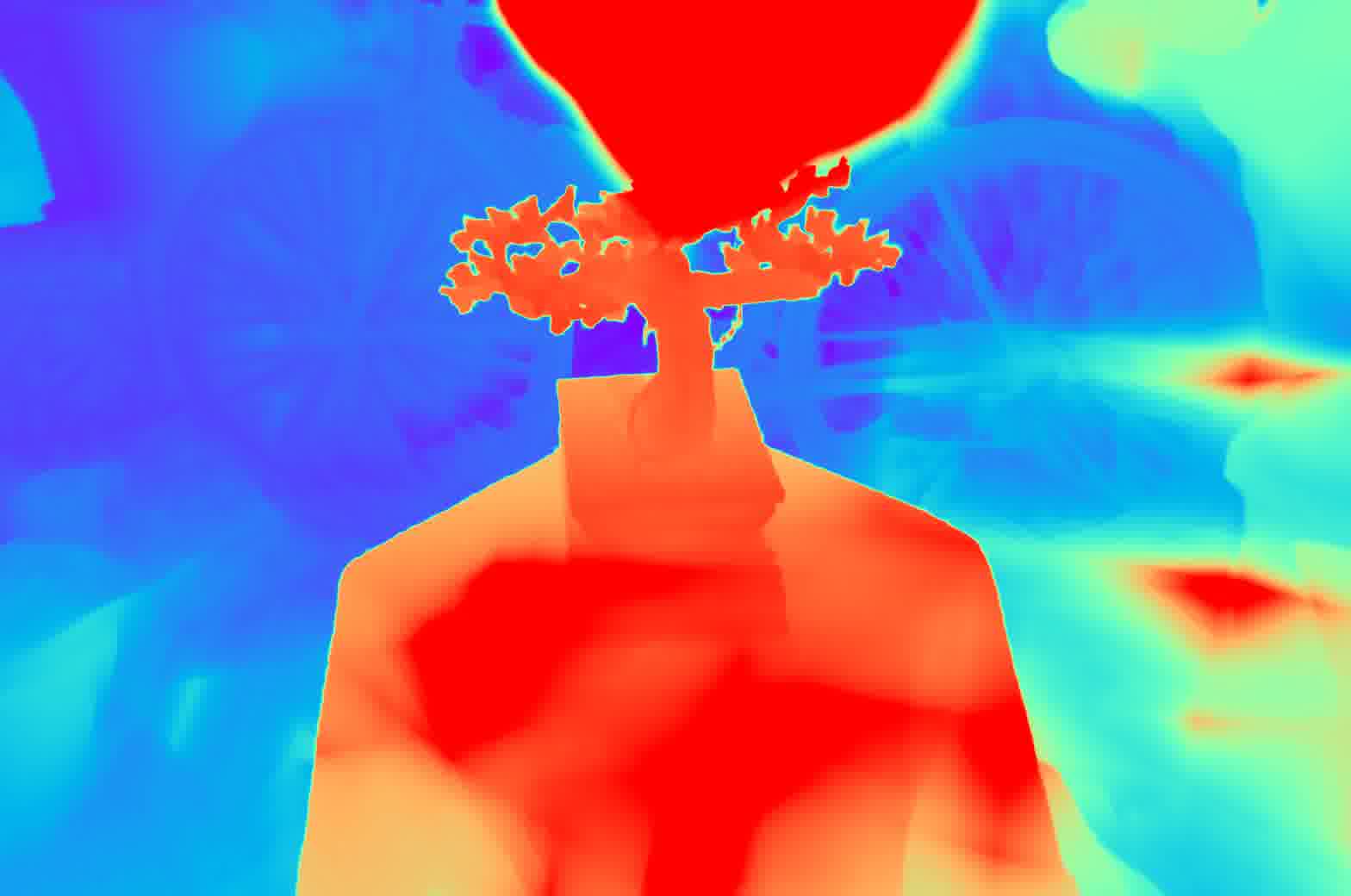}{(0.8, 1.2)}{(1.85, 1.6)}
     \end{subfigure}
     \hfill
     \begin{subfigure}[b]{0.155\textwidth}
         \centering
         \footnotesize Rendering
         \includegraphics[width=\textwidth]{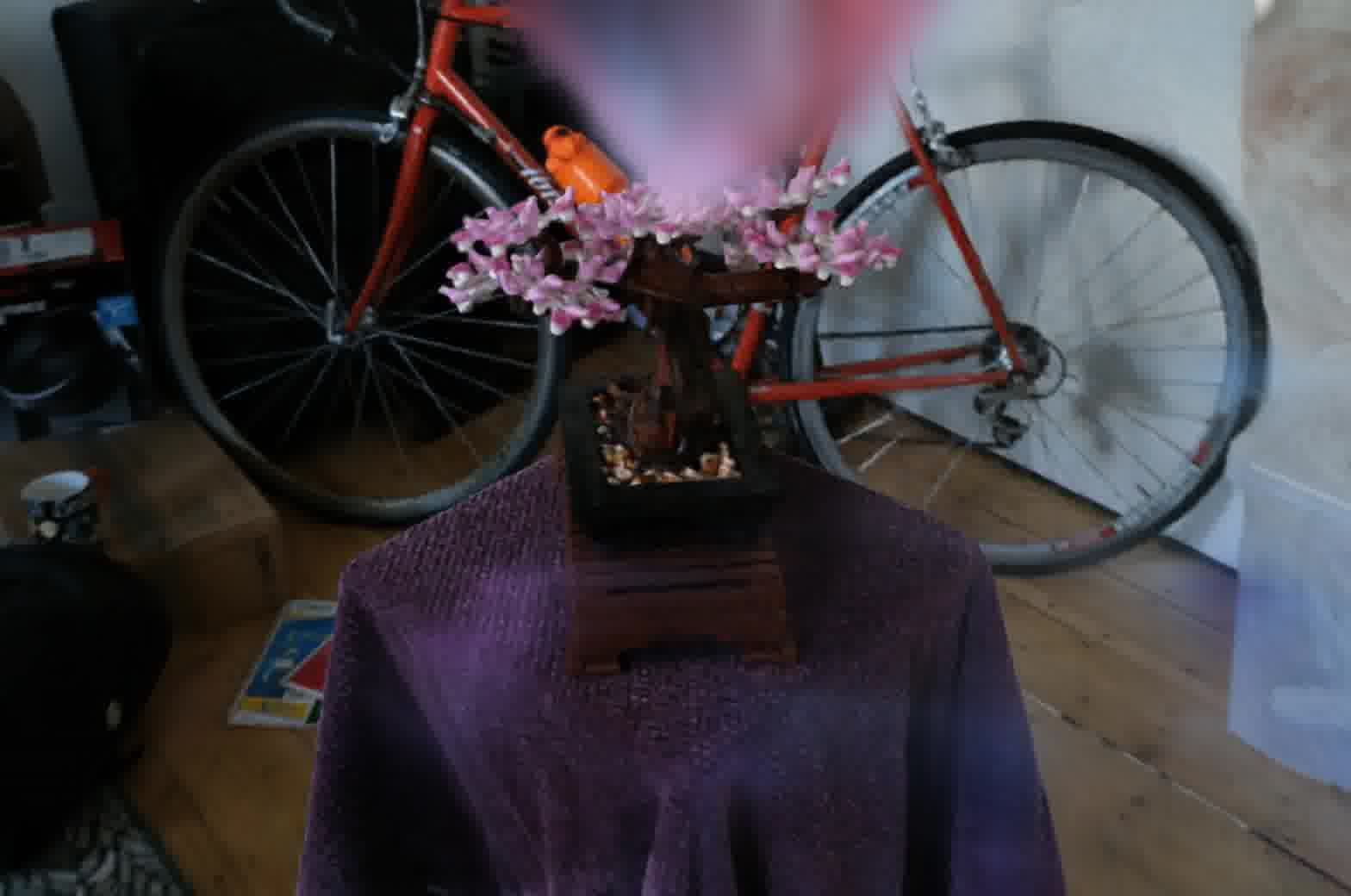}
    \end{subfigure}
     \hfill
     \begin{subfigure}[b]{0.155\textwidth}
     \centering \footnotesize Depth
         \imagewithsquare{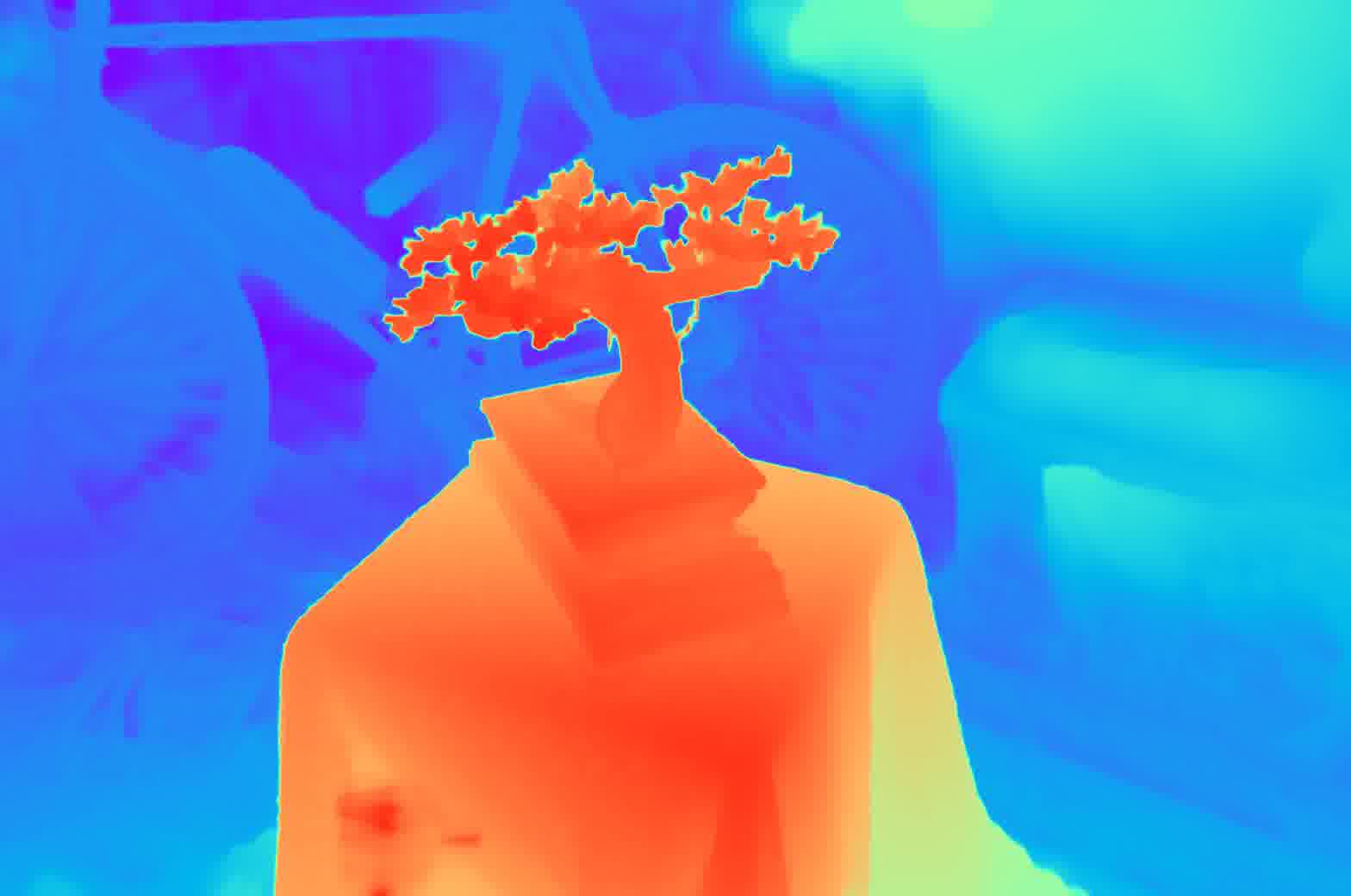}{(0.5, 0)}{(1.2, 0.3)}
     \end{subfigure}
     \hfill
    \begin{subfigure}[b]{0.155\textwidth}
         \centering
         \footnotesize Rendering
         \includegraphics[width=\textwidth]{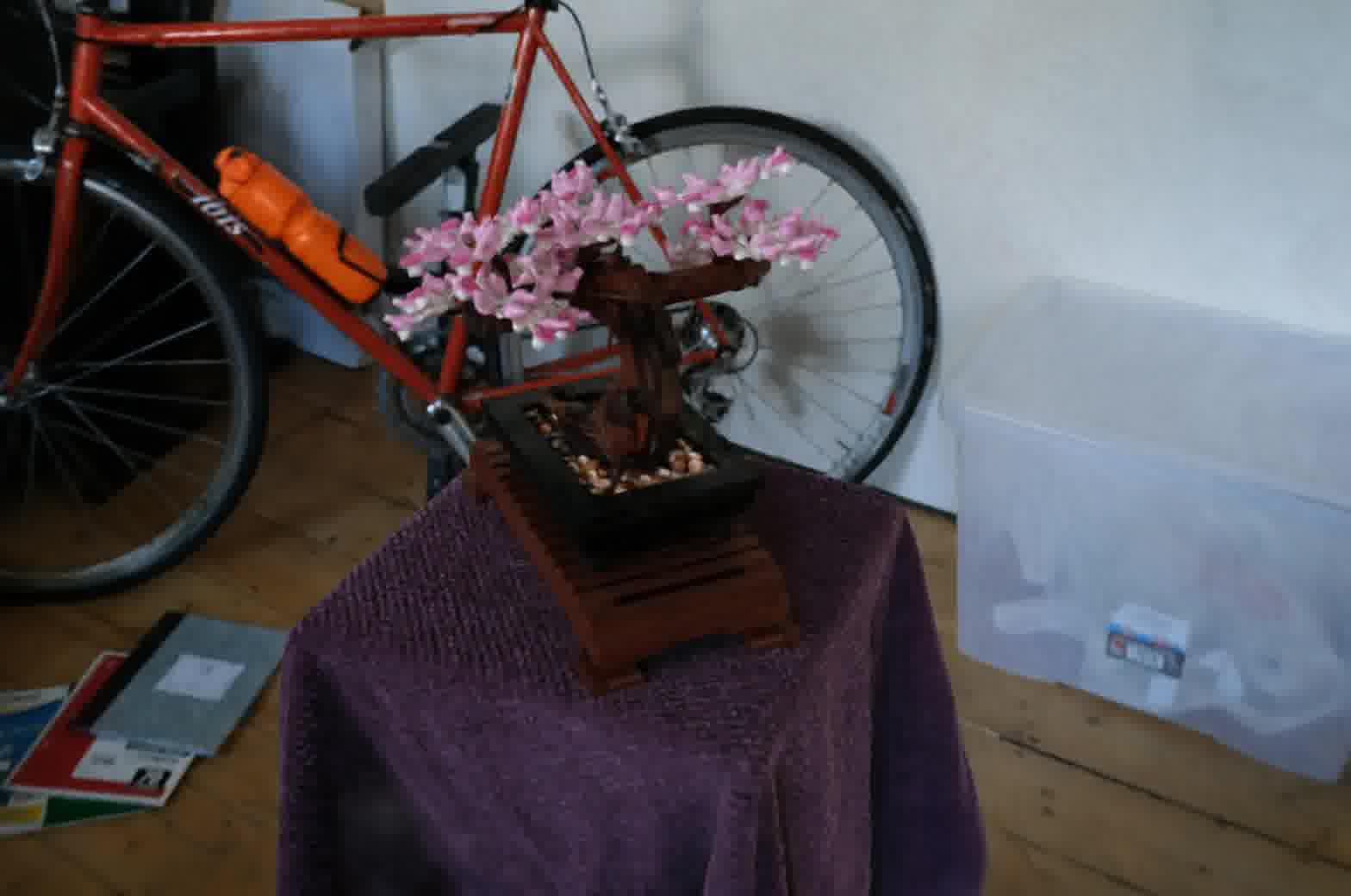}
     \end{subfigure}
    \hfill 
     \begin{subfigure}[b]{0.155\textwidth}
     \centering \footnotesize Depth
         \imagewithsquare{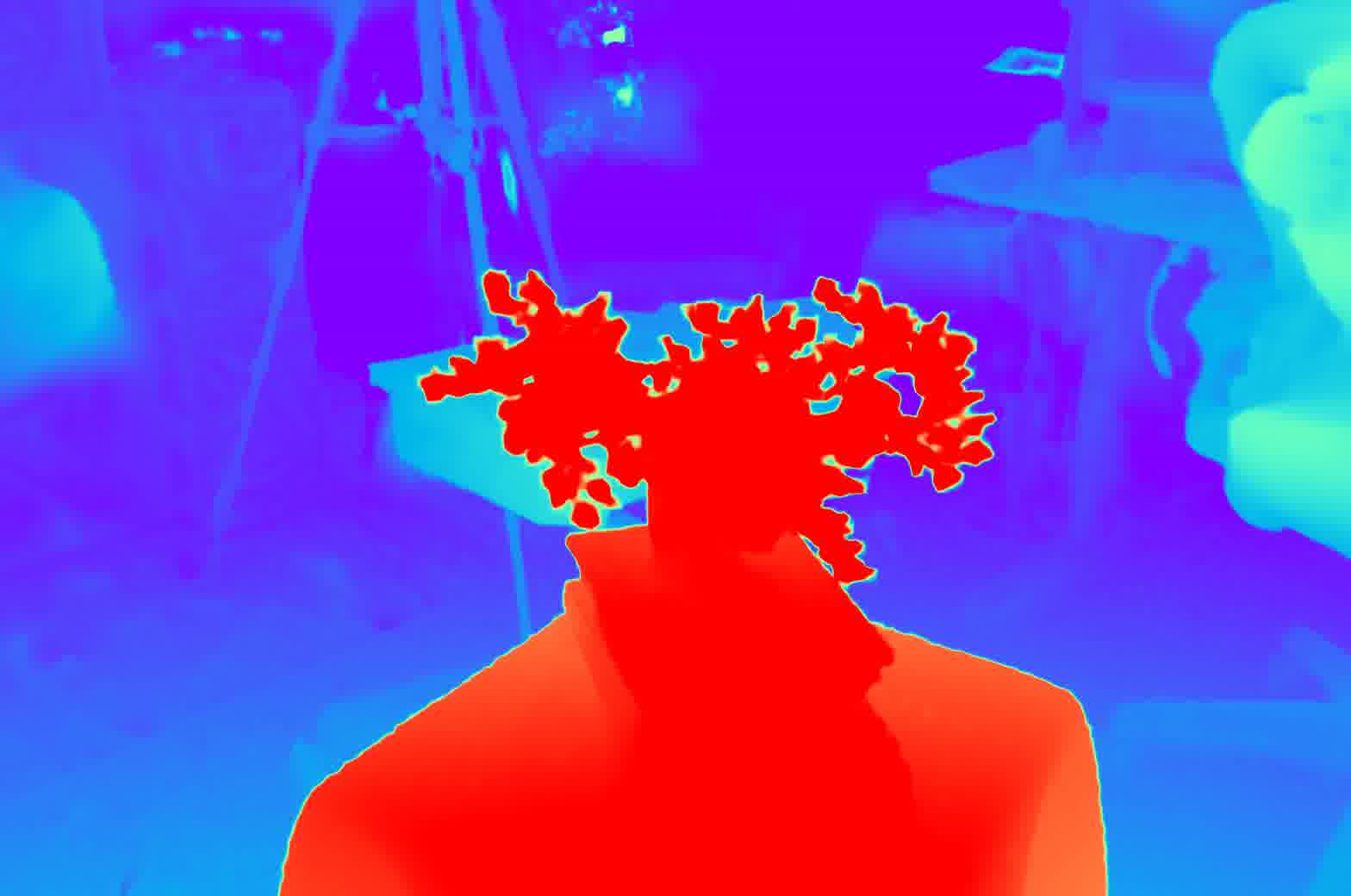}{(0.3, 1.2)}{(1.4, 1.6)}
     \end{subfigure}
     \hfill
    \begin{subfigure}[b]{0.155\textwidth}
         \centering
         \footnotesize Rendering
         \includegraphics[width=\textwidth]{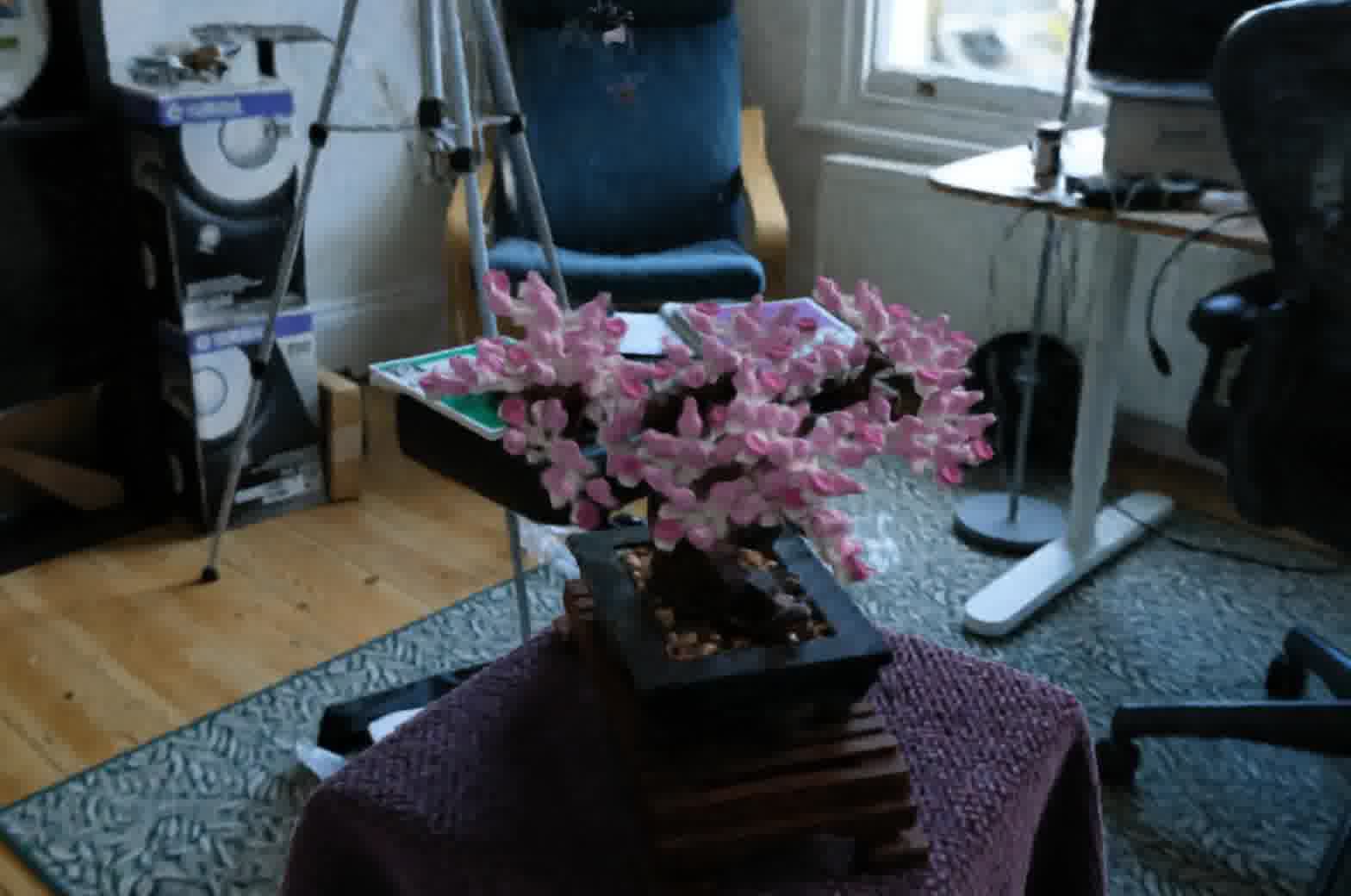}
     \end{subfigure}
     \\
     \rotatebox{90}{\hspace{0.5mm}\footnotesize With Scaling}
     \begin{subfigure}[b]{0.155\textwidth}
         \centering
         \includegraphics[width=\textwidth]{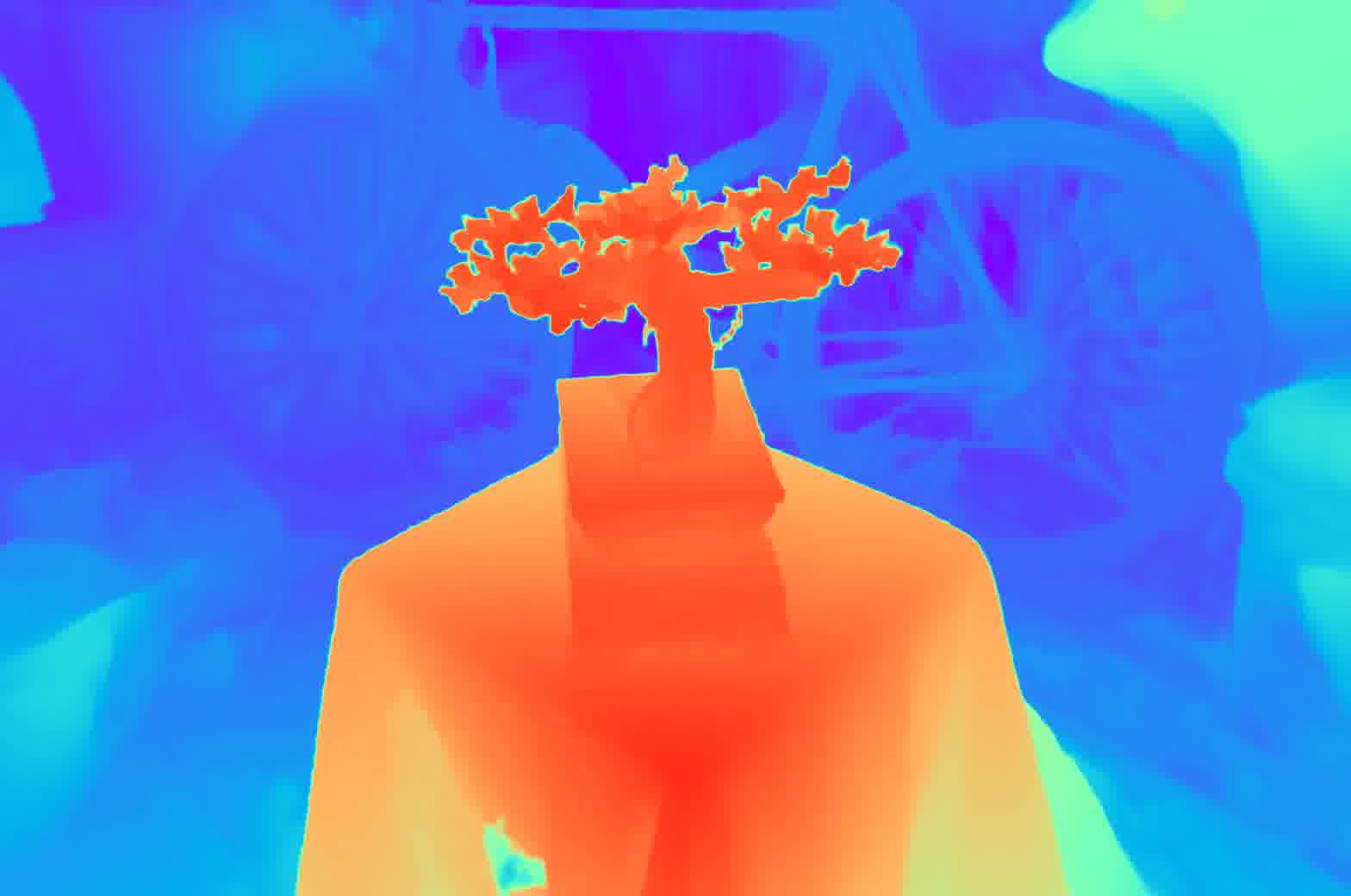}
     \end{subfigure}
     \hfill
     \begin{subfigure}[b]{0.155\textwidth}
         \centering
         \includegraphics[width=\textwidth]{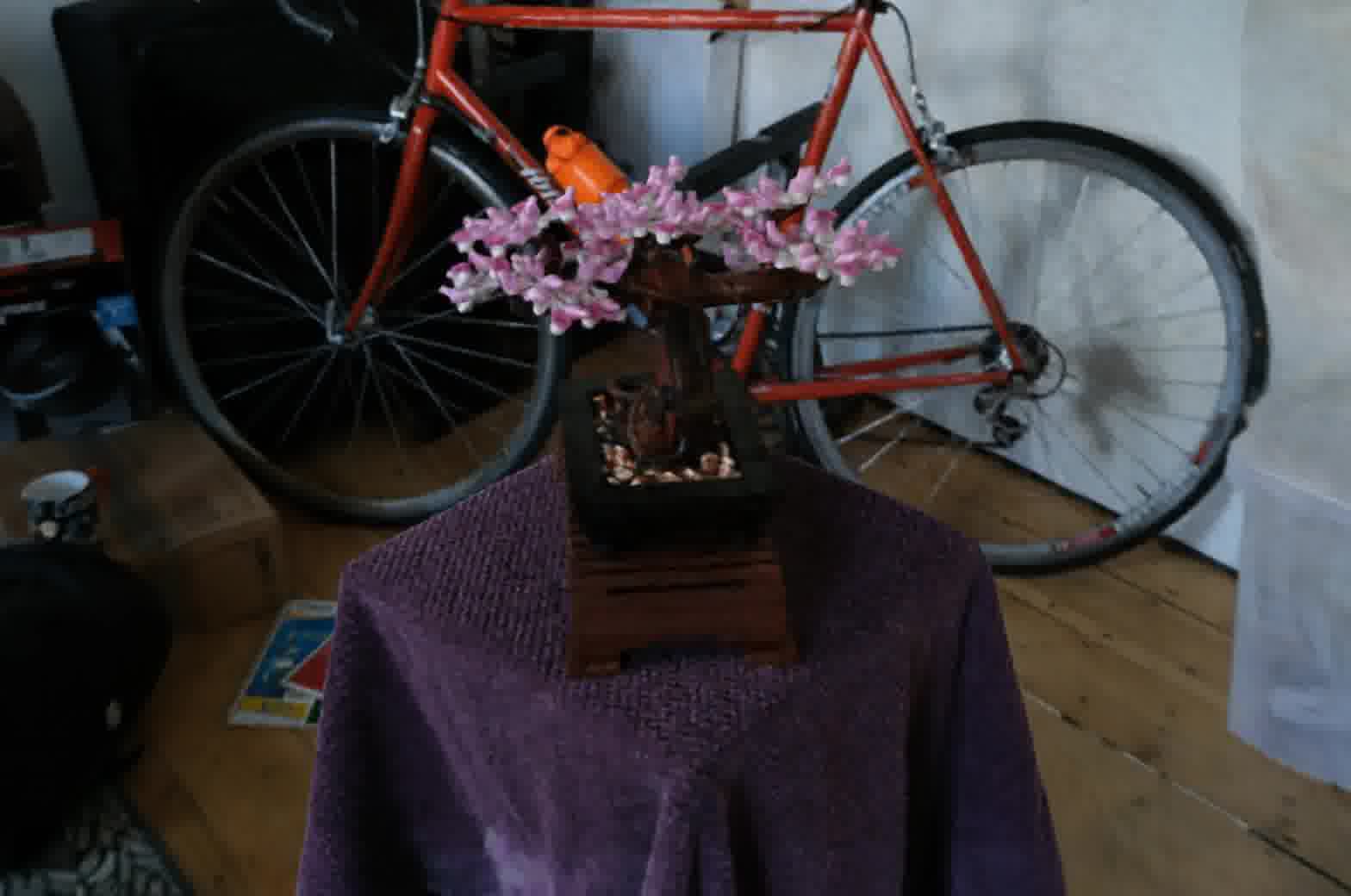}
     \end{subfigure}
     \hfill
     \begin{subfigure}[b]{0.155\textwidth}
         \centering
         \includegraphics[width=\textwidth]{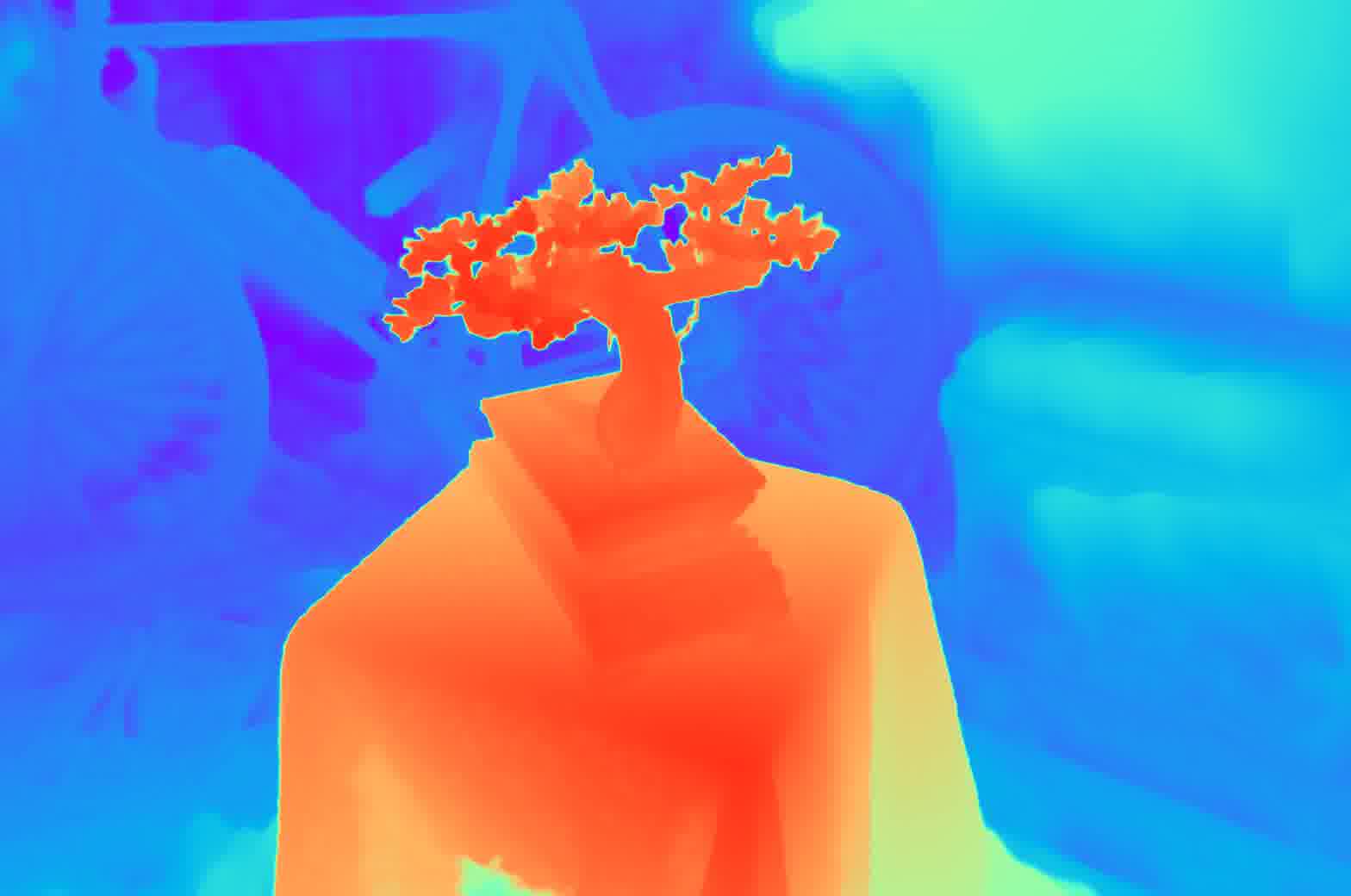}
     \end{subfigure}
     \hfill
    \begin{subfigure}[b]{0.155\textwidth}
         \centering
         \includegraphics[width=\textwidth]{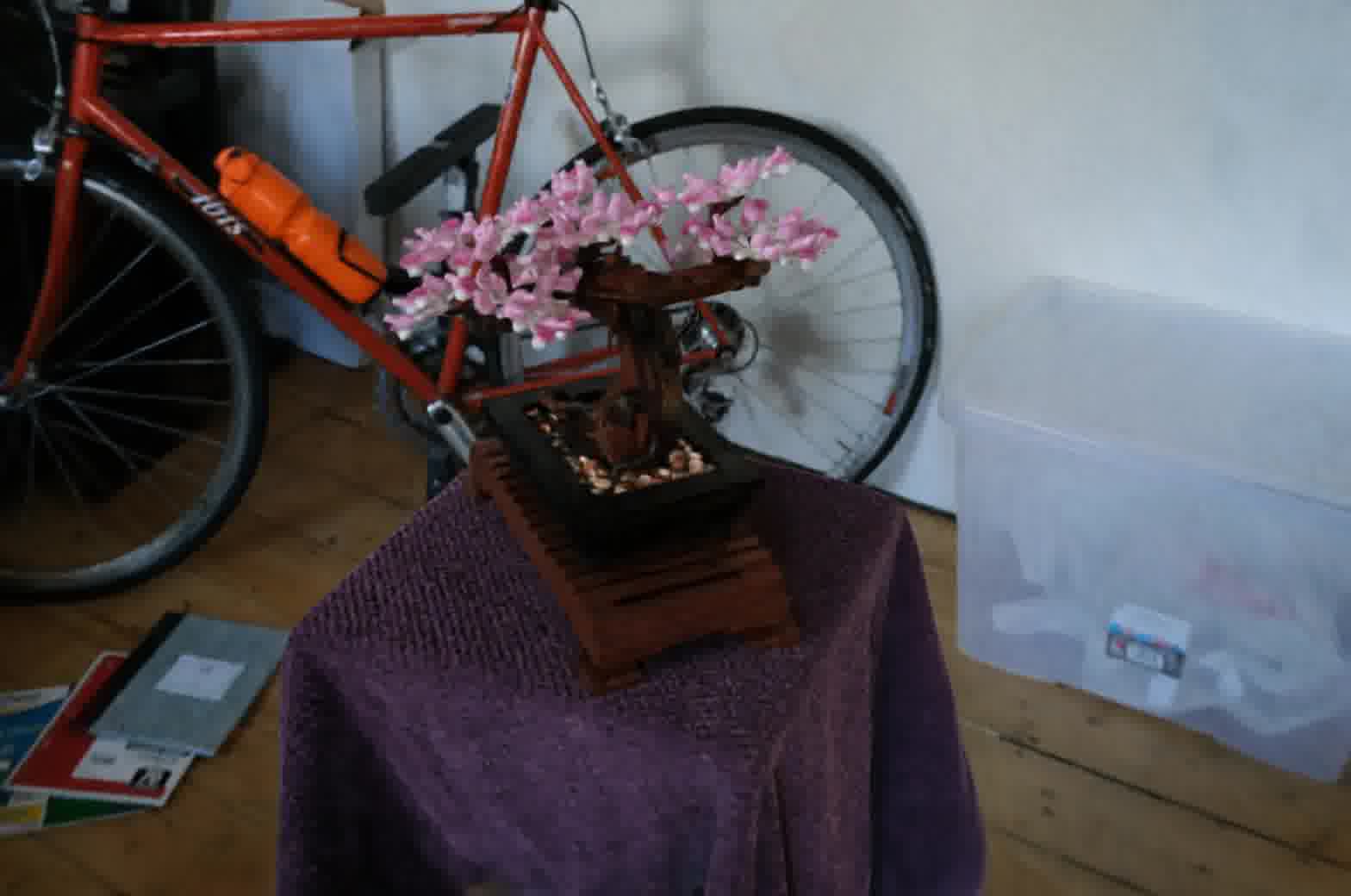}
     \end{subfigure}
    \hfill 
     \begin{subfigure}[b]{0.155\textwidth}
         \centering
         \includegraphics[width=\textwidth]{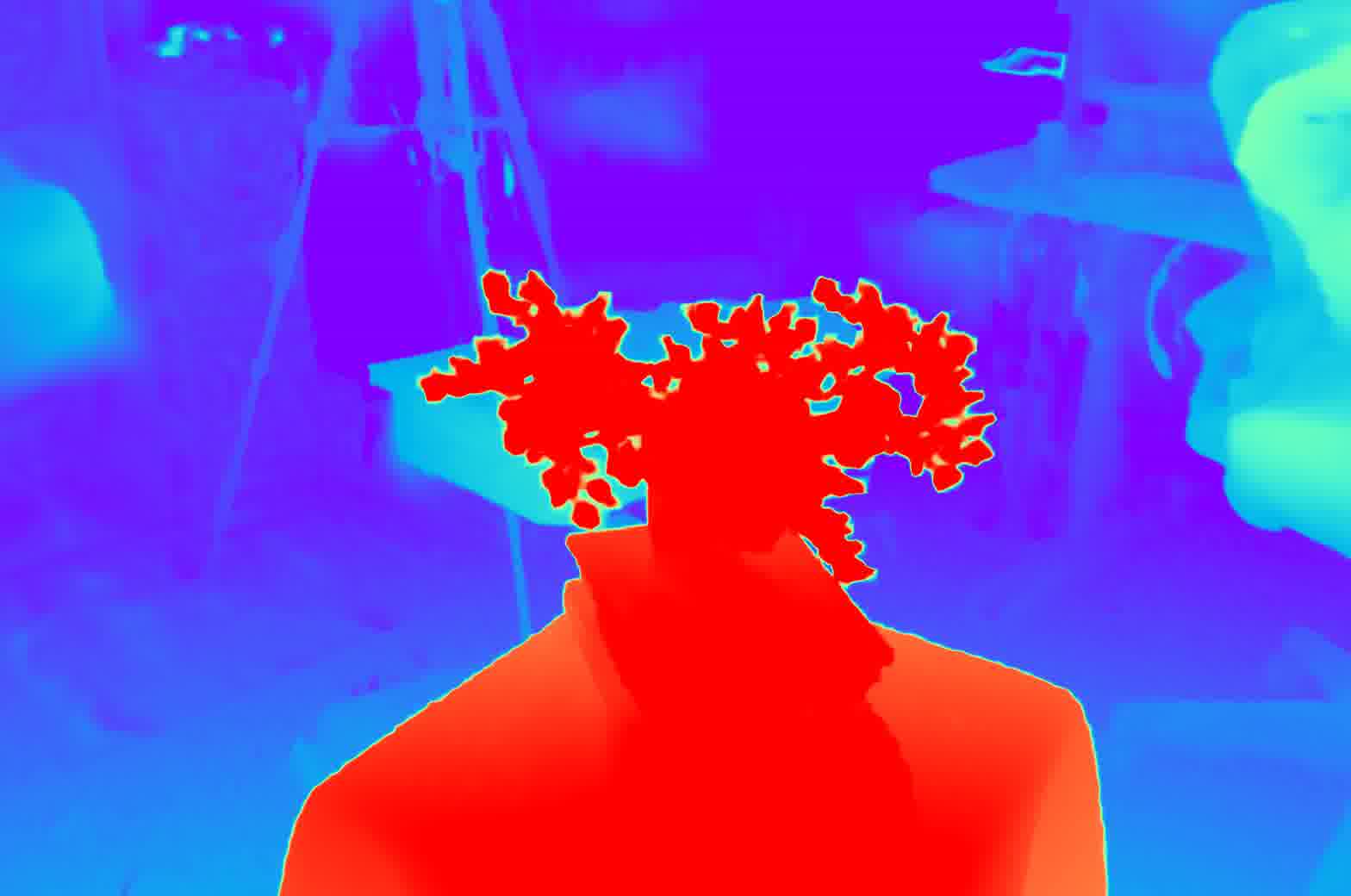}
     \end{subfigure}
     \hfill
    \begin{subfigure}[b]{0.155\textwidth}
         \centering
         \includegraphics[width=\textwidth]{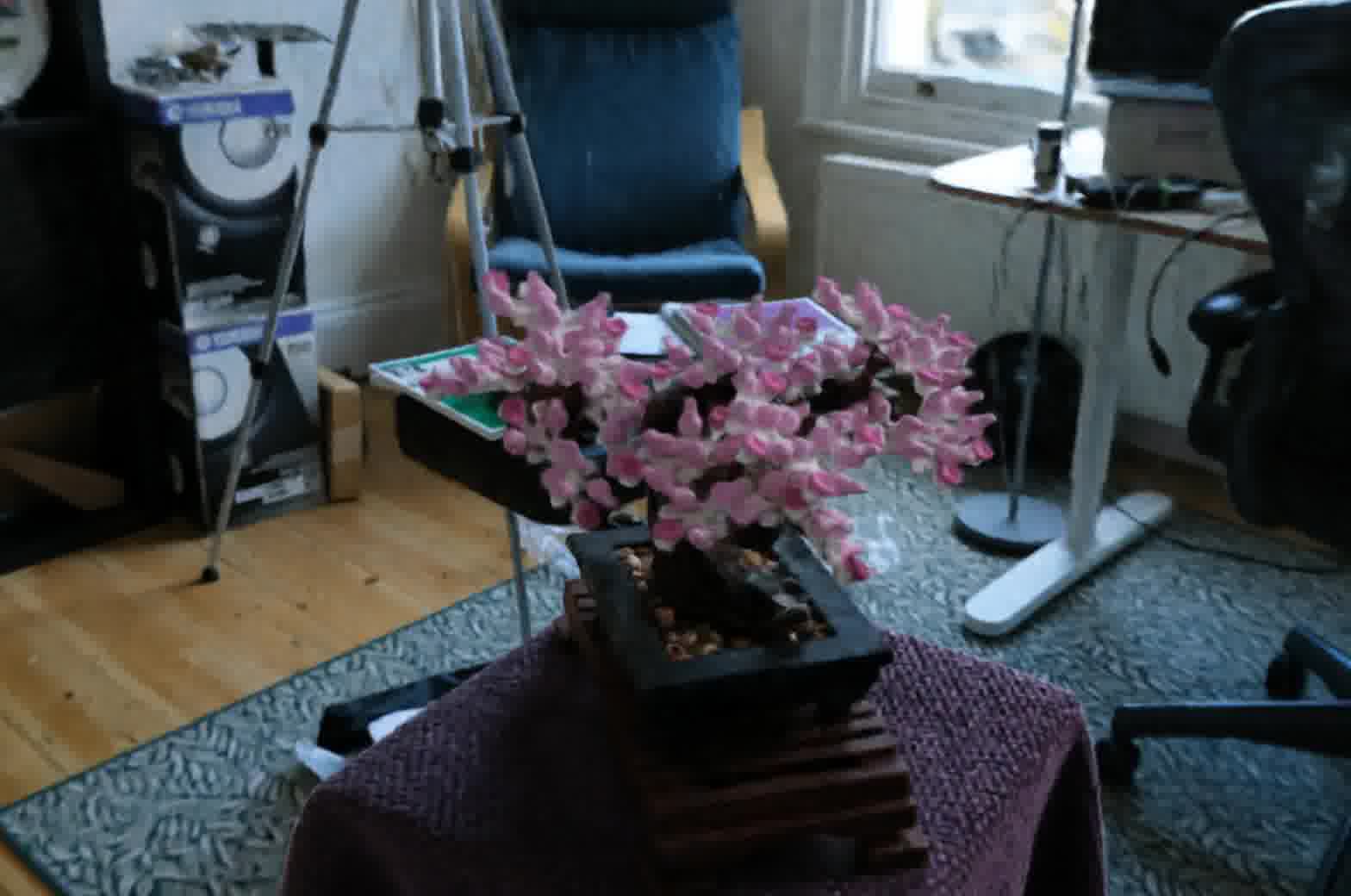}
     \end{subfigure}

          \rotatebox{90}{\hspace{1.5mm} \footnotesize No Scaling}
     \begin{subfigure}[b]{0.155\textwidth}
         \imagewithsquare{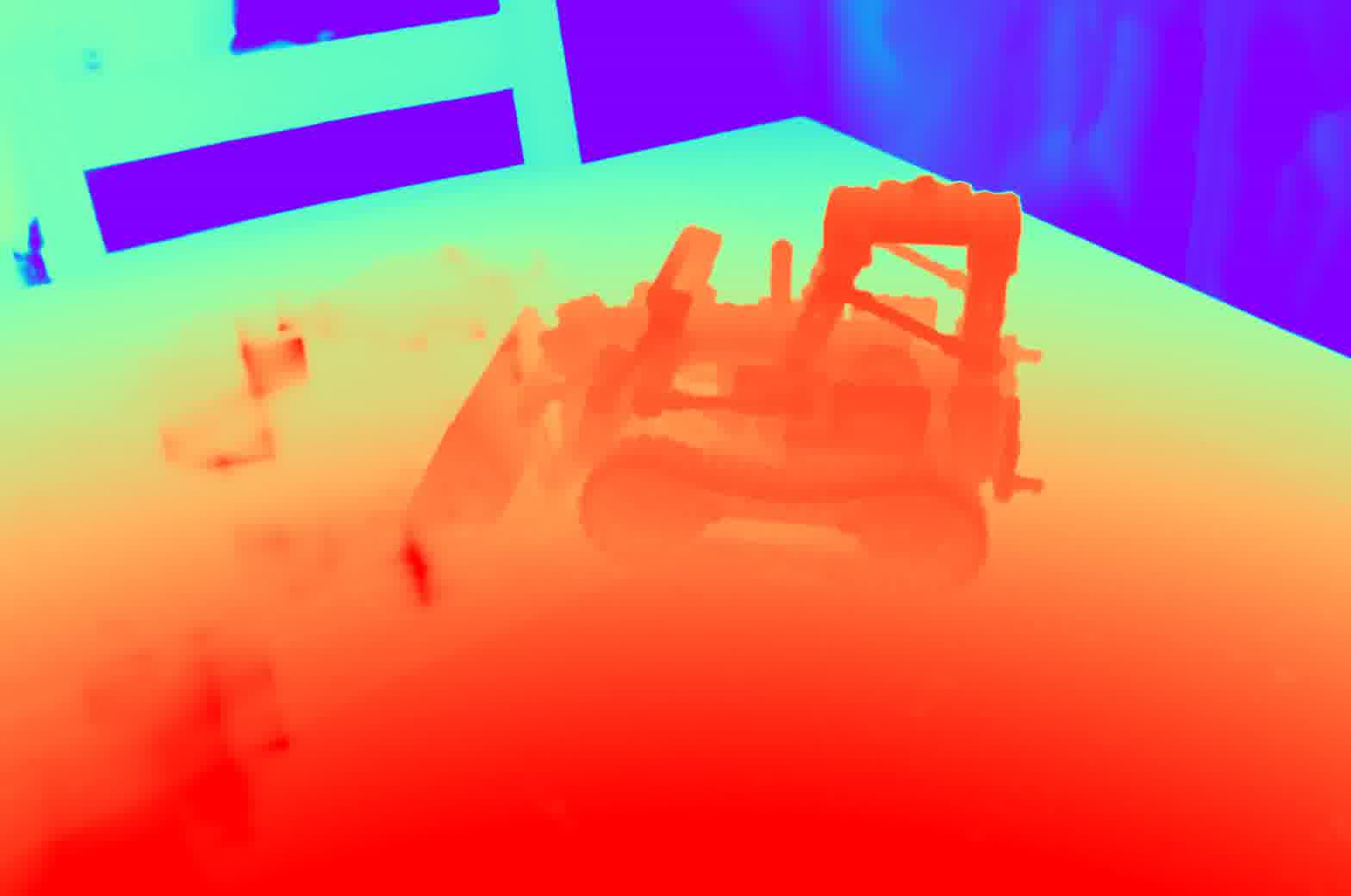}{(0.1, 0.1)}{(1.2, 1.2)}
     \end{subfigure}
     \hfill
     \begin{subfigure}[b]{0.155\textwidth}
         \centering
         \includegraphics[width=\textwidth]{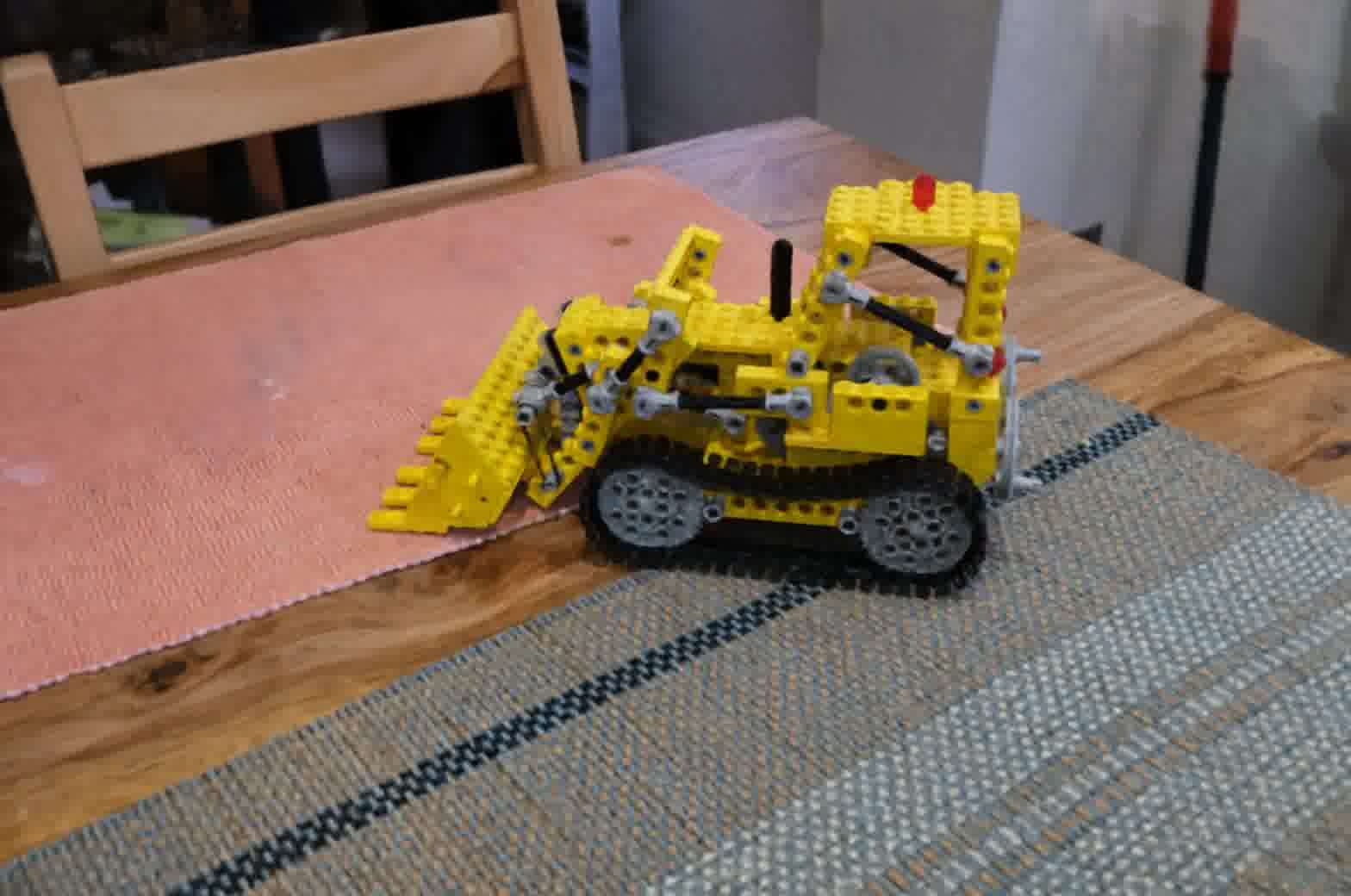}
     \end{subfigure}
     \hfill
     \begin{subfigure}[b]{0.155\textwidth}
         \imagewithtwosquare{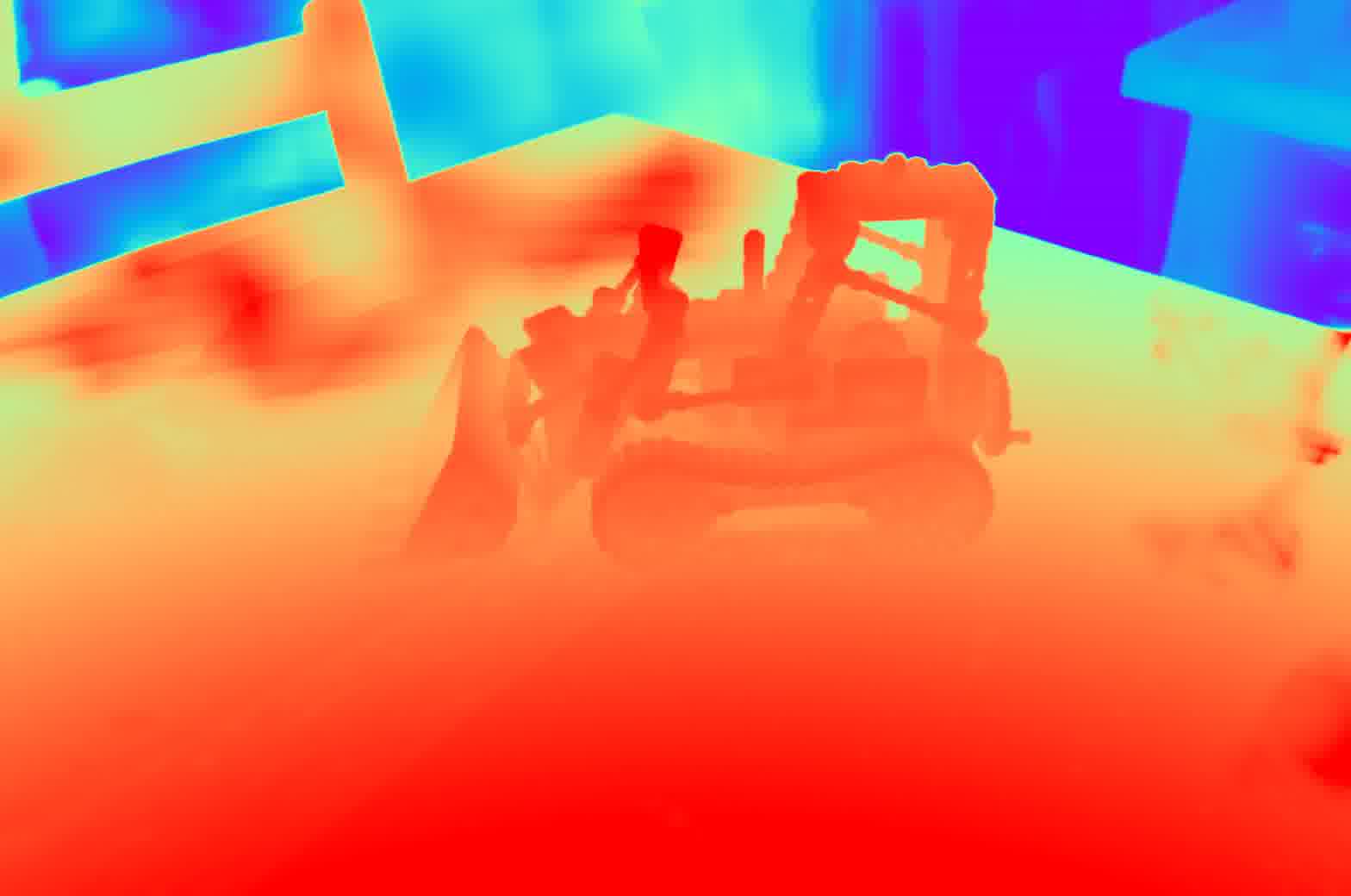}{(0, 0.8)}{(1.5, 1.6)}{(1.9, 0.3)}{(2.4, 1.25)}
     \end{subfigure}
     \hfill
    \begin{subfigure}[b]{0.155\textwidth}
         \includegraphics[width=\textwidth]{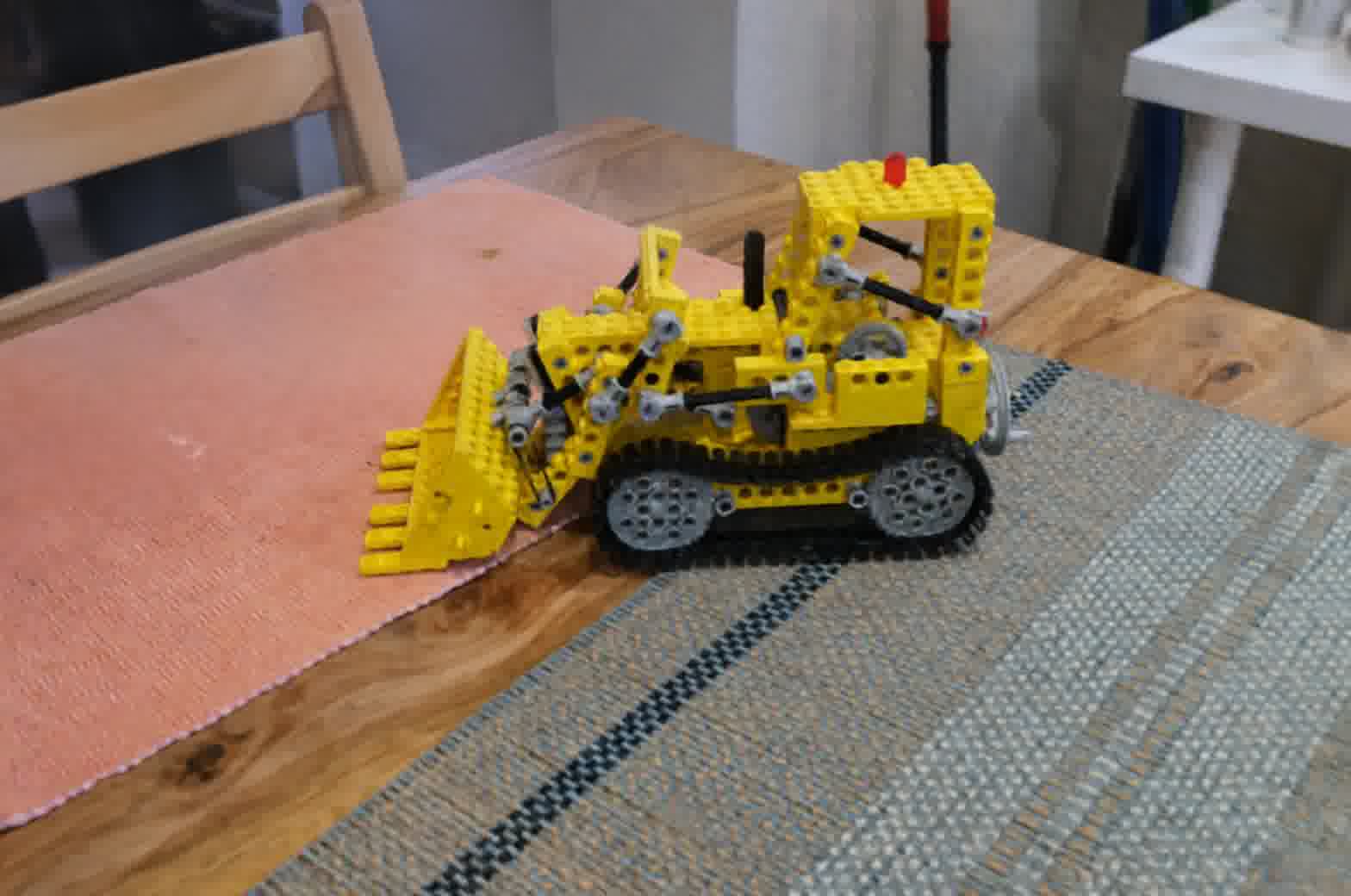}
     \end{subfigure}
    \hfill 
     \begin{subfigure}[b]{0.155\textwidth}
         \imagewithsquare{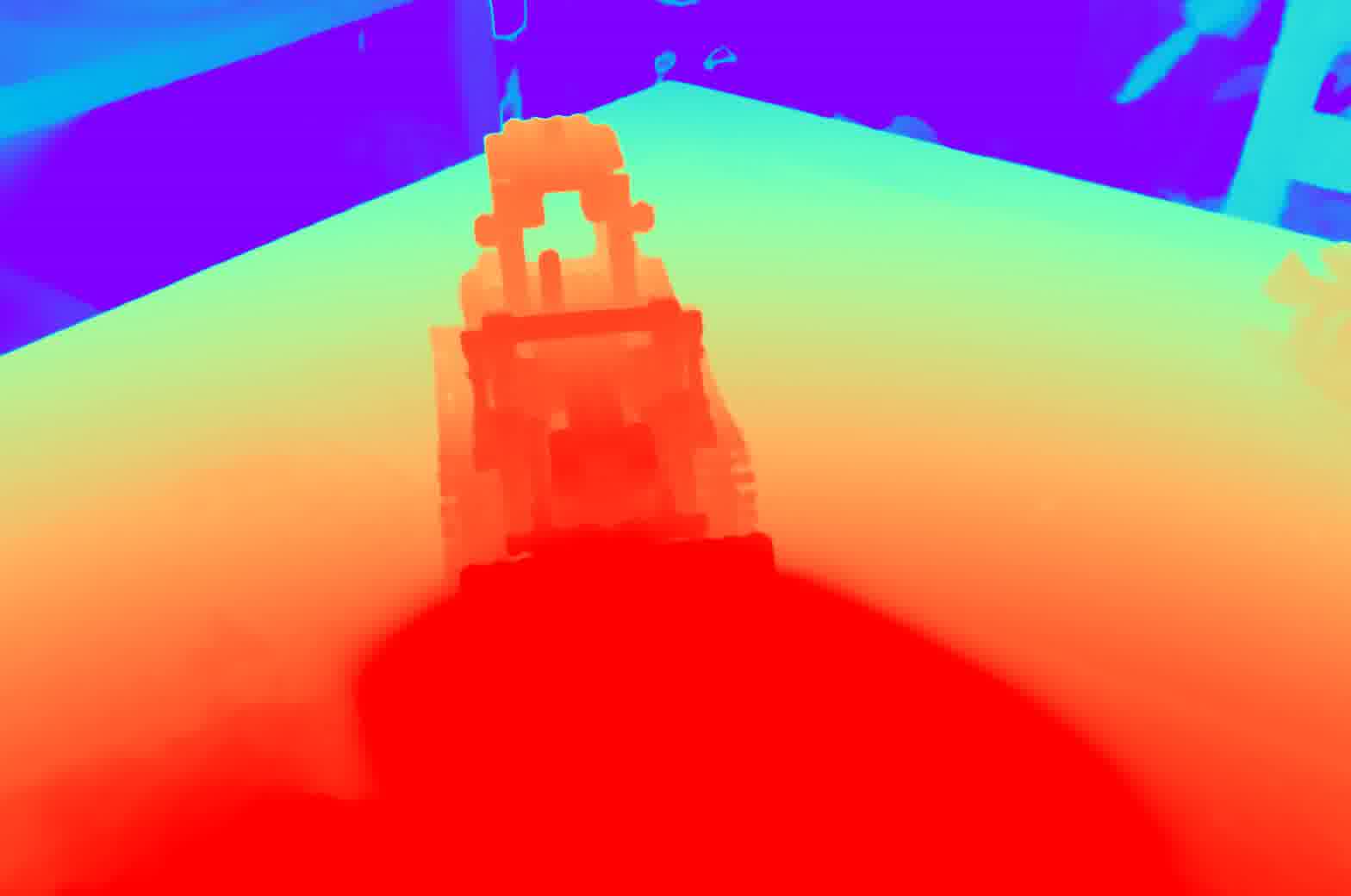}{(0.5, 0)}{(2, 0.75)}
     \end{subfigure}
     \hfill
    \begin{subfigure}[b]{0.155\textwidth}
         \centering
         \includegraphics[width=\textwidth]{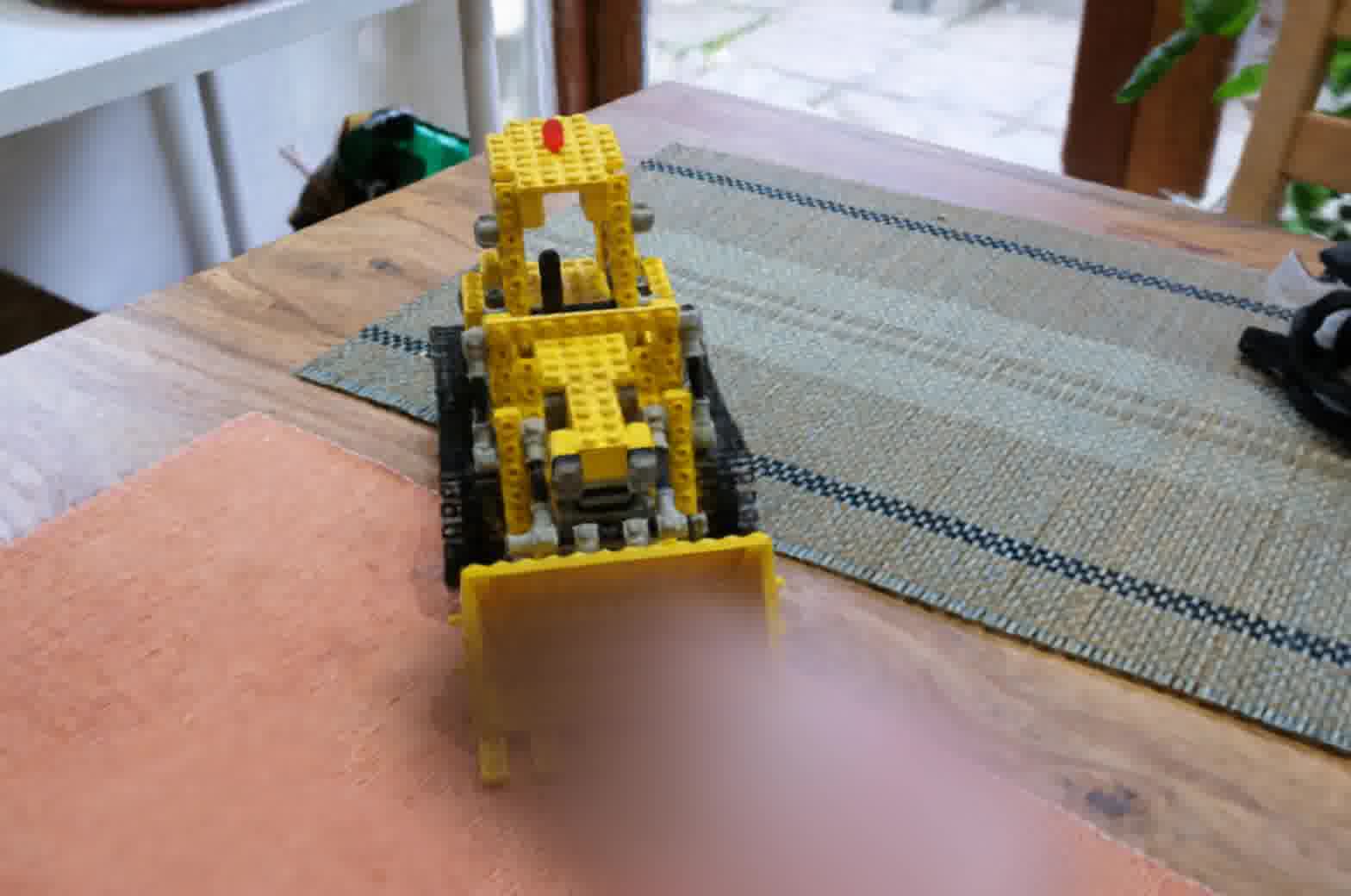}
     \end{subfigure}
     \\
     \rotatebox{90}{\hspace{0.5mm}\footnotesize With Scaling}
     \begin{subfigure}[b]{0.155\textwidth}
         \centering
         \includegraphics[width=\textwidth]{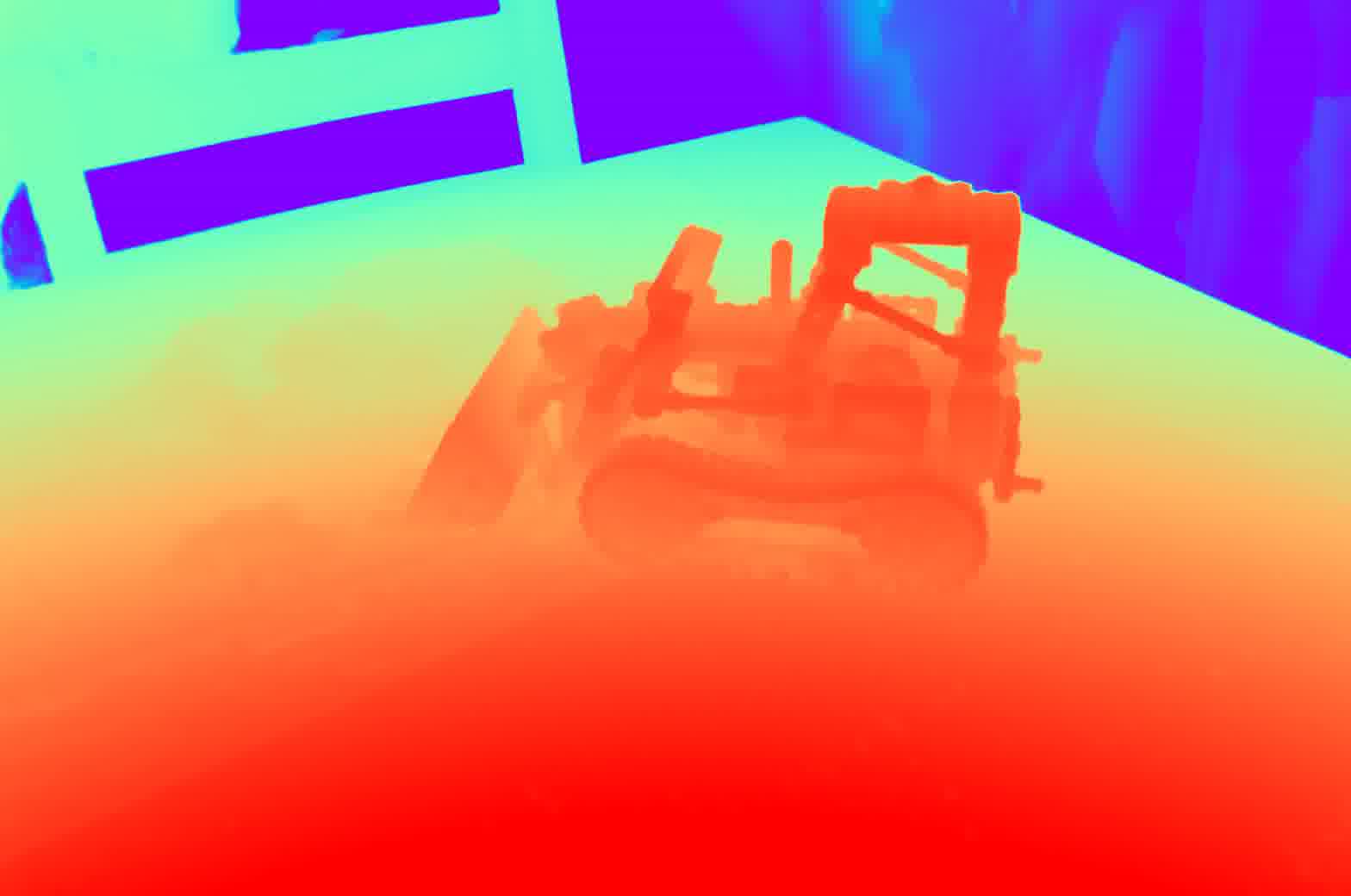}
     \end{subfigure}
     \hfill
     \begin{subfigure}[b]{0.155\textwidth}
         \centering
         \includegraphics[width=\textwidth]{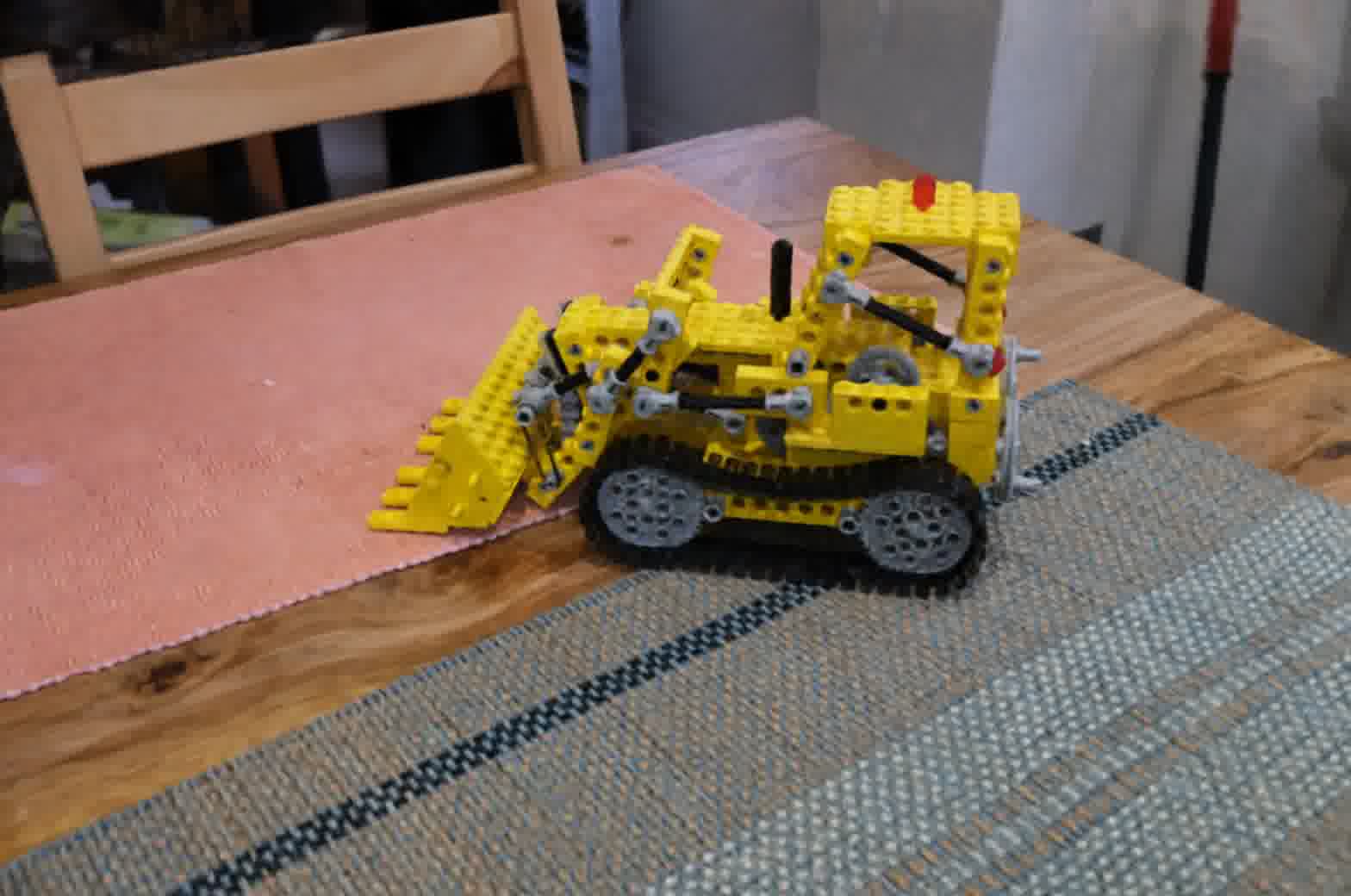}
     \end{subfigure}
     \hfill
     \begin{subfigure}[b]{0.155\textwidth}
         \centering
         \includegraphics[width=\textwidth]{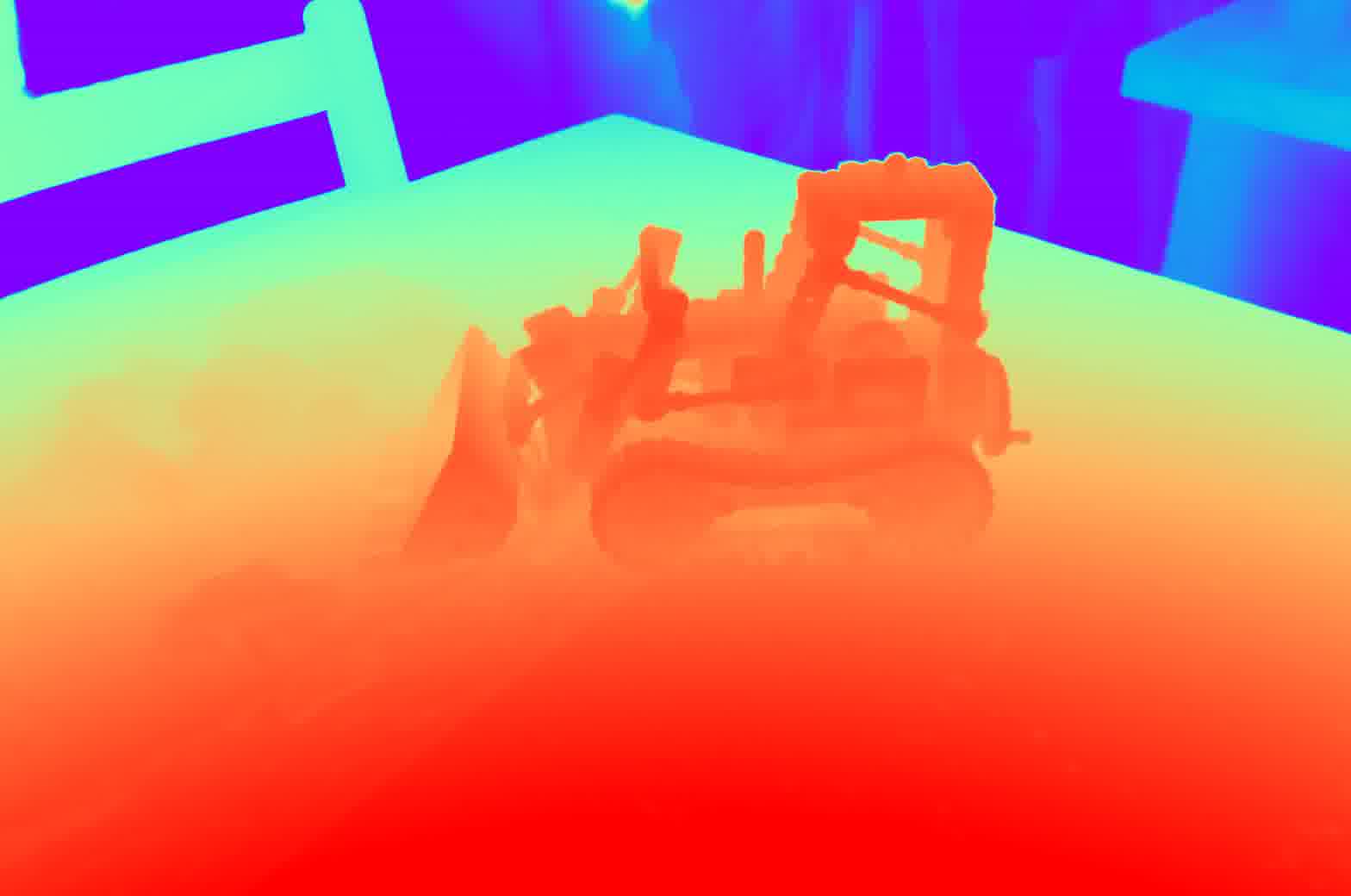}
     \end{subfigure}
     \hfill
    \begin{subfigure}[b]{0.155\textwidth}
         \centering
         \includegraphics[width=\textwidth]{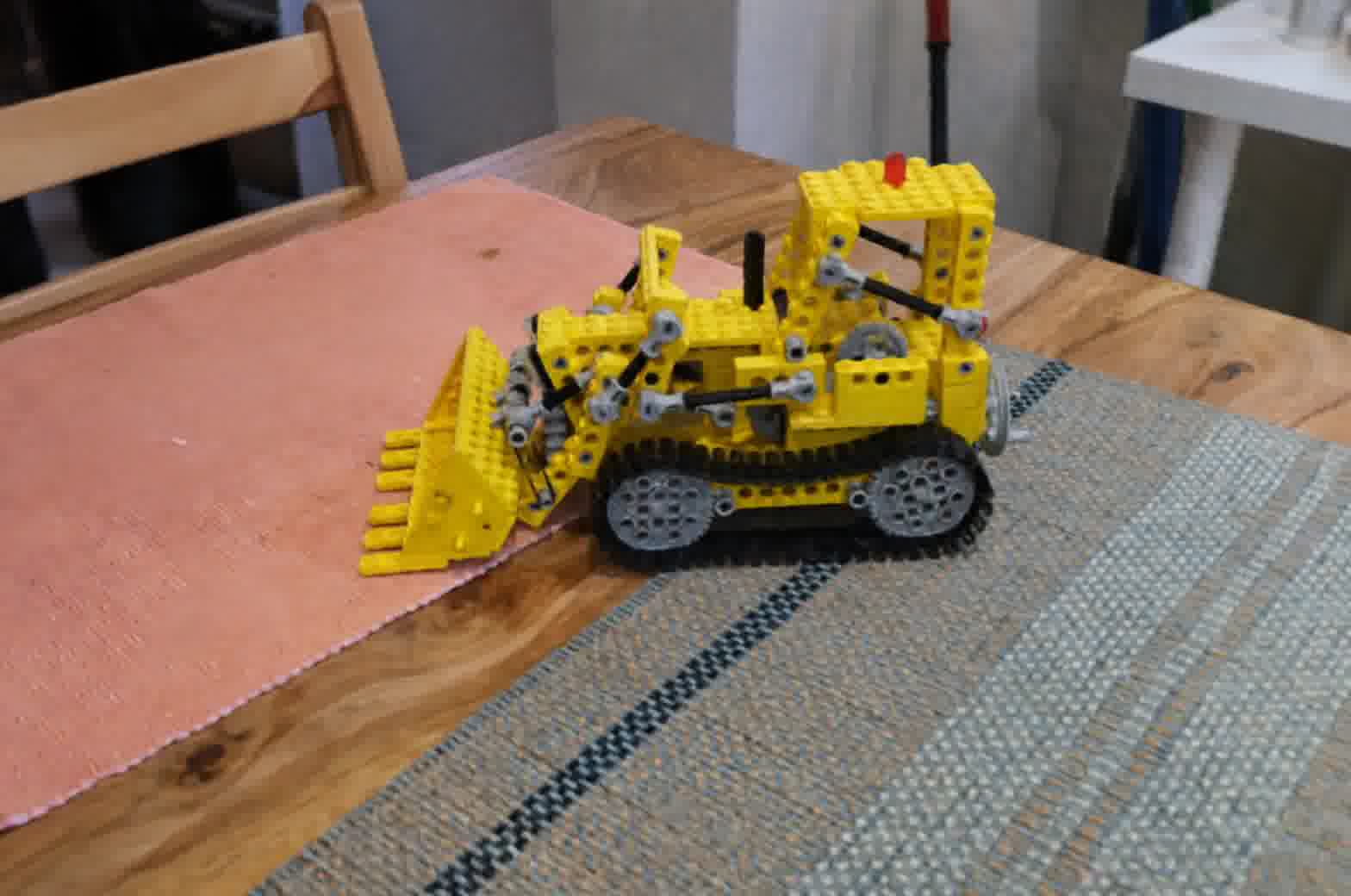}
     \end{subfigure}
    \hfill 
     \begin{subfigure}[b]{0.155\textwidth}
         \centering
         \includegraphics[width=\textwidth]{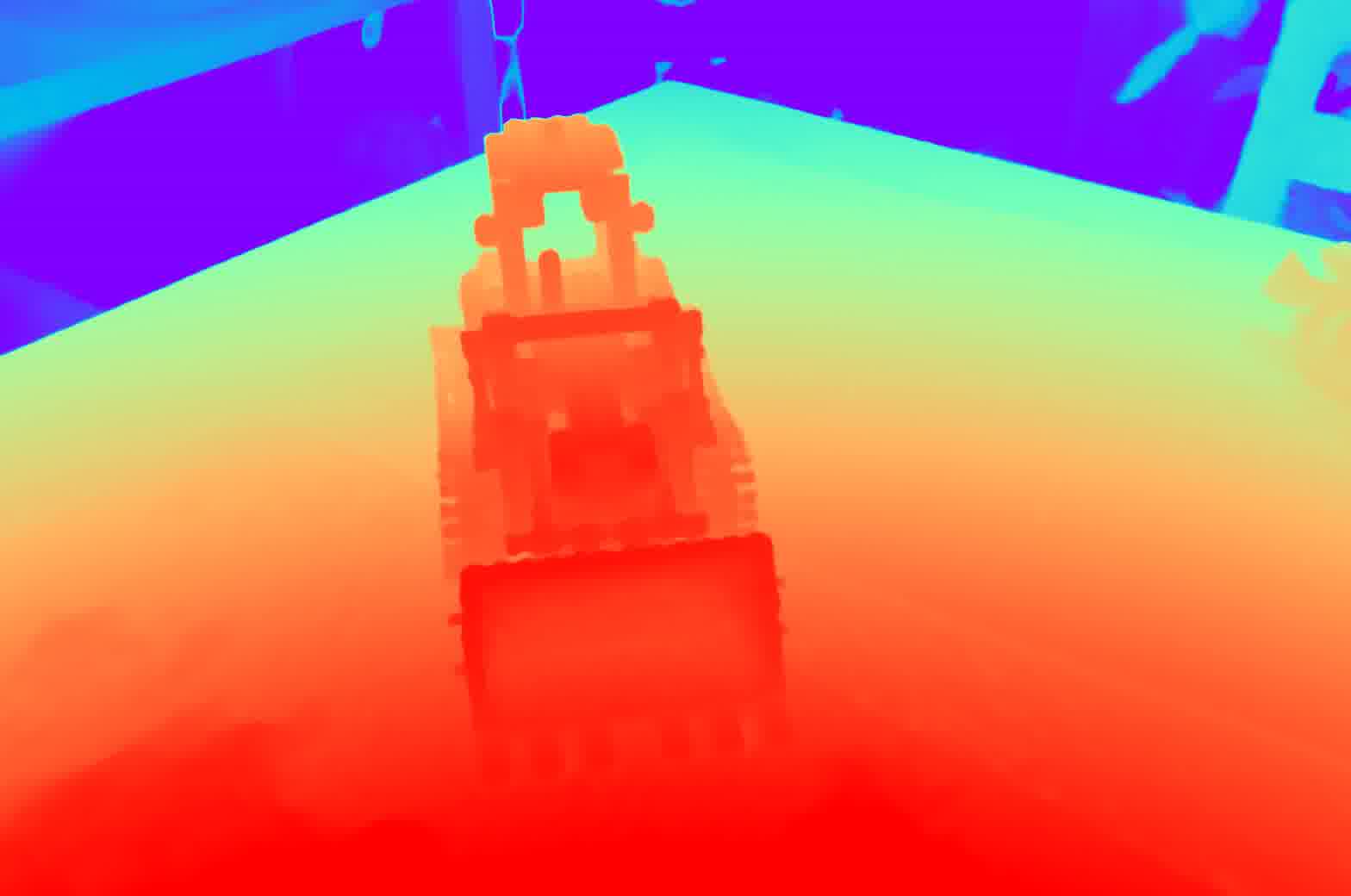}
     \end{subfigure}
     \hfill
    \begin{subfigure}[b]{0.155\textwidth}
         \centering
         \includegraphics[width=\textwidth]{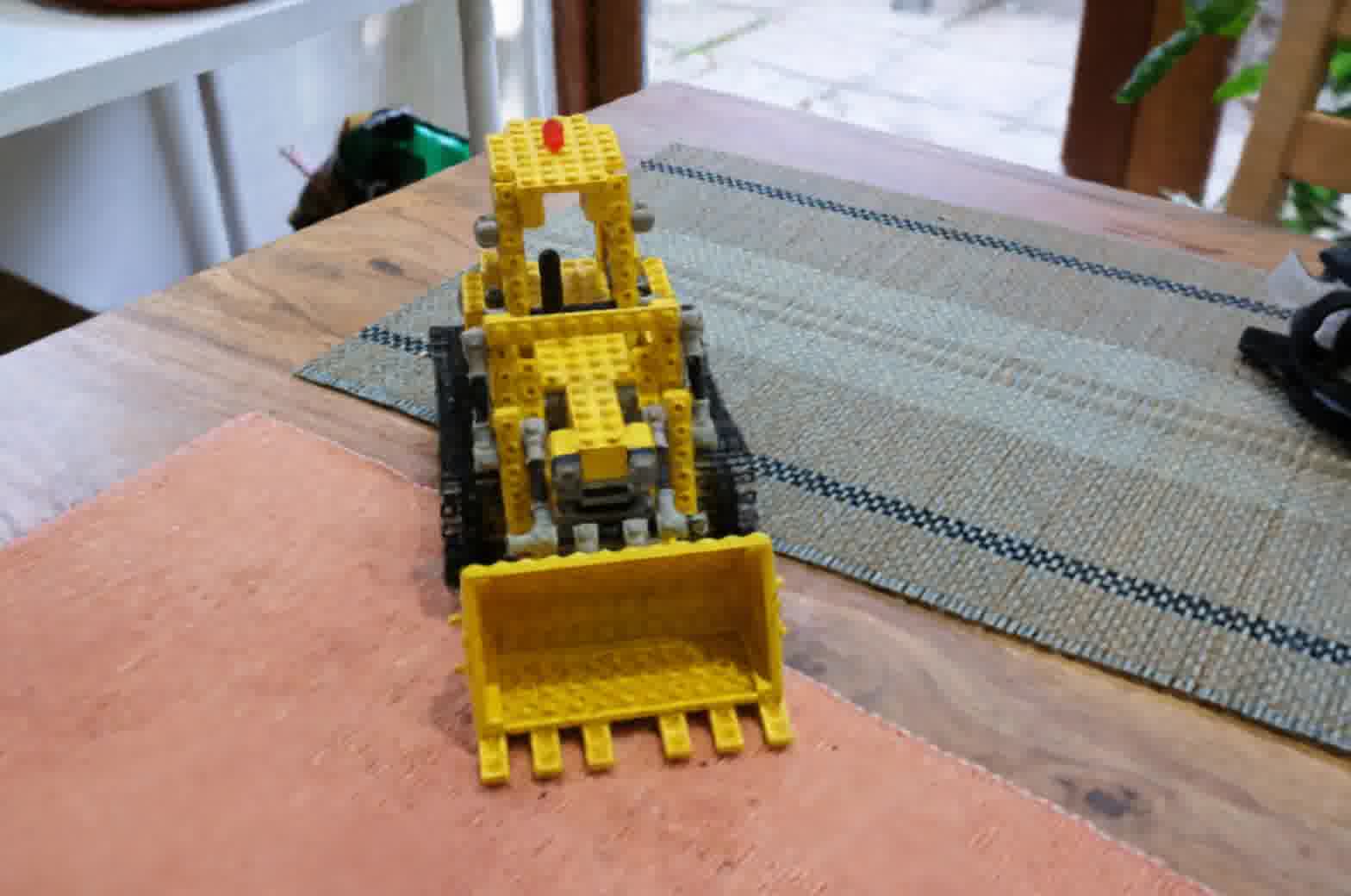}
     \end{subfigure}

        \caption{Results of \final{Improved} Direct Voxel Grid Optimization~\cite{SunSC22,SunSC22_2} original method (No Scaling) and improved with our proposed gradient scaling. Regions showing significant artifacts are highlighted with white rectangles. We can see that depth is more coherent with our gradient scaling, preventing floaters to appear. The depth is represented here with a warm-cool color palette, with red close to the camera and blue far. This is particularly visible in the top left and bottom right results and in the videos in Supplemental Materials.}
        \label{fig:dvgo_comp}
\end{figure*}
\begin{figure}
     \centering
    \rotatebox{90}{\hspace{4.5mm}\footnotesize No Scaling}
     \begin{subfigure}[b]{0.105\textwidth}
        \centering 
        \footnotesize Depth 
        \imagewithsquare{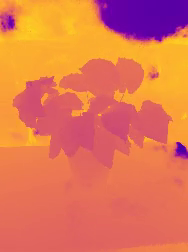}{(0.8,1.8)}{(1.7,2.3)}
     \end{subfigure}
     \begin{subfigure}[b]{0.105\textwidth}
         \centering
         \footnotesize Rendering
         \includegraphics[width=\textwidth]{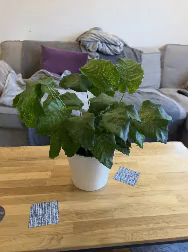}
     \end{subfigure}
     \begin{subfigure}[b]{0.105\textwidth}
         \centering
         \footnotesize Depth
         \imagewithtwosquare{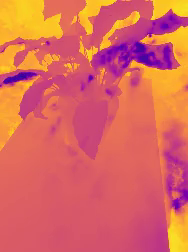}{(0,1.4)}{(1.65,2.2)}{(1.4,0.2)}{(1.7,1.3)}
     \end{subfigure}
    \begin{subfigure}[b]{0.105\textwidth}
         \centering
          \footnotesize Rendering
         \includegraphics[width=\textwidth]{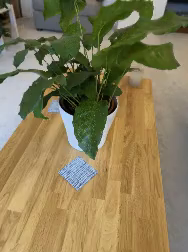}
     \end{subfigure}\\
         \rotatebox{90}{\hspace{3.5mm}\footnotesize With Scaling}
     \begin{subfigure}[b]{0.105\textwidth}
         \centering
         \includegraphics[width=\textwidth]{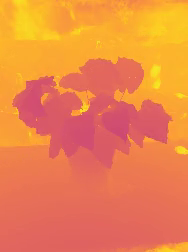}
     \end{subfigure}
     \begin{subfigure}[b]{0.105\textwidth}
         \centering
         \includegraphics[width=\textwidth]{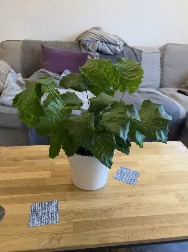}
     \end{subfigure}
     \begin{subfigure}[b]{0.105\textwidth}
         \centering
         \includegraphics[width=\textwidth]{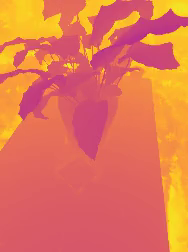}
     \end{subfigure}
    \begin{subfigure}[b]{0.105\textwidth}
         \centering
         \includegraphics[width=\textwidth]{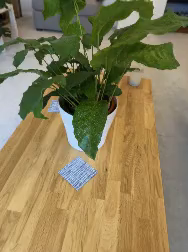}
     \end{subfigure}\\
         \rotatebox{90}{\hspace{4.5mm}\footnotesize No Scaling}
     \begin{subfigure}[b]{0.105\textwidth}
         \imagewithtwosquare{{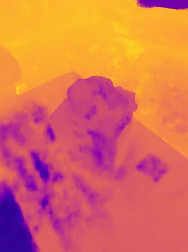}}{(0,0)}{(1.4,1.6)}{(1.12,2.05)}{(1.7,2.3)}
     \end{subfigure}
     \begin{subfigure}[b]{0.105\textwidth}
         \centering
         \includegraphics[width=\textwidth]{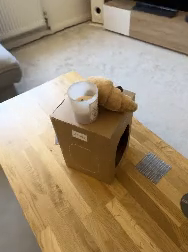}
     \end{subfigure}
     \begin{subfigure}[b]{0.105\textwidth}
         \imagewithtwosquare{{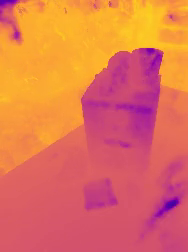}}{(0.7,0)}{(1.7,1.8)}{(0,1.8)}{(1.7,2.3)}
     \end{subfigure}
    \begin{subfigure}[b]{0.105\textwidth}
         \centering
         \includegraphics[width=\textwidth]{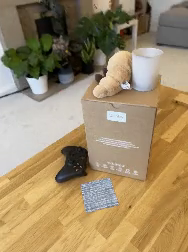}
     \end{subfigure}\\
    \rotatebox{90}{\hspace{3.5mm}\footnotesize With Scaling}
     \begin{subfigure}[b]{0.105\textwidth}
         \centering
         \includegraphics[width=\textwidth]{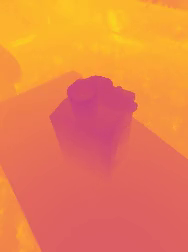}
     \end{subfigure}
     \begin{subfigure}[b]{0.105\textwidth}
         \centering
         \includegraphics[width=\textwidth]{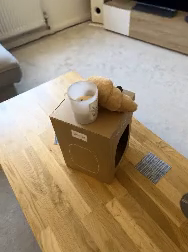}
     \end{subfigure}
     \begin{subfigure}[b]{0.105\textwidth}
         \centering
         \includegraphics[width=\textwidth]{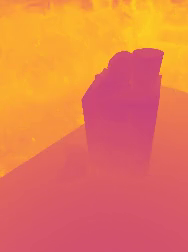}
     \end{subfigure}
    \begin{subfigure}[b]{0.105\textwidth}
         \centering
         \includegraphics[width=\textwidth]{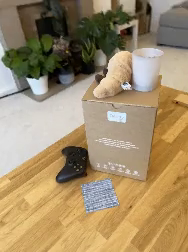}
     \end{subfigure}

        \caption{Results of InstantNGP~\cite{mueller2022instant} with and without our scaling. \textbf{Views shown here are training views}. Regions showing significant artifacts are highlighted with white rectangles. As with other volumetric representations, our proposed scaling helps reconstruct more coherent scenes. This is particularly visible in the videos available in Supplemental Materials. The depth is represented here with the plasma color palette, with purple close to the camera and yellow far.}
        \label{fig:ingp_comp}
\end{figure}

\subsection{Clamped quadratic scaling}
Given the nature of the problem, a straightforward candidate could be to scale gradients by $(\delta^i_{p})^2$ to compensate for the sampling density quadratic decay. While this indeed solves the near-camera sampling \final{imbalance} problem, it leads to very strong gradients far from the camera, preventing correctly learning surfaces as shown in Figure~\ref{fig:comp} (middle-row).

As described in equation \ref{eq:rho}, the sampling density at a given point is the result of the sum of sampling densities from all cameras.
 This means that the assumption of an inverse quadratic nature of the sampling density for a volume element is mostly valid near each given camera, as sampling from other cameras is negligible in this area.
 We thus opted not to modify the gradient when reaching the main content of the scene (around a depth of 1) and let the distribution of the camera guide the sampling density in these regions, as illustrated in Fig~\ref{fig:vis_cam2}.
 
 \final{Having more cameras seeing a point has a potentially positive effect} as it focuses samples in interesting regions as opposed to the near-camera \final{sampling imbalance} which is purely due to the nature of the ray-marching process and produces artifacts.
 We illustrate in Figure~\ref{fig:comp} that our gradient scaling approach helps focus the gradient near the center of the scene, prevents background collapse, and that using a purely quadratic scaling results in worse convergence and reconstruction. 
 
We compare three different approaches of gradient scaling using a custom implementation of NGP \cite{mueller2022instant}. For all of them, we set the near plane to 0.
In the top row we can see that without any scaling, density builds up very quickly near the training camera, and while some of it is removed throughout the optimization, some of this close-camera density remains at the end.
In the middle row, we can see that scaling purely quadratically leads to a bias toward far density. The table is reconstructed in the background at first and the optimization does not recover from this bad initialization.
We present the results of our clamped gradient scaling in the bottom row. We clearly observe the advantage of the method, the geometry of the table is quickly and well reconstructed, and it converges towards a better estimate without background collapse.
\begin{figure*}
     \centering
    \rotatebox{90}{\hspace{4mm}\footnotesize No Scaling}
     \begin{subfigure}[b]{0.24\linewidth}
        \centering \footnotesize Depth
         \imagewithsquare{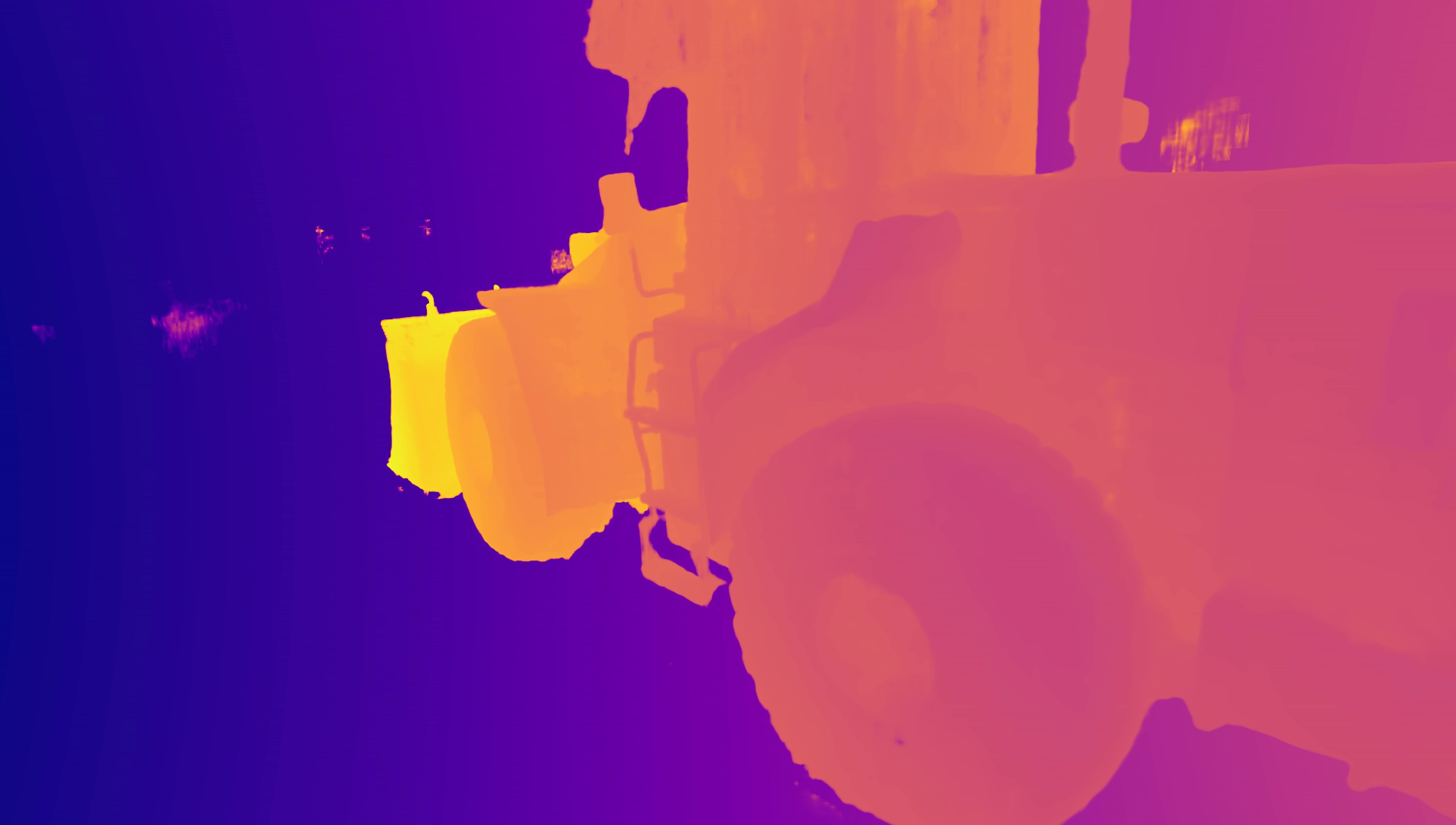}{(0.2,1.0)}{(1.2,1.8)}
     \end{subfigure}
     \begin{subfigure}[b]{0.24\linewidth}
         \centering
         \footnotesize Rendering
         \includegraphics[width=\textwidth]{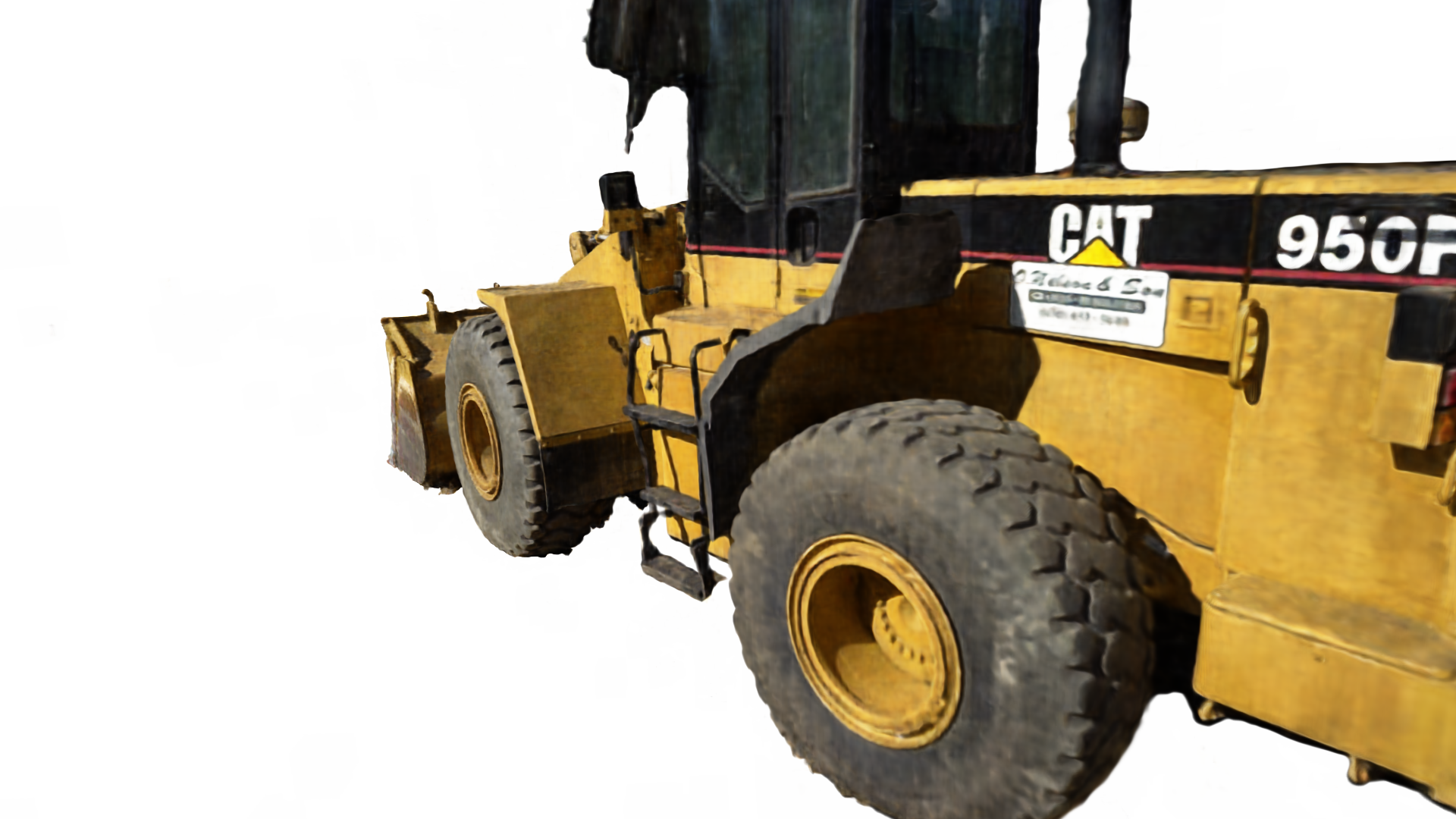}
     \end{subfigure}
     \begin{subfigure}[b]{0.24\linewidth}
         \centering
         \footnotesize Depth
         \imagewithtwosquare{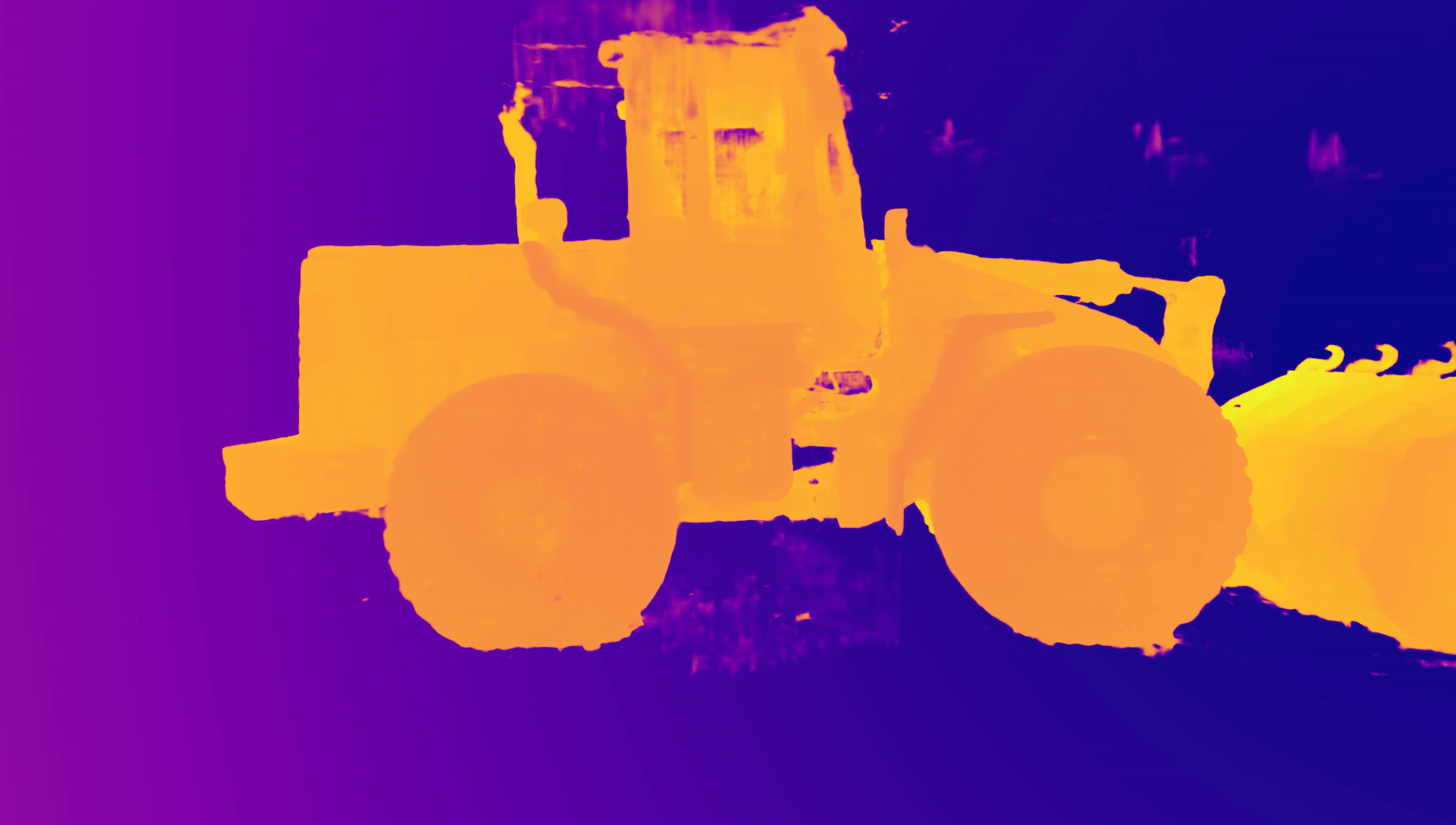}{(1.2,1.4)}{(3.65,2.1)}{(1.5,0.2)}{(2.4,1.0)}
     \end{subfigure}
    \begin{subfigure}[b]{0.24\linewidth}
         \centering
         \footnotesize Rendering
         \includegraphics[width=\textwidth]{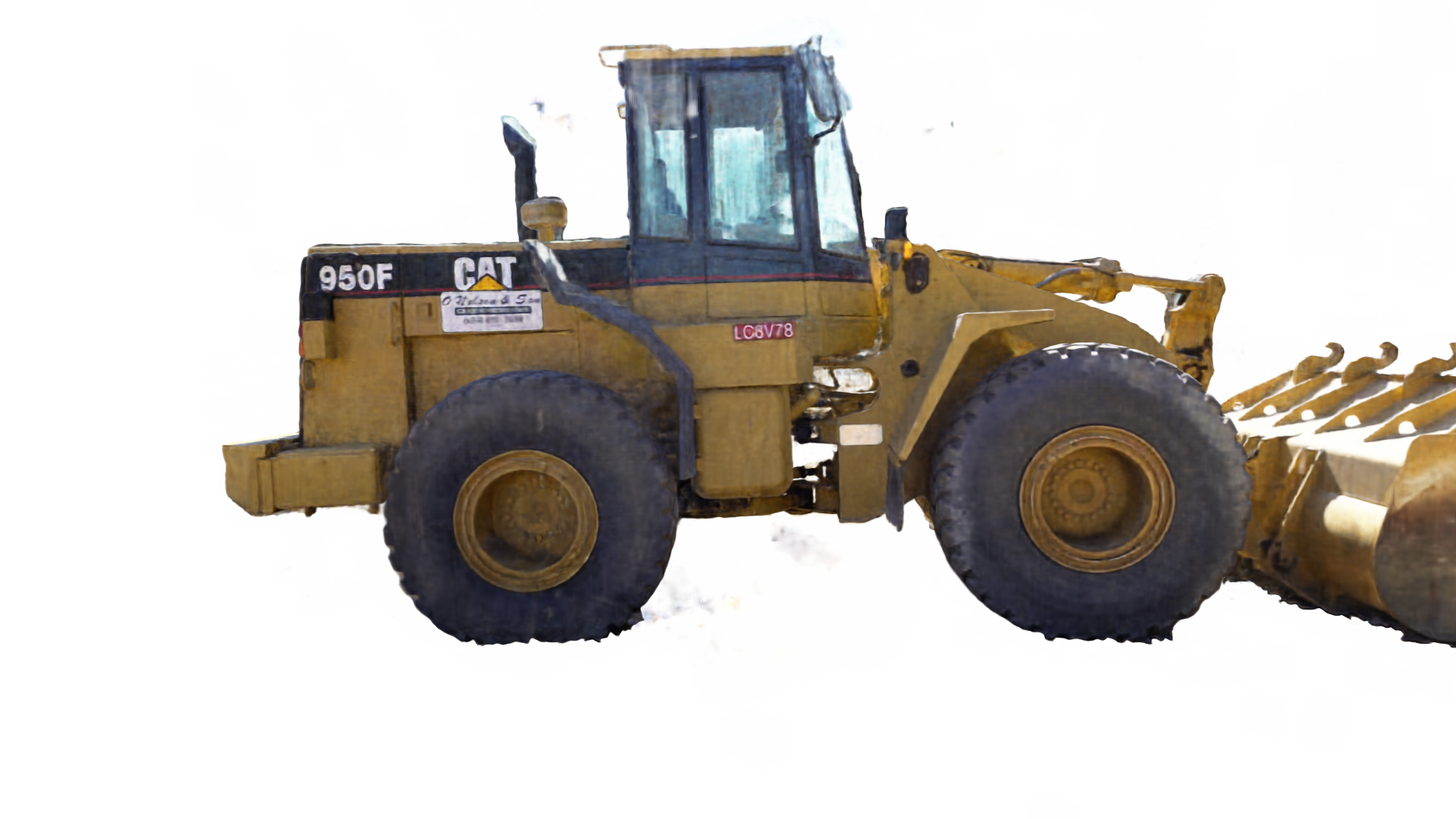}
     \end{subfigure}
         \rotatebox{90}{\hspace{4mm}\footnotesize With Scaling}
     \begin{subfigure}[b]{0.24\linewidth}
         \centering
         \includegraphics[width=\textwidth]{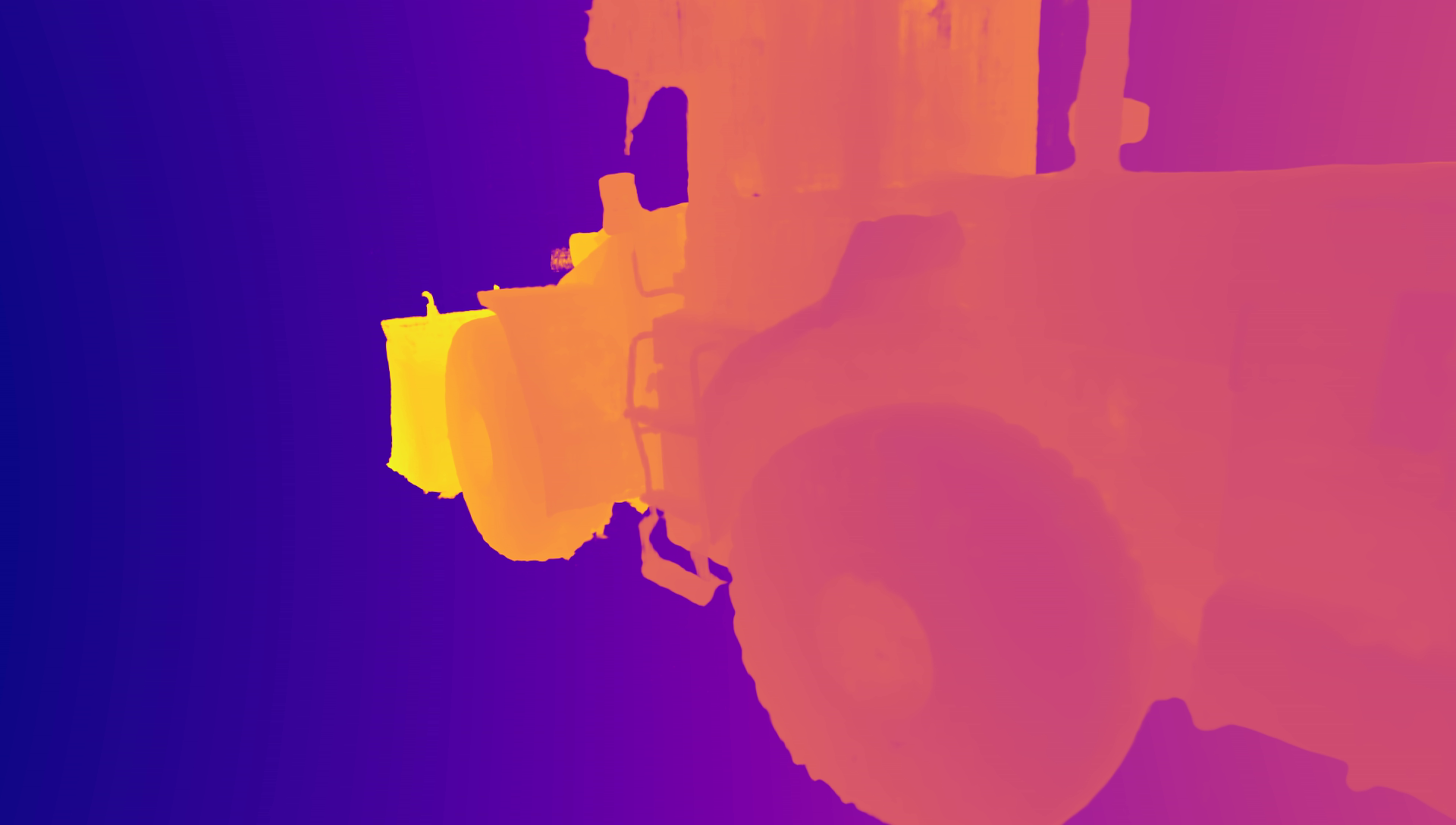}
     \end{subfigure}
     \begin{subfigure}[b]{0.24\linewidth}
         \centering
         \includegraphics[width=\textwidth]{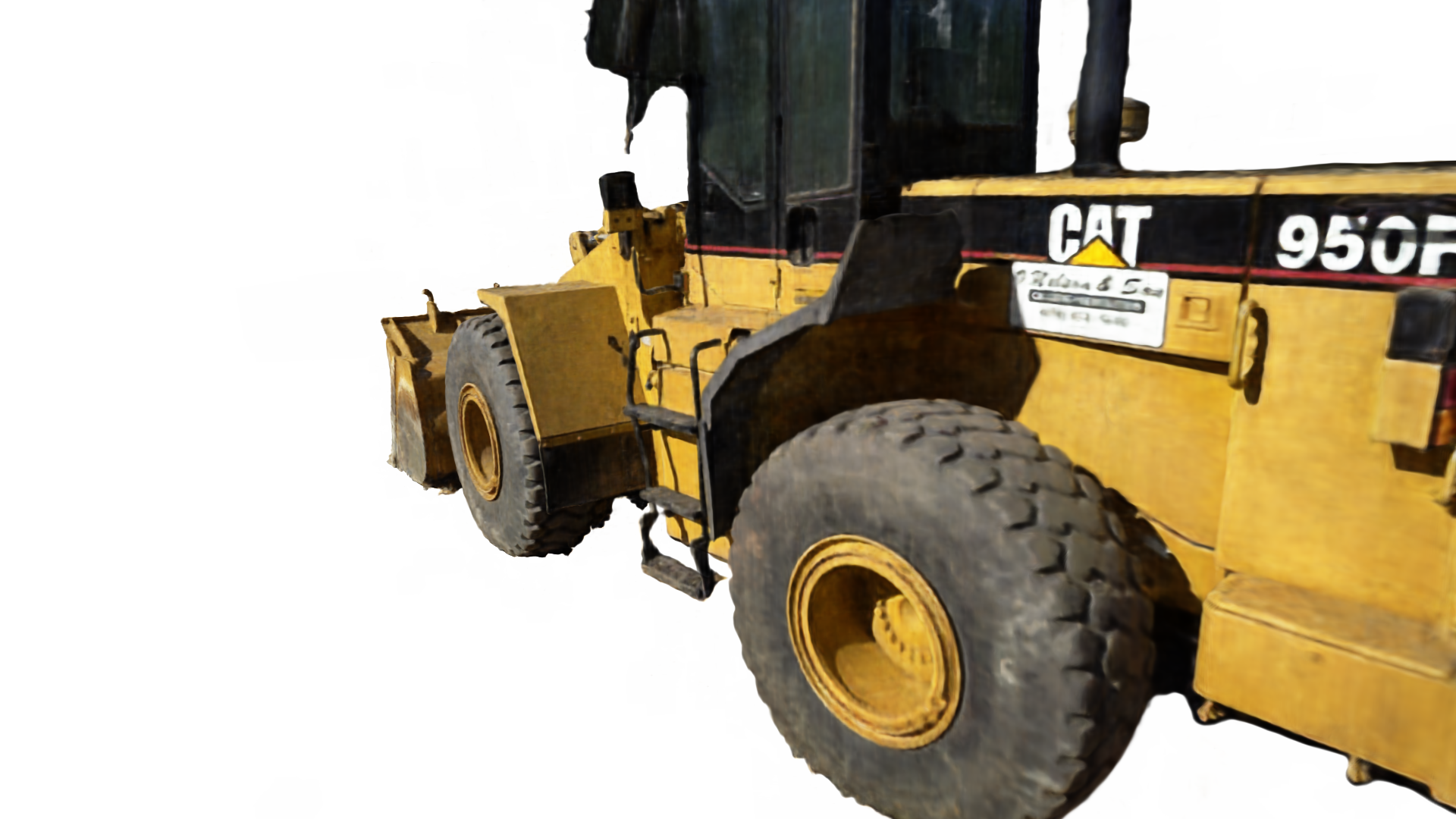}
     \end{subfigure}
     \begin{subfigure}[b]{0.24\linewidth}
         \centering
         \includegraphics[width=\textwidth]{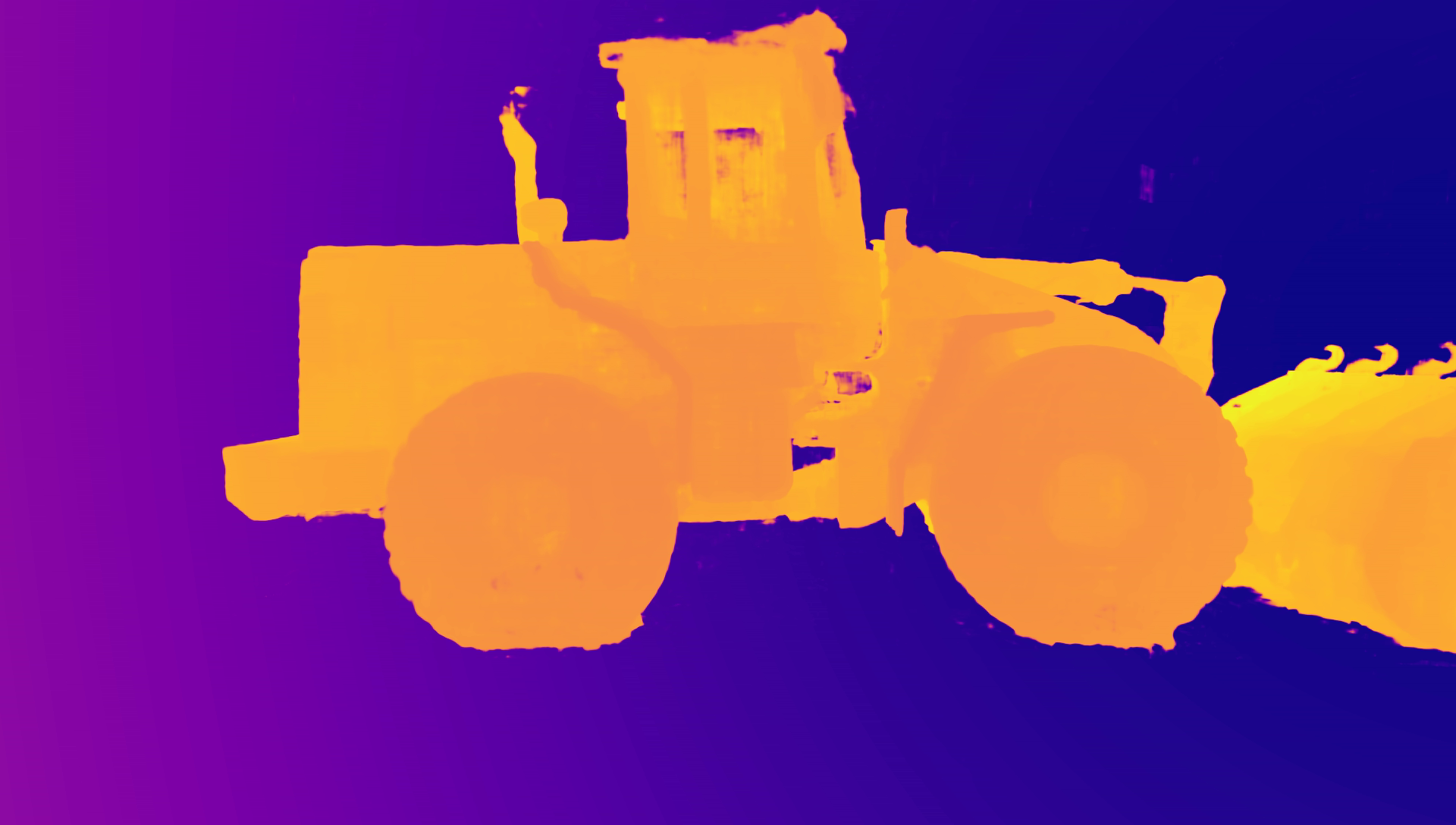}
     \end{subfigure}
    \begin{subfigure}[b]{0.24\linewidth}
         \centering
         \includegraphics[width=\textwidth]{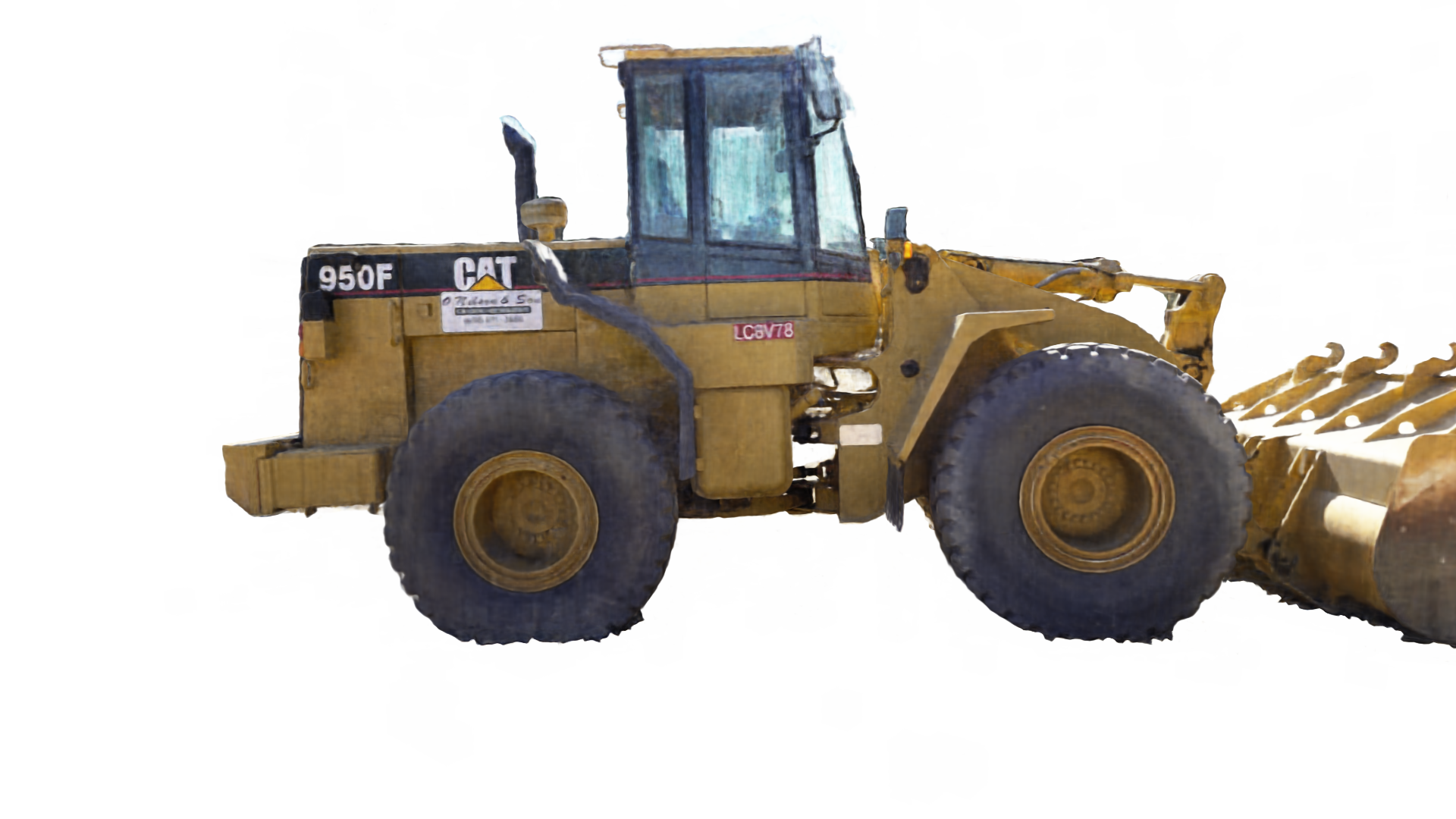}
     \end{subfigure}

         \rotatebox{90}{\hspace{4mm}\footnotesize No Scaling}
     \begin{subfigure}[b]{0.24\linewidth}
         \imagewithtwosquare{{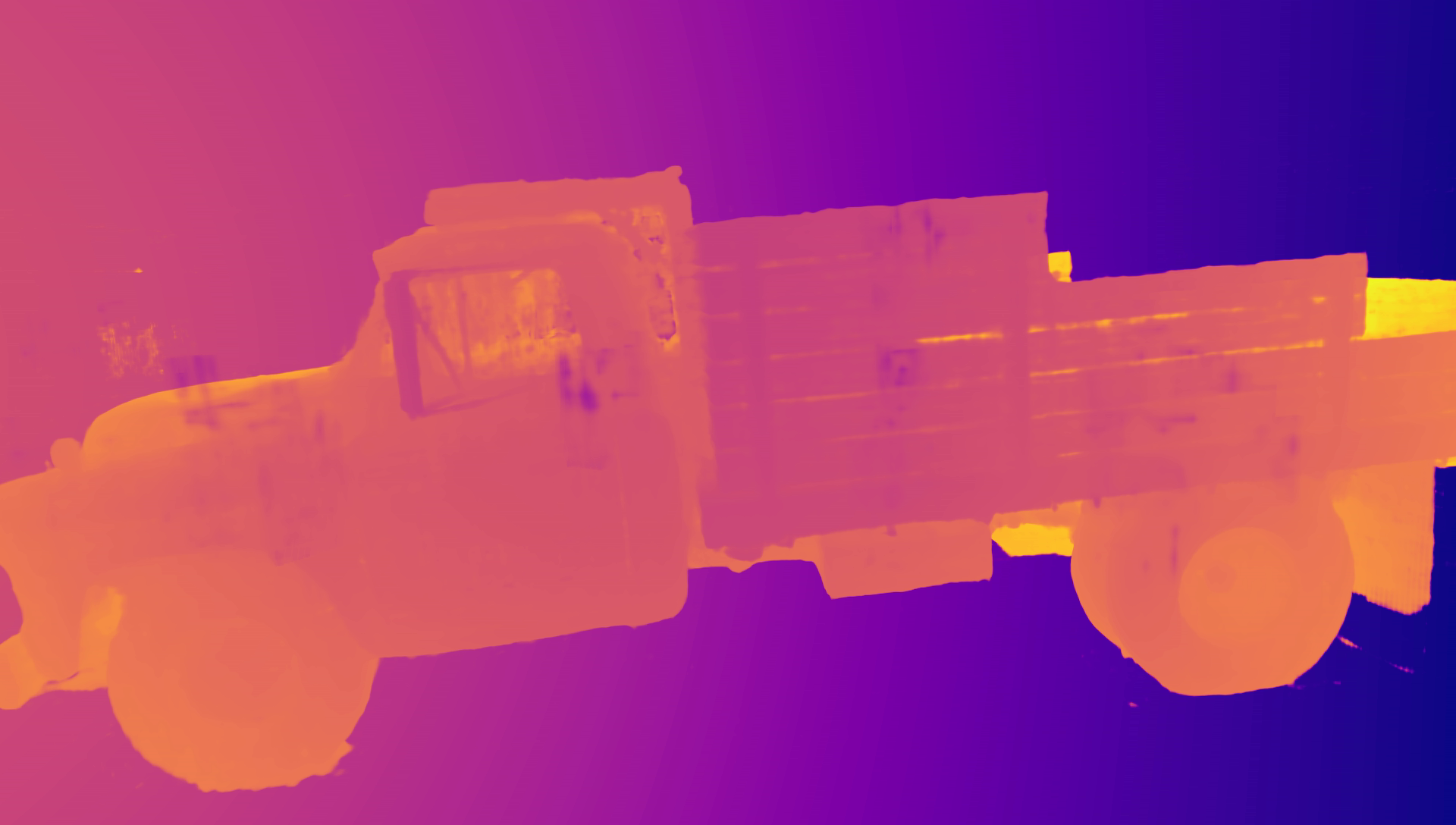}}{(0.1,0.8)}{(0.9,1.6)}{(1.3,0.9)}{(1.8,1.4)}

     \end{subfigure}
     \begin{subfigure}[b]{0.24\linewidth}
         \centering
         \includegraphics[width=\textwidth]{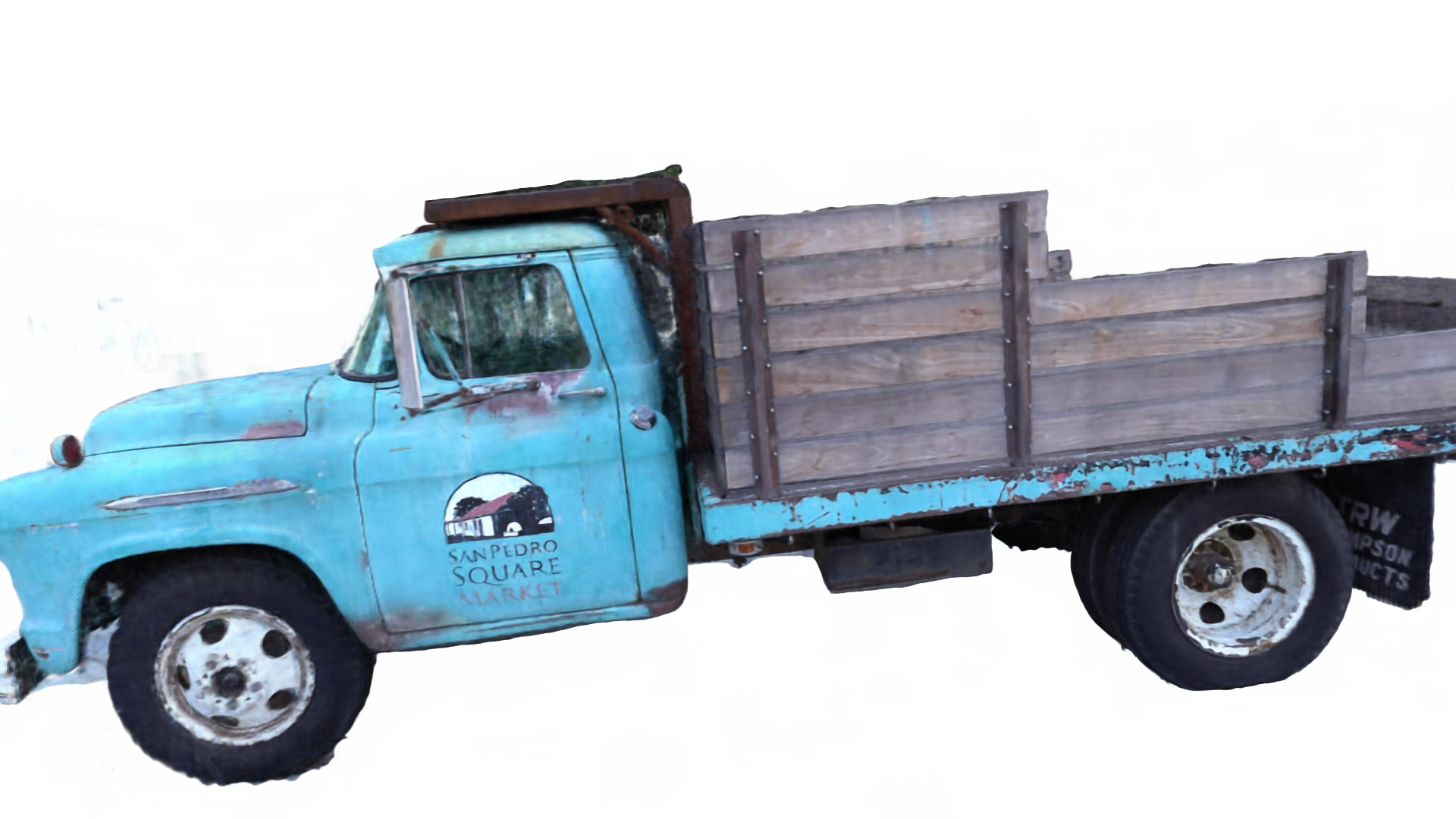}
     \end{subfigure}
     \begin{subfigure}[b]{0.24\linewidth}
         \imagewithtwosquare{{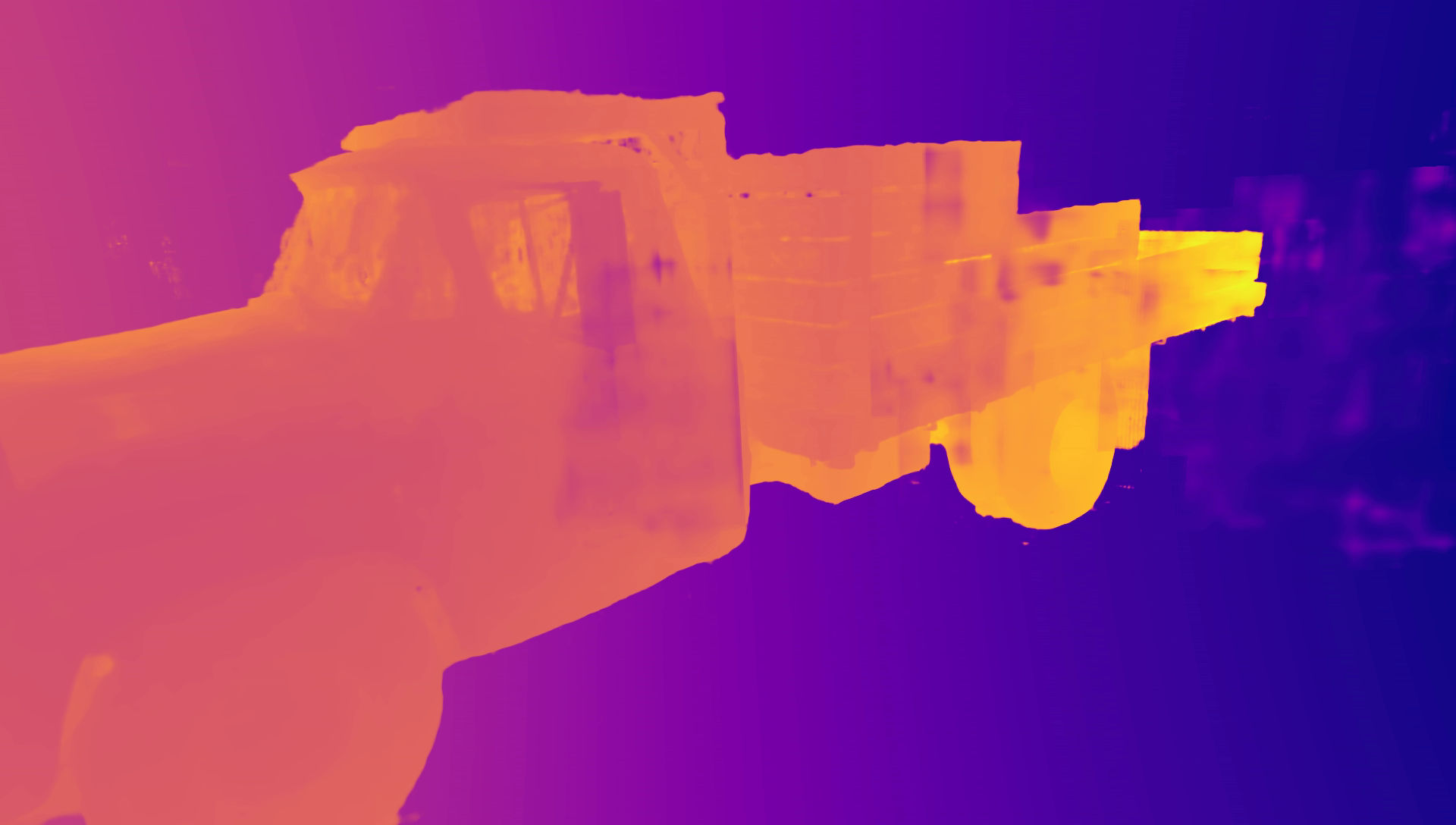}}{(1.2,0.8)}{(1.8,1.7)}{(2.1,0.6)}{(3.8,1.9)}
     \end{subfigure}
    \begin{subfigure}[b]{0.24\linewidth}
         \centering
         \includegraphics[width=\textwidth]{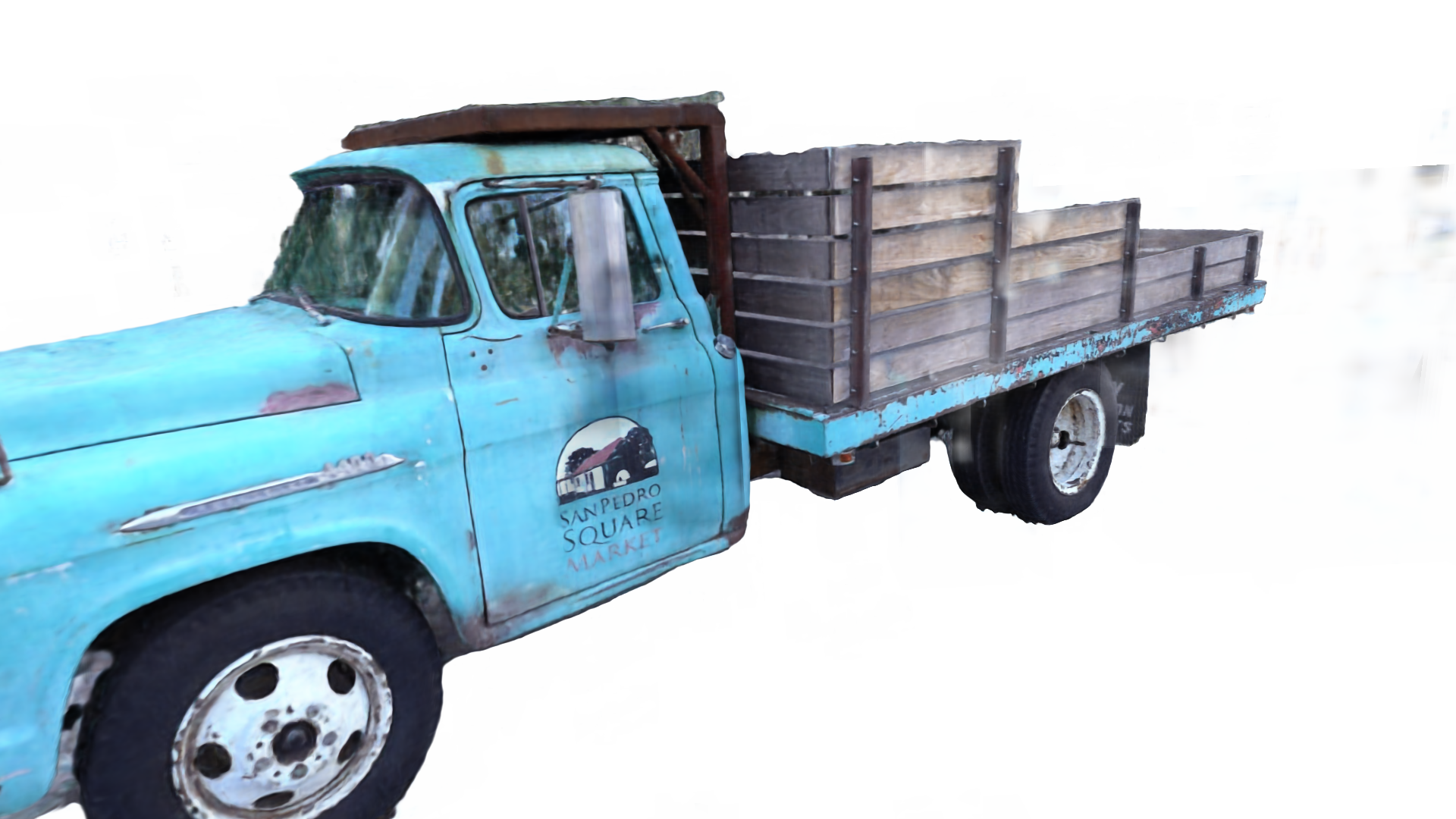}
     \end{subfigure}
    \rotatebox{90}{\hspace{4mm}\footnotesize With Scaling}
     \begin{subfigure}[b]{0.24\linewidth}
         \centering
         \includegraphics[width=\textwidth]{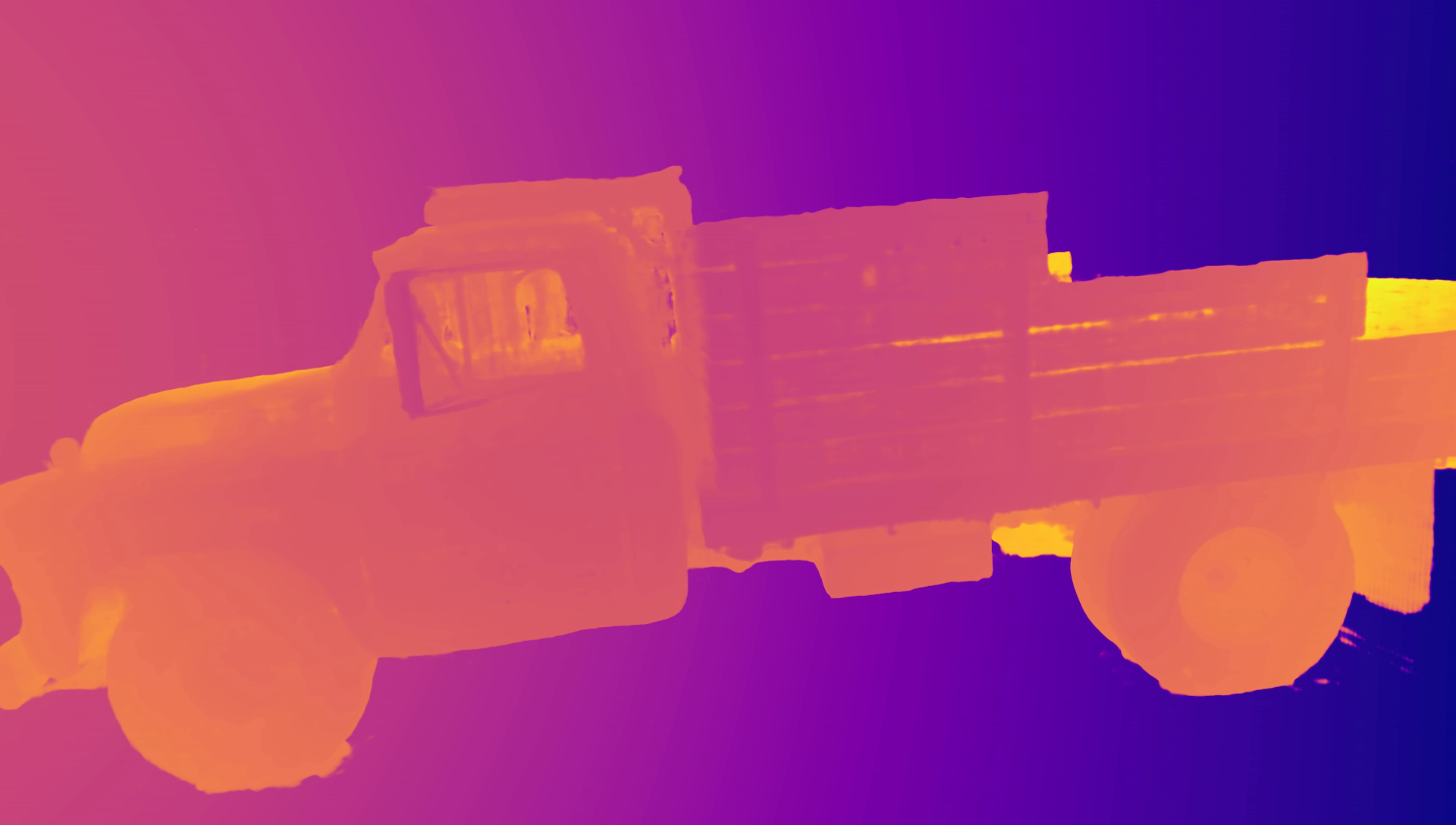}
     \end{subfigure}
     \begin{subfigure}[b]{0.24\linewidth}
         \centering
         \includegraphics[width=\textwidth]{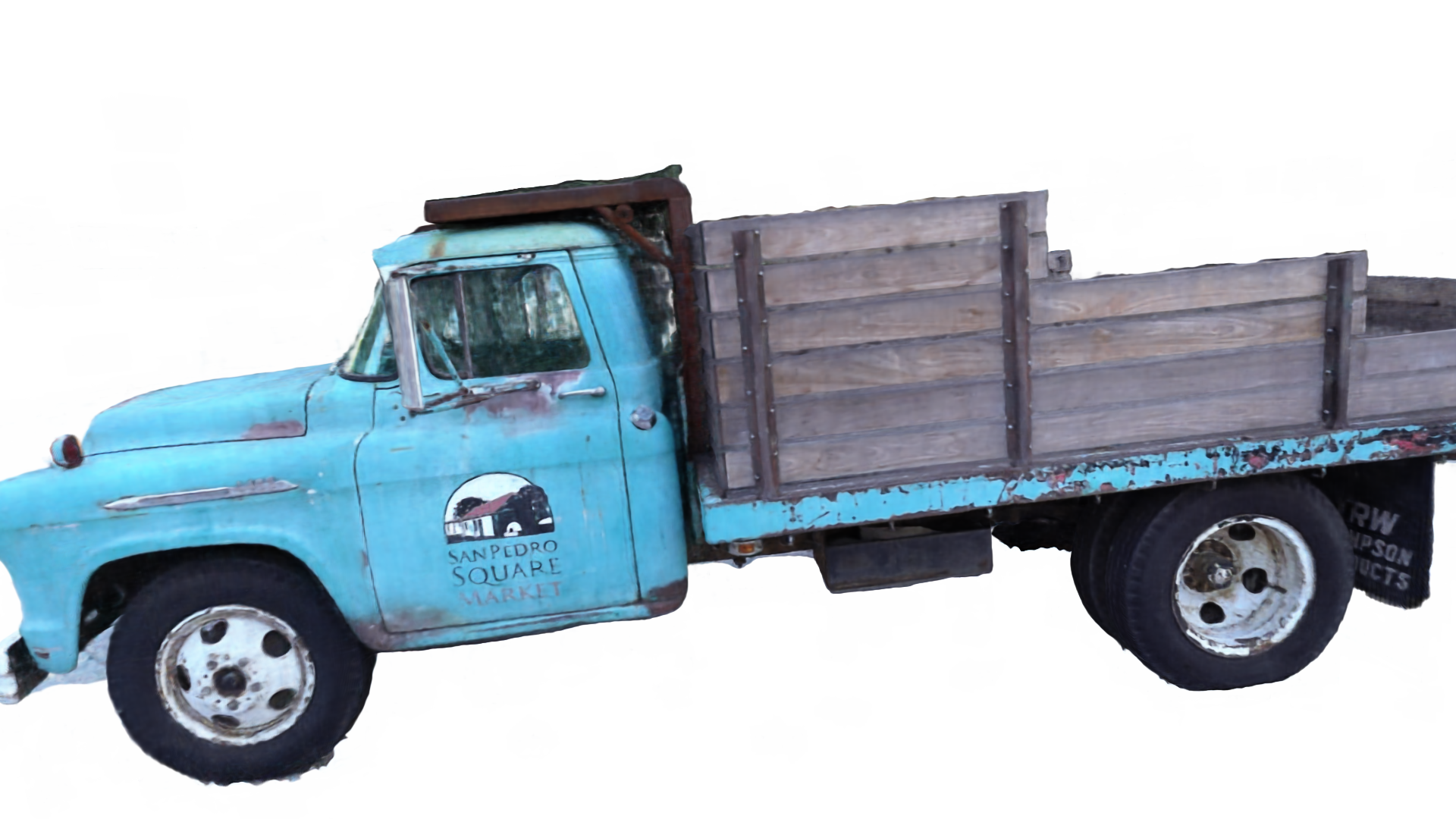}
     \end{subfigure}
     \begin{subfigure}[b]{0.24\linewidth}
         \centering
         \includegraphics[width=\textwidth]{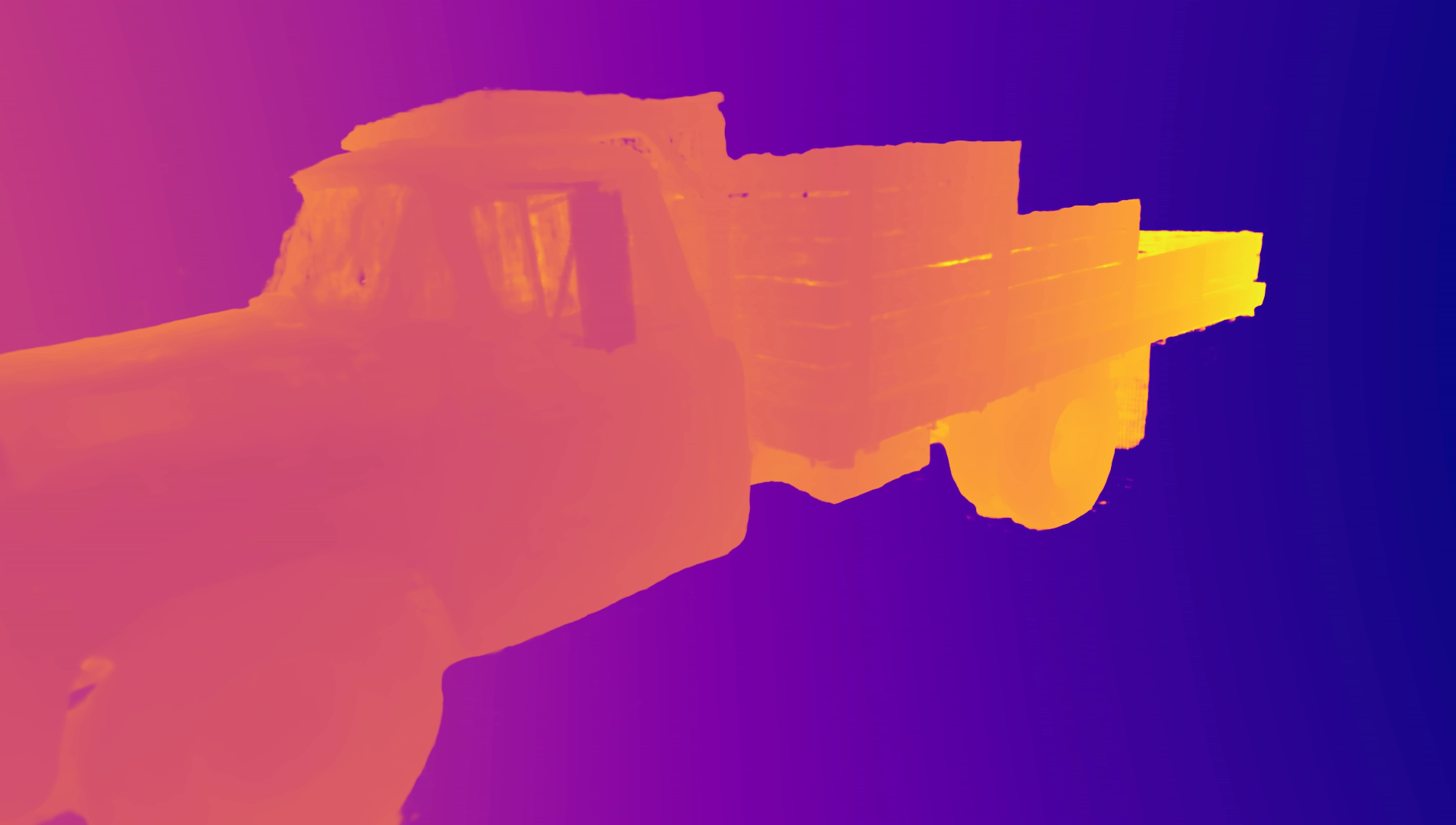}
     \end{subfigure}
    \begin{subfigure}[b]{0.24\linewidth}
         \centering
         \includegraphics[width=\textwidth]{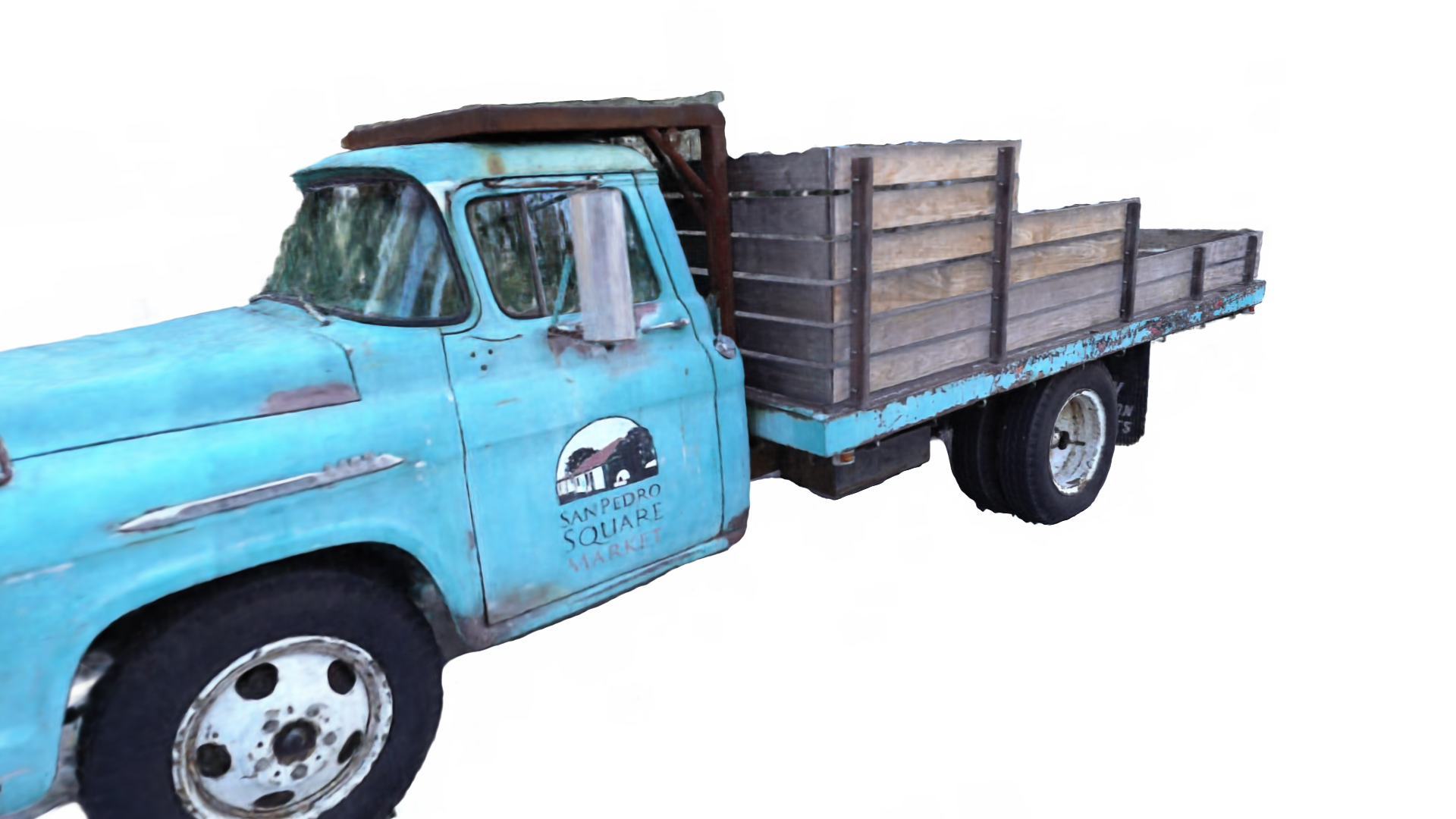}
     \end{subfigure}

        \caption{Results of TensoRF~\cite{Chen2022ECCV} with and without our scaling. Regions showing significant artifacts are highlighted with white rectangles. Here too, our scaling proves efficient in removing floaters due to background-collapse. TensoRF depth visualization attributes blue to red color to the background, the object depth follows the plasma color palette, with purple close to the camera and yellow far.}
        \label{fig:trf_comp}
\end{figure*}

\subsection{Gradient scaling for various NeRF representation}
In this section, we show the effect of adding gradient scaling to various existing methods and show that it removes background collapse effects for all of them. \emph{This is particularly visible in the videos available in Supplemental Materials}. For each method we use their implementation, as much as possible, leading to different color codings of the depth, we define it for each figure. \final{Unless mentioned otherwise we present testing views.}

\subsubsection{\final{Improved} Direct Voxel Grid Optimization}
\final{DVGO~\cite{SunSC22} uses a scalar voxel grid to represent density and a grid of features with a shallow MLP for colors.}
We use the author's implementation \final{which includes the improvement proposed in the follow-up work \cite{SunSC22_2}} and adjust it to enable our gradient scaling scheme using code similar to Fig~\ref{code}. In Fig~\ref{fig:dvgo_comp} we show a comparison for two scenes with and without gradient scaling. Using our gradient scaling approach \final{drastically reduces the amount of} near-camera floaters.

\subsubsection{Instant NGP}
\final{Instant NGP~\cite{mueller2022instant} uses a multilevel hash-grid of features and a shallow MLP to represent both density and colors of a scene.}
We use a \final{custom} PyTorch NGP implementation with the gradient scaling scheme implemented. In Fig~\ref{fig:ingp_comp} we show a comparison for two scenes (Exotic Plant and Croissant) with and without gradient scaling, this time \emph{we visualize training frames} and their respective depth after training. This shows the root cause of the problem: density appears very near cameras, perfectly matching the RGB appearance for this given camera, but leading to floaters when seen from novel viewpoints. Using our gradient scaling approach, density does not build up near cameras.

\subsubsection{TensoRF}
\final{TensoRF~\cite{Chen2022ECCV} uses a factorized decomposition of a 4D tensor representing the scene to model both density and color.}
We show in Fig~\ref{fig:trf_comp} a comparison to TensoRF and see similar improvement to the other volumetric representations. The results using our scaling show more coherent depth and do not exhibit floaters.

\subsubsection{MipNeRF360}
\begin{figure*}
     \centering
    \rotatebox{90}{\hspace{0.85cm}\footnotesize RGB}
     \begin{subfigure}[b]{0.24\linewidth}
         \centering \small Original MipNERF360
         \includegraphics[width=\textwidth]{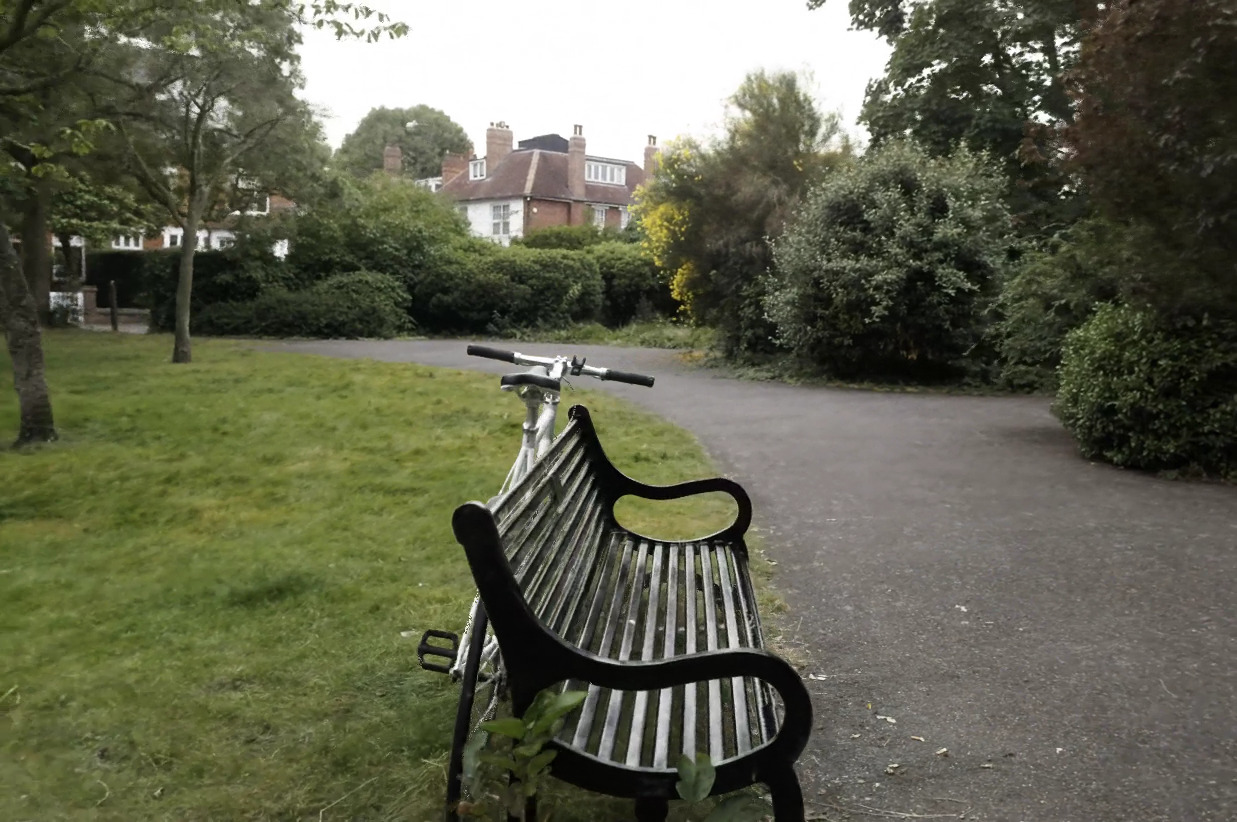}
     \end{subfigure}
     \begin{subfigure}[b]{0.24\linewidth}
         \centering \small No $\mathcal{L}_{\mathrm{dist}}$
         \includegraphics[width=\textwidth]{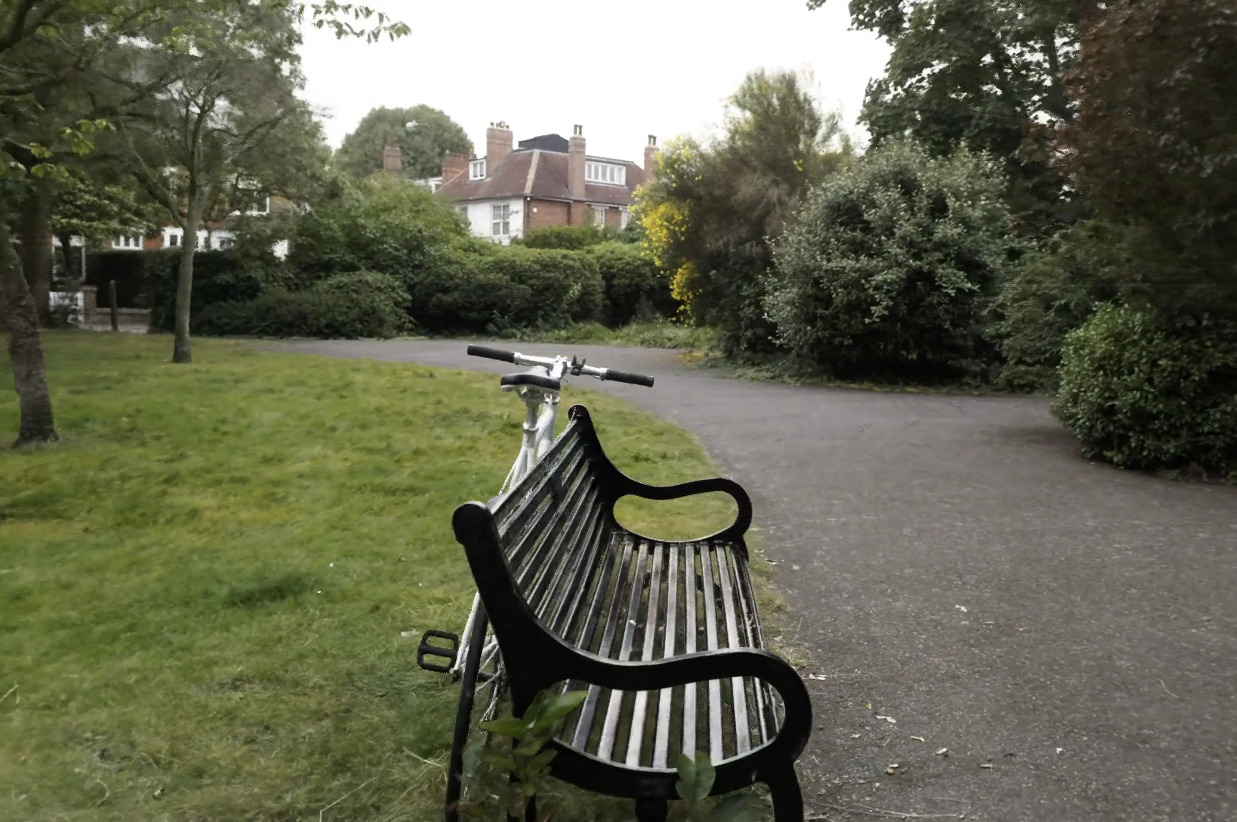}
     \end{subfigure}
    \begin{subfigure}[b]{0.24\linewidth}
         \centering No $\mathcal{L}_{\mathrm{dist}}$,  $np=0$, $s_{\nabla p}$
         \includegraphics[width=\textwidth]{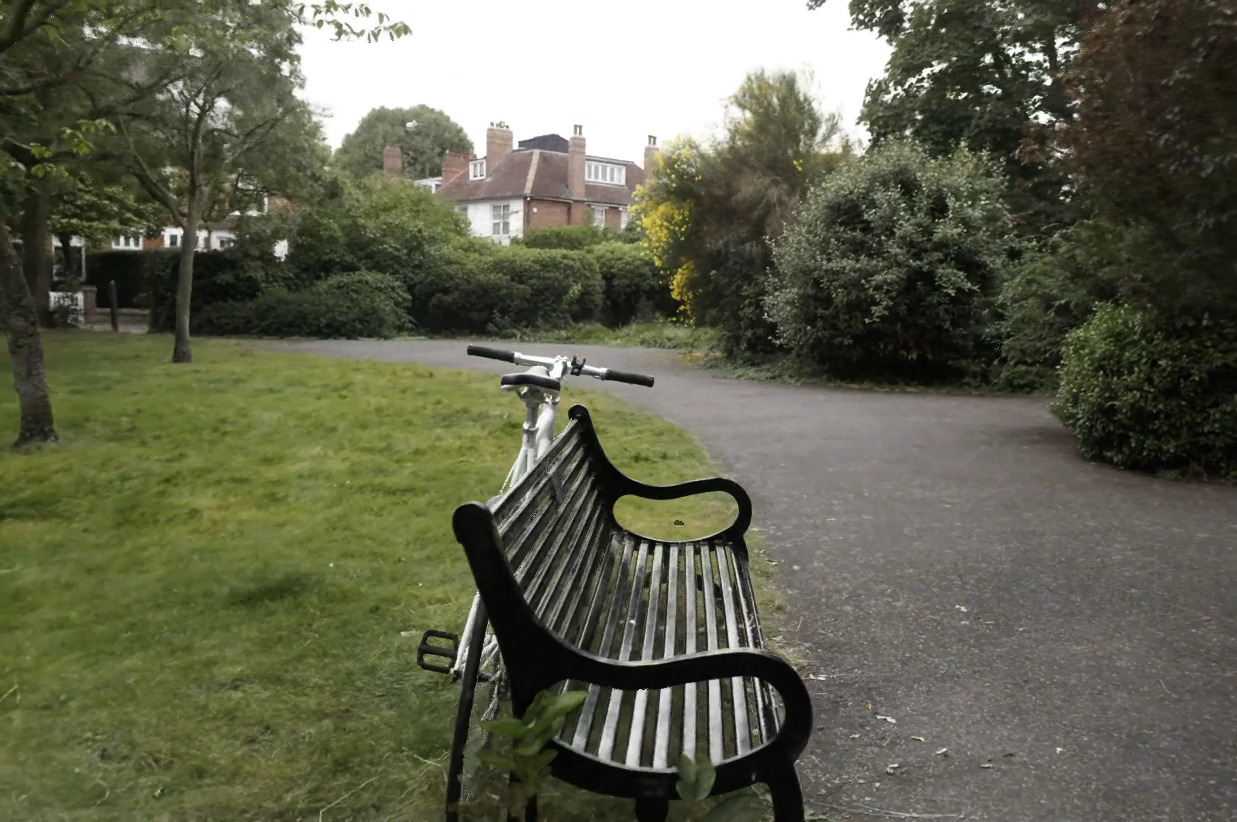}
     \end{subfigure}
     \begin{subfigure}[b]{0.24\linewidth}
         \centering Ours \\
         \centering $\mathcal{L}_{\mathrm{dist}}$,  $np=0$, $s_{\nabla p}$
         \includegraphics[width=\textwidth]{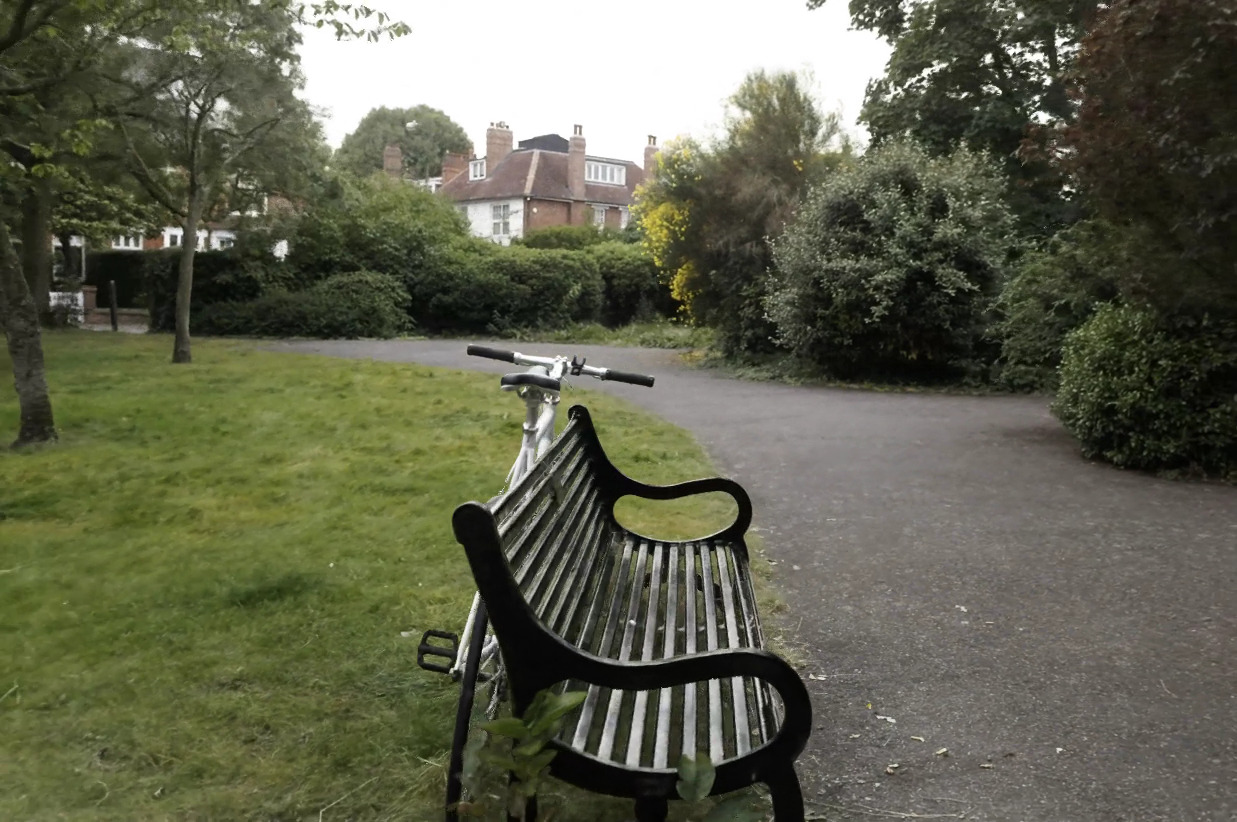}
     \end{subfigure}

     \centering
    \rotatebox{90}{\hspace{0.85cm}\footnotesize Depth}
    \hspace{-1.9mm}
     \begin{subfigure}[b]{0.24\linewidth} 
    \imagewithsquare{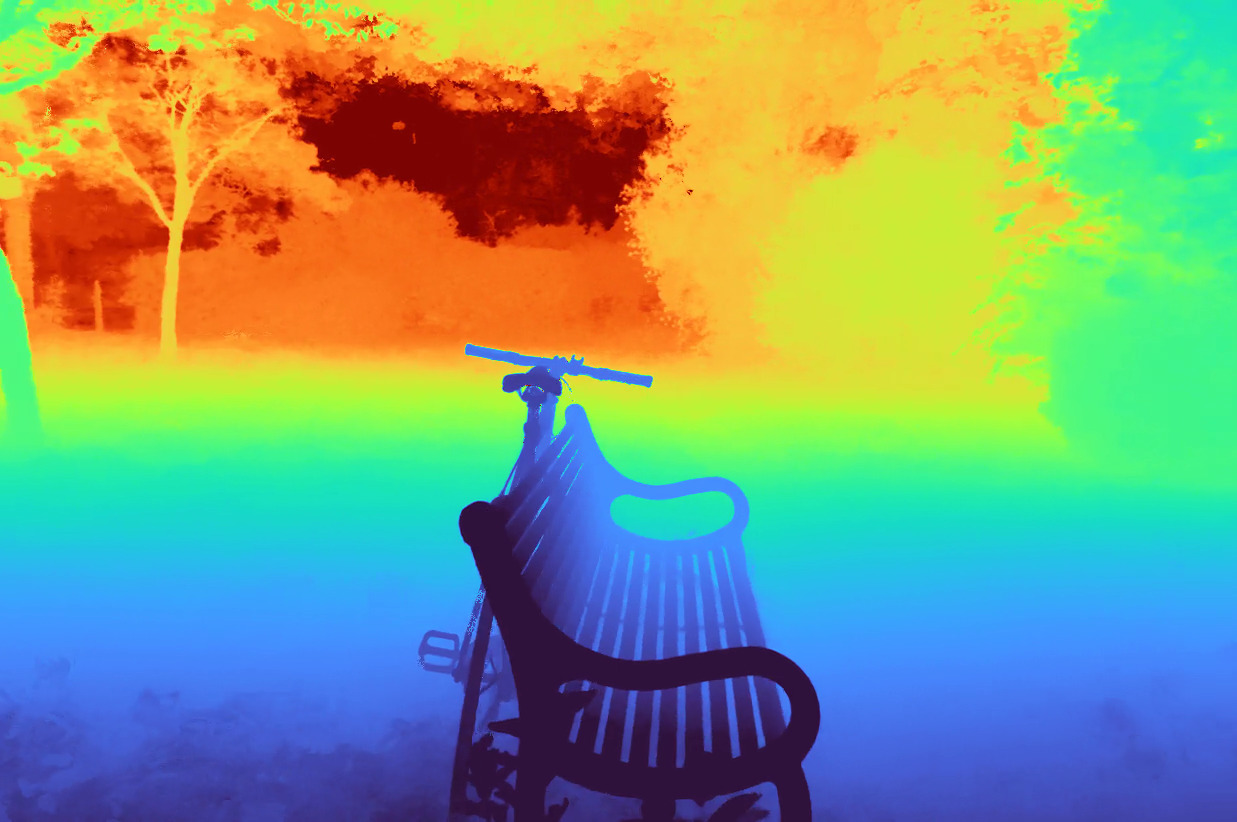}{(0, 0)}{(3, 0.5)}
     \end{subfigure}
     \begin{subfigure}[b]{0.24\linewidth}
    \imagewithsquare{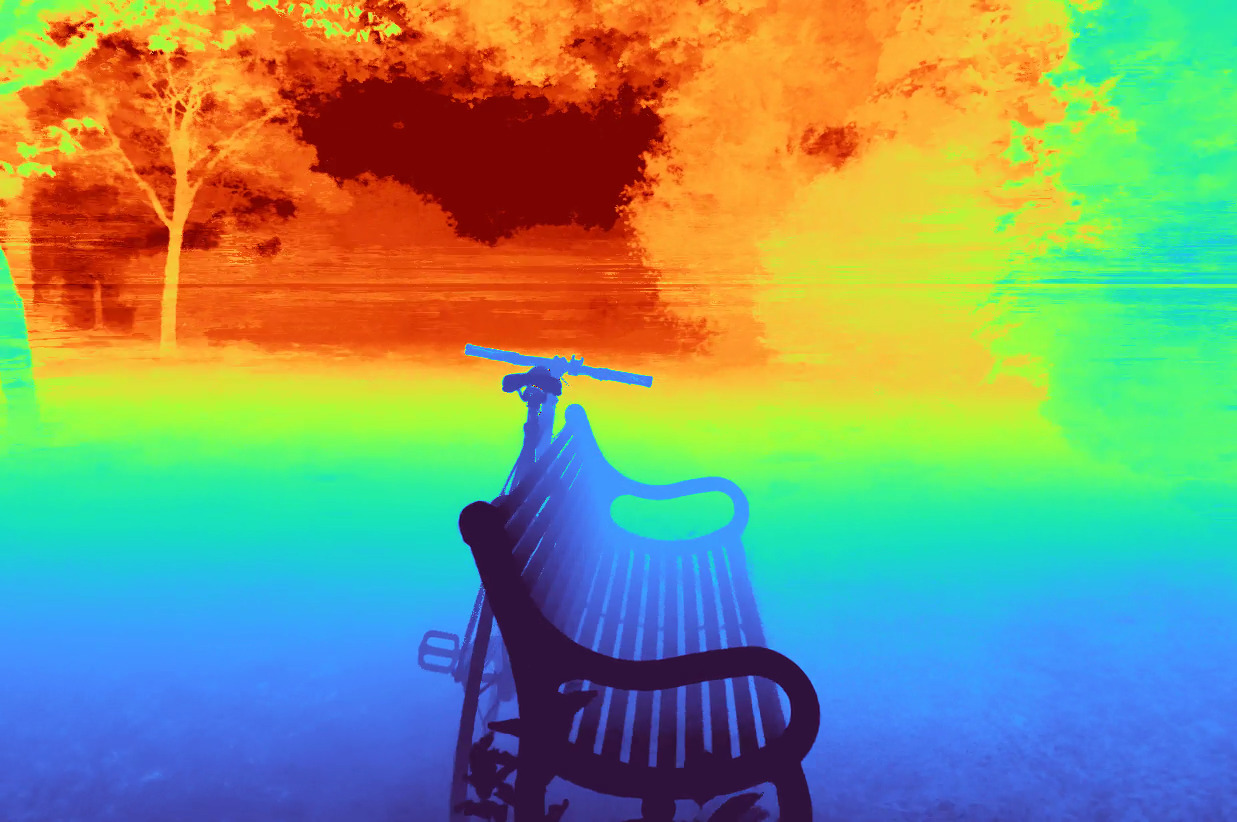}{(0, 0)}{(3, 0.5)}
     \end{subfigure}
    \begin{subfigure}[b]{0.24\linewidth}
\imagewithsquare{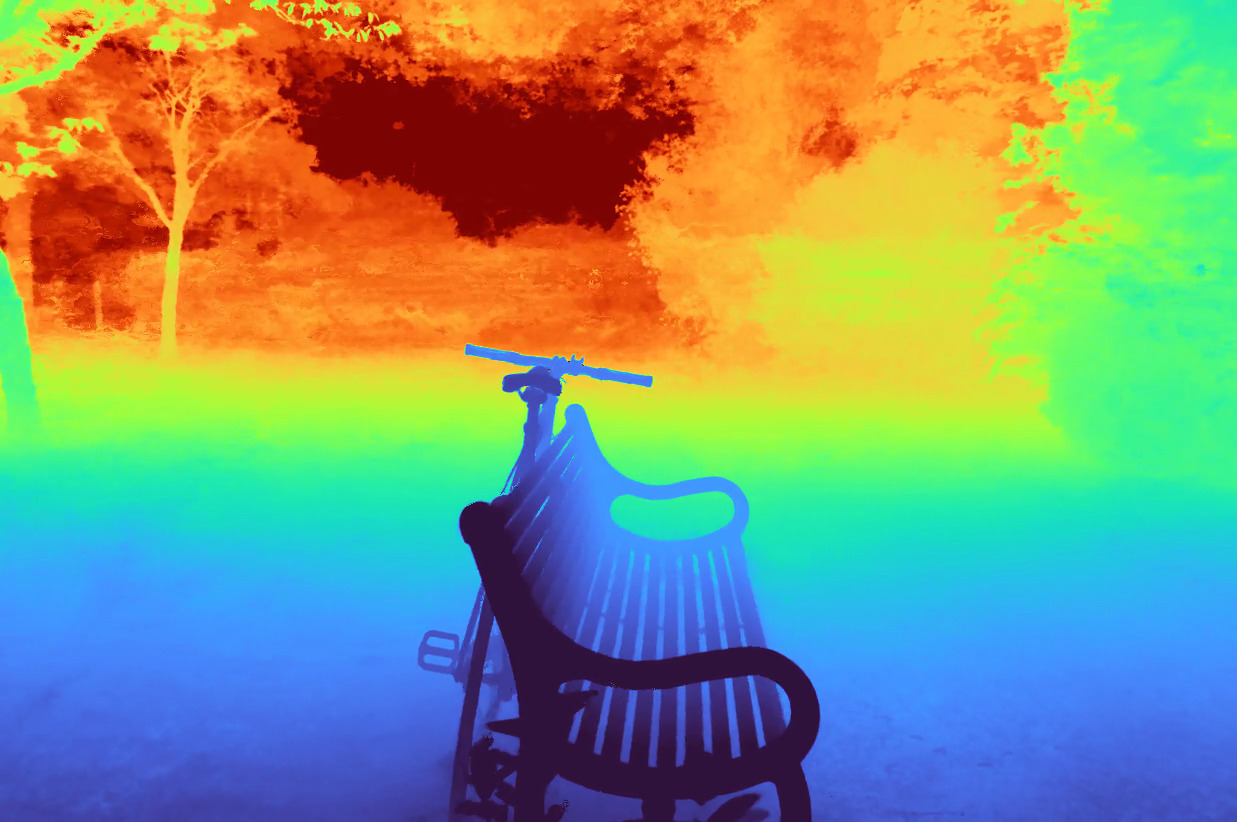}{(0, 0)}{(3, 0.5)}
     \end{subfigure}
     \begin{subfigure}[b]{0.24\linewidth}
         \imagewithsquare{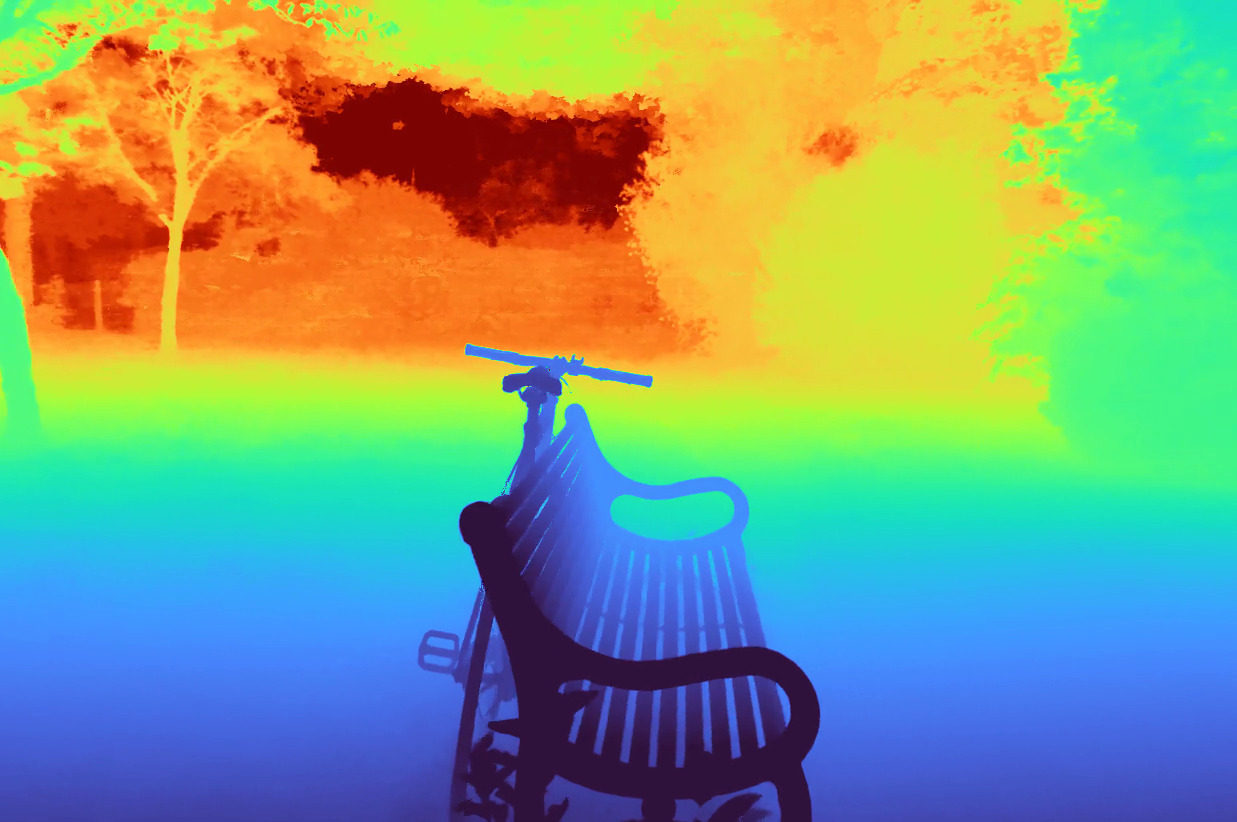}{(0, 0)}{(3, 0.5)}
     \end{subfigure}
     
        \caption{Comparison to MipNERF360~\cite{Barron2022MipNeRF3U} with different design choice, with and without $\mathcal{L}_{\mathrm{dist}}$, with a near plane at $0.2$ or $np=0$ and with or without our proposed gradient scaling $s_{\nabla p}$. We see here artifacts close to the camera near the floor. Our results combining $\mathcal{L}_{\mathrm{dist}}$ and $s_{\nabla p}$, shows the best reconstruction. The depth is represented with deep blue close to the camera and deep red far. As with other methods, this improvement is particularly visible in the videos available in Supplemental Materials.}
        \label{fig:mip360_comp}
\end{figure*}

We compare to MipNeRF360~\cite{Barron2022MipNeRF3U} in Fig.~\ref{fig:mip360_comp}. \final{MipNeRF360 uses an MLP and a custom frequency encoding to model the scene at several levels of detail, preventing aliasing. A non-linear scene parameterization allows to handle unbounded scenes.} As MipNeRF proposes a loss to limit the problem of background collapse, we show comparisons to different settings. We have three axes we act on, use of the  $\mathcal{L}_{\mathrm{dist}}$ proposed by MipNeRF360~\cite{Barron2022MipNeRF3U}, setting of the near plane to 0 ($np=0$) and use of our proposed scaling $s_{\nabla p}$. We can see that combining $\mathcal{L}_{\mathrm{dist}}$ with our scaling $s_{\nabla p}$ yields the best results.\final{ We believe the reason for this finding is that $\mathcal{L}_{\mathrm{dist}}$ has value beyond its partial effect on background collapse. It adds a solid density prior which is applicable to the scenes used in the paper, making the combination of our scaling and $\mathcal{L}_{\mathrm{dist}}$ beneficial.} As with the other volumetric representations, these effects are most visible in the videos available in Supplemental Materials.

\begin{table*}[h!]
  \begin{center}
    \begin{tabular}{l|l|c|c|c} 
        \hline
      \textbf{Method} & \textbf{Experiment} &\textbf{PSNR $\uparrow$} & \textbf{SSIM $\uparrow$} & \textbf{LPIPS $\downarrow$}\\
      \hline
      TensoRF & Truck w/o Gradient Scaling & 26.92 & 0.913 & 0.127\\
       & Truck with Gradient Scaling (Ours) & \textbf{27.13} & \textbf{0.914} & \textbf{0.124}\\
      \hline
      TensoRF & Caterpillar w/o Gradient Scaling & 25.84 & 0.910 & 0.136\\
       & Caterpillar with Gradient Scaling (Ours) & \textbf{26.19} & \textbf{0.912} & \textbf{0.133}\\
      \hline
      DVGOv2 & Kitchen w/o Gradient Scaling & 25.84 & 0.708 & 0.408\\
       & Kitchen with Gradient Scaling (Ours) & \textbf{25.91} & \textbf{0.712} & \textbf{0.400}\\
      \hline
      DVGOv2 & Bonsai w/o Gradient Scaling & 27.77 & 0.826 & 0.404\\
       & Bonsai with Gradient Scaling (Ours) & \textbf{28.03} & \textbf{0.828} & \textbf{0.403}\\
      \hline
      NGP & Croissant w/o Gradient Scaling & 30.94 & 0.941 & 0.101\\
       & Croissant with Gradient Scaling (Ours) & \textbf{31.22} & \textbf{0.943} & \textbf{0.097}\\
      \hline
      NGP & Exotic Plant w/o Gradient Scaling & 29.67 & 0.932 & 0.112\\
       & Exotic Plant with Gradient Scaling (Ours) & \textbf{29.92} & \textbf{0.935} & \textbf{0.110}\\
      \hline
       MipNeRF360 & Bicycle Original MipNeRF360 & \textbf{24.79} & \textbf{0.685} & \textbf{0.223}\\

        & Bicycle No $\mathcal{L}_{\mathrm{dist}}$ & 24.77 & 0.680 & 0.227\\
        & Bicycle No $\mathcal{L}_{\mathrm{dist}}$,  $np=0$, $s_{\nabla p}$ & 24.65 & 0.679 & 0.225 \\
        & Bicycle $\mathcal{L}_{\mathrm{dist}}$,  $np=0$, $s_{\nabla p}$ (Ours) & 24.75 & 0.682 & \textbf{0.223}\\
      \hline
       MipNeRF360 & Stump Original MipNeRF360 & 26.61 & 0.743 & 0.160\\

        & Stump No $\mathcal{L}_{\mathrm{dist}}$ & \textbf{26.64} & \textbf{0.744} & \textbf{0.159}\\
        & Stump No $\mathcal{L}_{\mathrm{dist}}$,  $np=0$, $s_{\nabla p}$ & 26.60 & 0.741 & 0.164 \\
        & Stump $\mathcal{L}_{\mathrm{dist}}$,  $np=0$, $s_{\nabla p}$ (Ours) & 26.62 & 0.742 & 0.161\\
      \hline
    \end{tabular}
    \caption{We compare PSNR, SSIM, and LPIPS metrics for the evaluated methods(\cite{Barron2022MipNeRF3U, Chen2022ECCV, mueller2022instant, SunSC22, SunSC22_2}) with and without our gradient scaling. Overall our scaling leads to better metrics for all representations, with MipNeRF360~\cite{Barron2022MipNeRF3U} leading to a similar quality. The floater effect has a relatively low impact on visual metrics, but is perceptually disturbing and clearly visible in the qualitative evaluations and videos in the supplemental material.}
    \label{tab:metric}
  \end{center}
\end{table*}
\paragraph*{Details}
MipNeRF360 uses a \final{non-linear} coordinate \final{parameterization} to fit an unbounded space in a bounded box and distributes samples linearly in disparity space to compensate for the contraction of coordinate, using a function $g(x) = 1/x$ to parametrize the rays.
Using this same function with a near plane very close to 0 would put most samples just in front of the camera, in our tests when setting the near plane to zero we use $g(x) = 1/(1+x)$.
\final{As described in Sec.\ref{sec:method_jacobian}, we don't use the Jacobian rescaling for MipNeRF360 because the central part of the scene's coordinates, which contains most cameras, is unaffected by the reparameterization}.

\subsection{Quantitative evaluation and discussions}
We report PSNR, SSIM, and LPIPS \cite{zhang2018perceptual} metrics for the evaluated methods with and without our gradient scaling in Tab.~\ref{tab:metric} and for the different variants of MipNeRF360.
\final{All metrics are computed on held-out test images, provided with the dataset when possible, or randomly selected.}

Overall our method leads to better metrics for most tests except for MipNeRF360. While the quantitative difference is small, it mainly depends on the visibility of the floaters in test views, which might be a relatively sparse event depending on the dataset. Further, setting the near plane to zero allocates slightly fewer samples to the center of the scene which can have a slightly negative impact on reconstruction when there is no near-camera geometry.
Floaters are more likely to be visible when test cameras are slightly further away from the scene center than training cameras. We find that the qualitative evaluation test paths, shown in Supplemental Material, better show the improvement provided by our approach.

\section{Conclusion}
We present a simple, yet efficient, gradient scaling approach, removing the need for near-plane setting in NeRF-like methods while preventing background collapse. Our method is computationally efficient and solves a \final{sampling imbalance} existing in most published approaches. This is particularly important in capture scenarios where objects are arbitrarily close or at varying distances from the cameras. Our scaling is directly applicable to most NeRF-like representations and can be easily integrated with a few lines of code.

\section*{Acknowledgements}
We thank Jon Barron and Peter Hedman for their help in designing Figure 2.


\bibliographystyle{eg-alpha-doi}  
\bibliography{main.bib}        


\end{document}